\documentclass[
nolists,
nofoot,
nocopyright,
subscriptcorrection,
upint,
varvw,
barcolor=Goldenrod3,
mathalfa=cal=euler,
balance,
hyphenate,
french]{asmejourarxiv} %

\usepackage{epsfig}
\graphicspath{ {./images/} }
\usepackage{subcaption}
\usepackage{lipsum}
\usepackage{booktabs}
\usepackage{multirow}
\usepackage{enumitem}

\usepackage[svgnames]{xcolor}
  \definecolor{diffstart}{named}{Grey}
  \definecolor{diffincl}{named}{Green}
  \definecolor{diffrem}{named}{OrangeRed}

\usepackage{amsthm}
\usepackage{algorithm,algpseudocode} 

\newtheoremstyle{asmeplain}%
  {6pt}{6pt}
  {\itshape}
  {}
  {\scshape}
  {.}
  {.5em}
  {}

\newtheoremstyle{asmedefn}%
  {6pt}{6pt}%
  {}%
  {}%
  {\scshape}%
  {.}%
  {.5em}%
  {}

\theoremstyle{asmeplain}
\newtheorem{boxedtheorem}{Theorem}
\newtheorem{boxedlemma}[boxedtheorem]{Lemma}

\newtheorem{boxedcorollary}[boxedtheorem]{Corollary}
\newtheorem{boxeddefinition}[boxedtheorem]{Definition}

\renewcommand{\Im}{\text{Im}}
\def\Ker{\text{Ker}}

\usepackage{scalerel}

\DeclareMathOperator*{\argmin}{arg\,min}

\def\bb{\boldsymbol{b}}
\def\bc{\boldsymbol{c}}

\def\bf{\boldsymbol{f}}
\def\bg{\boldsymbol{g}}

\def\bn{\boldsymbol{n}}

\def\bp{\boldsymbol{p}}
\def\bq{\boldsymbol{q}}
\def\br{\boldsymbol{r}}
\def\bs{\boldsymbol{s}}

\def\bu{\boldsymbol{u}}
\def\bv{\boldsymbol{v}}
\def\bw{\boldsymbol{w}}
\def\bx{\boldsymbol{x}}
\def\by{\boldsymbol{y}}
\def\bz{\boldsymbol{z}}
\def\bA{\boldsymbol{A}}

\def\bC{\boldsymbol{C}}

\def\bF{\boldsymbol{F}}

\def\bI{\boldsymbol{I}}

\def\bP{\boldsymbol{P}}

\def\bT{\boldsymbol{T}}
\def\bU{\boldsymbol{U}}

\def\bZ{\boldsymbol{Z}}

\def\bAcal{\mathbf{{A}}}

\def\bCcal{\mathbf{{C}}}
\def\bDcal{\mathbf{{D}}}

\def\bFcal{\mathbf{{F}}}
\def\bGcal{\mathbf{{G}}}

\def\bJcal{\mathbf{{J}}}

\def\bLcal{\mathbf{{L}}}
\def\bMcal{\mathbf{{M}}}

\def\bUcal{\mathbf{{U}}}

\def\bPhi{\boldsymbol{\Phi}}

\def\bgamma{\boldsymbol{\gamma}}

\def\bomega{\boldsymbol{\omega}}
\def\bOmega{\boldsymbol{\Omega}}

\def\btau{\boldsymbol{\tau}}

\newcommand{\beq}{\begin{equation}}
\newcommand{\eeq}{\end{equation}}

\hypersetup{%
	pdfauthor={Bryce Palmer},                       		   	
	pdftitle={palmer_relcp_2026},                  	
	pdfkeywords={Rigid body dynamics, Nonsmooth dynamics, Complementarity-based time-stepping, Linear complementarity problem, ReLCP}, 
	pdfsubject = {Reformulating complementarity-based time-stepping for nonsmooth contact between smooth rigid bodies},			
	pdflicenseurl={https://ctan.org/pkg/asmejour},
}

\JourName{ArXiv}

\begin{document}


\SetAuthorBlock{Bryce Palmer\CorrespondingAuthor}{Center for Computational Biology,\\
   Flatiron Institute,\\
   New York, NY 10010, USA\\
   email: bpalmer@flatironinstitute.org} 

\SetAuthorBlock{H. Metin Aktulga}{Department of Computer Science,\\
   Michigan State University,\\
   East Lansing, MI 48824 USA\\
   email: hma@msu.edu} 

\SetAuthorBlock{Tong Gao}{Department of Mechanical Engineering,\\
   Tufts University,\\
   Medford, MA 02155 USA\\
   email: tong.gao@tufts.edu} 


\title{Smooth-Rigid-Body Contact as a ReLCP: A Recursively Generated Linear Complementarity Problem}


\begin{abstract}
This paper reformulates complementarity-based time-stepping for frictionless nonsmooth contact between smooth rigid bodies as a recursively generated linear complementarity problem (ReLCP), involving a sequence of LCPs of increasing dimension. Starting from a classical single-constraint shared-normal signed-distance (SNSD) LCP, the method adds unilateral constraints only when the discrete-time update predicted by the current contact set would violate nonpenetration of the underlying smooth surfaces. The resulting procedure acts directly on smooth geometry, enforces nonpenetration to a prescribed tolerance, and avoids the oversampling inherent to proxy-surface contact models such as tessellations or multi-sphere decompositions, for which improved geometric fidelity can drive rapid growth in constraint count and cost. For strictly convex bodies, we prove that an initially overlap free configuration with sufficiently small timestep sizes, imply finite termination of the adaptive augmentation, and yield a unique discrete-time velocity update. In the limit $\Delta t \to 0$ and for any fixed overlap-free discrete state with a fixed geometric overlap tolerance, we prove that the recursion terminates after the initial solve, reducing the method to the classical single-constraint SNSD LCP and retaining the usual consistency of complementarity time-stepping with the underlying differential variational inequality. Numerical tests on colliding ellipsoids, compacting ellipsoid suspensions, growing bacterial colonies, and taut chainmail networks demonstrate stable large-timestep behavior, bounded interpenetration without discretization-induced surface roughness, and substantial reductions in both active constraint counts and runtime relative to representative discrete-surface complementarity formulations.
\end{abstract}

\date{Version \versionno, \today}

\maketitle 


\section{Introduction}
The simulation of collision and contact dynamics in rigid and flexible body systems has long occupied a central place in scientific computing, engineering, and computer graphics. Broadly speaking, computational approaches to contact may be grouped into three classes: piecewise-smooth methods, smooth penalty-based methods, and nonsmooth complementarity-based methods. Piecewise-smooth formulations seek to identify the precise times and locations of collision events and to apply instantaneous impulses so as to enforce conservation laws exactly. Although these methods are conceptually natural and analytically appealing, they are seldom used in large-scale simulations because accurate event detection for complex geometries is prohibitively difficult and because densely collisional systems impose severe timestep restrictions.

Smooth methods avoid explicit event resolution by regularizing contact through repulsive potentials or penalty forces, thereby replacing discontinuous collision laws with ordinary differential equations. Their algorithmic simplicity, computational efficiency, and ease of implementation have made them among the most widely used approaches in many engineering and animation applications. This regularization, however, introduces well-known trade-offs. Stronger repulsive potentials reduce overlap but increase numerical stiffness and therefore require smaller timesteps for stability; pairwise additive contact forces limit the transmission of force through contact networks, thereby introducing finite speed-of-sound effects; and the resulting contact laws depend on non-physical parameters that must be tuned to balance stability against accuracy \cite{arman_demp_vs_demc_2017}.

Nonsmooth methods likewise avoid explicit collision-time resolution, but instead of smoothing the contact law they discretize the underlying continuous-time differential variational inequalities governing contact and impact. This complementarity-based time-stepping perspective permits any number of bodies to enter or leave contact within a single timestep and supports stable timesteps that may exceed those of smooth methods by several orders of magnitude for the same system \cite{servin_nonsmooth_vs_smooth_dem_2014}, without introducing fitting parameters or finite speed-of-sound effects \cite{arman_demp_vs_demc_2017}. The framework is also mathematically mature: convergence as timestep size decreases, together with uniqueness of the resulting center-of-mass dynamics under reasonable assumptions, has been established even in the presence of bilateral constraints and friction \cite{anitescu_convergence_2008, anitescu_2006, anitescu_hart_2004}. Its principal drawback is instead algorithmic and computational. Relative to smooth methods, complementarity-based formulations require more elaborate constraint generation and incur higher per-timestep setup and solve costs. Even so, as algorithmic and computational advances continue to reduce this overhead, nonsmooth methods remain a compelling route to high-fidelity, scalable time integration in contact-dominated systems.

Independent of the choice of smooth or nonsmooth time integration, simulations of non-spherical bodies must specify the surface on which contact is evaluated; this surface representation fixes the local gaps and normals seen by the contact law and can therefore change particle-scale interactions and the bulk behavior produced by many contacts. Even when the intended body is smooth or piecewise smooth, the numerical method often acts on a proxy surface: a tessellation of planar faces, a multi-sphere decomposition, or some other discrete approximation. In the context of smooth discrete element methods, extensive studies have shown that coarse proxies introduce artificial surface roughness that can alter bulk behavior through artifacts such as premature jamming or modified mixing rates \cite{markauskas_multisphere_adequacy_2010, hohner_temporal_comparison_2011, hohner_shape_effects_dem_2012, hohner_dem_2015, liu_multi_super_ellipsoid_2020, you_dem_comparison_se_2018, liu_comparative_2021}. Increasing geometric resolution mitigates these artifacts but drives rapidly growing costs \cite{markauskas_multisphere_adequacy_2010}, and classical multi-sphere DEM does not, in general, converge to the dynamics of the underlying smooth body without corrective terms \cite{hohner_temporal_comparison_2011}. These observations motivate approaches that operate directly on smooth surfaces, whether through surface-to-surface minimization \cite{liu_multi_super_ellipsoid_2020, you_dem_comparison_se_2018, liu_comparative_2021} or geometrically informed potentials~\cite{zorin_geom_pot_2025}.

In contrast, the role of surface representation in nonsmooth multibody dynamics has received comparatively little systematic attention. Many formulations adopt specific proxy surfaces without examining the consequences of that choice beyond visual plausibility; a notable exception is the review of Andres, Erleben, and Ferguson~\cite{erleben_contact_class_2022} and the lines of works on geometrically implicit methods~\cite{chakraborty_geom_implicit_2014,xie_line_surface_contact_2016, xie_multiple_contact_patches_2018}. The relevant issue is not merely how contacts are detected, but which directions of relative motion are constrained and at what geometric resolution. More broadly, the limited discussion of proxy surfaces in nonsmooth dynamics has left important gaps in understanding regarding the validity and limitations of particular contact constructions and has led to the use of some techniques in settings where their physical justification is weak. There is therefore a clear need to examine how surface representation affects nonsmooth multibody dynamics and to develop formulations that address this issue directly.

The present work revisits complementarity-based time integration for frictionless unilateral contact between smooth rigid bodies in both inertial and overdamped settings. We show that single-constraint shared-normal signed-distance methods can become unstable at large timestep sizes because unconstrained orthogonal motions may induce overlap, whereas multi-sphere constructions enforce non-overlap only to the proxy resolution and incur rapidly growing constraint counts as the proxy is refined. Against this background, the main contributions are fourfold. First, we reformulate smooth-surface contact time-stepping as a recursively generated sequence of LCPs whose dimension grows only when additional constraints are required by the trial update. Second, we abstract this construction into a new problem class, termed a recursively generated linear complementarity problem (ReLCP), and prove general sufficient conditions for convergence in finite iterations and uniqueness of the converged result, or of an induced quantity of interest. Third, we specialize this framework to frictionless contact between strictly convex smooth bodies and prove, for each fixed overlap-free discrete state, that sufficiently small timestep sizes imply finite termination of the contact augmentation together with existence and uniqueness of the discrete-time velocity update. In the same contact setting, we prove that in the limit as $\Delta t \to 0$, the trial configuration produced by the initial single-constraint SNSD-LCP solve already satisfies any fixed geometric overlap tolerance, so that no further constraints are generated and the formulation reduces to the classical single-constraint SNSD LCP, thereby recovering the standard small-timestep consistency with the underlying differential variational inequality. Fourth, we provide numerical examples that examine convergence, stability, and computational scaling relative to representative single-constraint and multi-sphere formulations.

\subsection{Related work}
Linear-complementarity time-stepping for nonsmooth contact is classical. Foundational formulations by Stewart and Trinkle, Anitescu and Potra, and related analyses of Stewart, together with the general theory of complementarity problems developed by Cottle, Pang, and Stone, established LCP-based updates as a standard framework for rigid multibody contact \cite{stewart_first_lcp_1996, anitescu_potra_1997, anitescu_portra_time_stepping_2002, stewart_frictional_review_2000, CPS1992}. These classical formulations are geometrically explicit: contact geometry is frozen or linearized over a timestep, so each discrete update is a fixed LCP. This fixed-problem structure supports analyses of existence, convergence, and uniqueness of the induced dynamics under standard assumptions \cite{anitescu_convergence_2008, anitescu_2006, anitescu_hart_2004}, and it gives the formulation a practical numerical advantage: each step reduces to a structured LCP solvable by established complementarity or convex-QP algorithms, rather than to a nonlinear iteration whose success may depend on initialization or backtracking. 

Nevertheless, freezing or linearizing contact geometry can decouple the enforced constraints from the configuration produced by the update, allowing the resulting update to violate nonpenetration of the underlying smooth surfaces even when the initial geometry is non-overlapping. Unconstrained overlaps can be mitigated through smaller timesteps or finer proxy surfaces, but these remedies introduce their own costs and do not address the underlying issue of explicitly decoupling contact geometry from the resulting update. Geometrically implicit formulations remove this frozen-geometry approximation by coupling contact geometry to contact wrench at the cost of solving a mixed NCP at each step. Chakraborty et al. introduced this approach \cite{chakraborty_geom_implicit_2014}. Subsequent extensions treated multiple convex contact patches and non-point contact \cite{xie_line_surface_contact_2016, xie_multiple_contact_patches_2018}. These works prove a conditional consistency result: if the mixed NCP admits a solution, the resulting update satisfies the intended discrete nonpenetration conditions. Note, this result does not establish that such a solution exists or that it can be reached in finitely many iterations.

Other approaches retain the nonlinear contact equations and solve them directly at each timestep, using either fixed-point sequences of relinearized LCPs \cite{todorov_impl_nonlin_comp_2010} or nonsmooth Newton methods \cite{nvidia_nonsmooth_newton_2019}. Because the constitutive laws, constraint functions, and constraint Jacobians may be evaluated at the unknown post-update configuration, such formulations can accurately capture nonlinear material response and feedback from bilateral constraints within the timestep solve. The primary disadvantage of these methods is their tendency to require careful initialization or backtracking to ensure convergence and their potential for fluctuating or divergent iterations \cite{todorov_impl_nonlin_comp_2010}.

At the algorithmic level, ReLCP resembles active-set methods. Classical active-set algorithms solve a fixed inequality-constrained quadratic program: they update a working set of candidate active constraints, solve the associated reduced systems, and terminate when the optimality conditions of that fixed problem hold \cite{goldfarb_idnani_1983, nocedal_wright_2006}. Pivoting and active-set interpretations of monotone LCPs likewise select complementary variables within a fixed problem \cite{CPS1992, erleben_contact_class_2022}. ReLCP differs because the relevant geometric constraints are not drawn from a fixed, finitely enumerable candidate set; each trial configuration can expose constraints absent from the current complementarity problem. The recursion is therefore not an active-set solver for a fixed QP or LCP, but a formulation in which the complementarity problem itself is augmented by the predicted smooth-surface update.

\section{Preliminaries} \label{sec:preliminaries}
This section reviews the mathematical properties of complementarity-based time integration for nonsmooth multibody contact, with an emphasis on its core assumptions, the practical implications of these assumptions, and common misconceptions about these methods. The focus is on a subset of nonsmooth multibody dynamics related to inelastic collisions, as introduced by Stewart et al. \cite{stewart_first_lcp_1996}.

\subsection{Nonsmooth rigid multibody dynamics}
\subsubsection{Nonsmooth generalized dynamics}
Nonsmooth multibody dynamics is naturally expressed in an abstract form through a generalized configuration vector. For example, consider a system of \(N\) rigid bodies where each body's state is described by its center of mass, \(\bx_\ell(t)\), and its orientation, \(\bp_\ell(t)\) for $\ell\in[1, N]$. To avoid the singularities of Euler angles, orientations are often represented as unit quaternions, \(\bq_\ell(t) = [s_\ell,\bw_\ell]^\top\) (with scalar part \(s_\ell\) and vector part \(\bw_\ell\)). The complete configuration of the system is then given by the column vector
\(
\bCcal(t) = \left[\ldots, \bx_\ell^\top, s_\ell, \bw_\ell^\top, \ldots\right]^\top,
\)
which has 7 degrees of freedom per body (with one redundancy due to the unit-norm constraint on the quaternions). While the following discussion focuses on rigid-body collisions, the number of configurational variables naturally increases for systems with flexible bodies (e.g., a bacterial colony modeled as growing rods, as discussed in Section \ref{sec:examples}, gains additional configurational variables related to the dynamically changing lengths).

Under the action of center-of-mass forces \(\bf_{\ell}\) and torques \(\btau_{\ell}\), each rigid body moves with translational velocity \(\bu_{\ell}\) and angular velocity \(\bomega_{\ell}\). For simplicity, these body-level properties are again collected into system-level vectors denoted by their uppercase counterparts, i.e.,
\(
\bFcal(t) = \left[\ldots, \bf_\ell^\top, \btau_\ell^\top, \ldots\right]^\top \in \mathbb{R}^{6N}
\)
and 
\(
\bUcal(t) = \left[\ldots, \bu_\ell^\top, \bomega_\ell^\top, \ldots\right]^\top\in \mathbb{R}^{6N}.
\)
In classical mechanics, these velocities are assumed continuous, so that
\(
\frac{d \bx_\ell}{dt} = \bu_\ell,\quad \frac{d \bp_\ell}{dt} = \bomega_\ell
\)
or more generally, 
\(
\dot{\bCcal} = \bGcal(\bCcal) \bUcal,
\)
where $\bGcal$ is a configuration-specific linear mapping from standard to generalized coordinates. However, in the presence of collision events\textemdash where impulses induce discontinuities\textemdash the generalized velocity \(\dot{\bCcal}(t)\) may exhibit sudden jumps, making \(\dot{\bCcal} = \bGcal(\bCcal) \bUcal\) meaningless in the classical sense. Nevertheless, following Anitescu et al. \cite{anitescu_2006}, there are reasonable assumptions that one can make about contact dynamics that allow it to be mapped into a more amicable form: 
\begin{enumerate}[wide=0pt]
    \item \(\bCcal(t)\) is continuous; bodies do not ``teleport'' during collisions.
    \item The speed \(\|\dot{\bCcal}(t)\|\) remains finite, ensuring the system's energy remains bounded.
    \item \(\dot{\bCcal}(t)\) is a function of bounded variation; that is, the overall change in generalized velocity over any finite time interval\textemdash the total variation\textemdash is finite.
    \item The number of collision events in any finite time interval is finite, and outside of these events, the velocity is continuous.
\end{enumerate}
These assumptions imply that \(\bCcal(t)\) is locally Lipschitz and satisfies 
\begin{equation}
\bCcal(t) = \bCcal(0) + \int_{0}^t \dot{\bCcal}(t')dt'.
\end{equation}
For discrete-time simulations, these assumptions allow us to discretize this weak formulation by evolving the configuration from \(t^k\) to \(t^{k+1} = t^k + \Delta t\) via left or right-sided Riemann sums:
\begin{equation} \label{eq:con_disc}
\bCcal^{k+1} \approx \bCcal^k + \Delta t\dot{\bCcal}^k 
\quad | \quad 
\bCcal^{k+1} \approx \bCcal^k + \Delta t\dot{\bCcal}^{k+1},
\end{equation}
effectively localizing the discontinuities to the discrete timestep locations. While Eq.~\eqref{eq:con_disc} may \emph{look} like a first order Taylor expansion, it cannot be attained directly without first making similar assumptions about the continuity, boundedness, and well-behavedness of  \(\bCcal(t)\) and \(\dot{\bCcal}(t)\). 

\subsubsection{Laws of motion}
A key strength of complementarity-based nonsmooth multibody dynamics is that it imposes relatively few restrictions on how forces between constituent bodies translate into motion. In the simplest scenario\textemdash typical of dry granular matter (e.g., hopper flow, landslides) or classical robotics\textemdash Newton’s second and third laws entirely govern the dynamics: collisions and external forces act on inertial bodies and their velocity evolves accordingly. Yet these methods readily accommodate more complex laws of motion: bodies may be suspended in a viscous fluid (introducing drag or hydrodynamic coupling), or they may experience long-range interactions from electromagnetic or gravitational fields. The resulting equations can span purely inertial to strictly overdamped limits. As long as the discretized weak form of the dynamics can be expressed in a closed-form update rule mapping $\bFcal^k$ to $\bCcal^{k+1}$ (with certain reasonable restrictions), complementarity-based methods may be applied.

Presenting a fully generalized formulation of this framework is beyond the present scope. Instead, to illustrate its flexibility, two representative systems are considered: (i) dry, inertial dynamics and (ii) overdamped, local-drag dynamics. In the first scenario, the dynamics of constituent bodies follow Newton’s laws of motion:
\begin{equation}
M(\bCcal)\dot{\bUcal} = \bFcal,
\end{equation}
where \(M(\bCcal)\in \mathbb{R}^{6N \times 6N}\) is a block-diagonal, symmetric positive-definite mass matrix of the form
\begin{equation}
M(\bCcal)=\mathrm{diag}(\ldots,m_\ell\bI_{3\times 3},\bI_{\ell},\ldots),
\end{equation}
with \(m_\ell\) and \(\bI_\ell\) denoting the mass and moment of inertia of the \(\ell\)-th body, respectively. As discussed in Section \ref{sec:lcp_ccqpp}, this positive-definiteness is essential for ensuring the existence and uniqueness of solutions in nonsmooth methods. Following Anitescu et al. \cite{anitescu_2006}, the inertial formula is discretized as
\begin{equation}
\bUcal^{k+1} = \bUcal^k + \Delta tM^{-1}_k\bFcal^k.
\end{equation}
The necessary update rule, which is a closed-form map from $\bFcal^k$ to $\bCcal^{k+1}$, is then attained using a right-sided Riemann sum (Eq.~\eqref{eq:con_disc}):
\begin{equation} \label{eq:law_inertia}
\begin{aligned}
    \text{Implicit: }& &&\bCcal^{k+1} = \bCcal^{k} + \Delta t \bGcal^{k+1}\left(\bUcal^{k} + \Delta t M^{-1}_k \bFcal^{k}\right), \\
    \text{Implicit (linear): }& &&\bCcal^{k+1} = \bCcal^{k} + \Delta t \bGcal^k\left(\bUcal^{k} + \Delta tM^{-1}_k \bFcal^{k}\right). \\
\end{aligned} 
\end{equation}
Note, the original right-sided update depends implicitly on the unknown configuration $\bCcal^{k+1}$ since $\bGcal^{k+1} = \bGcal\left(\bCcal^{k+1}\right)$, preventing a closed-form solution. This is addressed by approximating $\bGcal^{k+1} \approx \bGcal^{k}$ by assuming that rotations are small, an approximation that holds to first order in $\Delta t$ (as can be seen by substituting in $\bGcal^{k+1} \approx \bGcal^{k} + \Delta t \dot{\bGcal}^{k}$ and canceling higher order terms).

In contrast, consider a suspension of rigid bodies immersed in a fluid whose viscous effects dominate inertial forces. Newton’s laws still apply, but the dominance of viscous drag permits the assumption that inertial-effects are negligible, meaning that the velocity \(\bUcal\) follows immediately from the balance of forces: 
\begin{equation}
\bUcal = \bMcal\bFcal,
\end{equation}
where \(\bMcal\in \mathbb{R}^{6N\times6N}\) is referred to as the mobility matrix. If long-range hydrodynamic interactions are included, \(\bMcal\) is dense; conversely, in dilute or simplified settings where body–body fluid coupling can be neglected, \(\bMcal\) reduces to a block-diagonal ``local-drag'' matrix. The time discretization for this overdamped system is straightforward:
\begin{equation}
\bUcal^{k} = \bMcal^{k}\bFcal^{k},
\end{equation}
In contrast to inertial systems, forces computed at each timestep instantaneously translate into velocities. As a result, the update rule for overdamped systems may instead employ left-sided Riemann sums (Eq.~\eqref{eq:con_disc}) to form a map from $\bFcal^k$ to $\bCcal^{k+1}$:
\begin{equation} \label{eq:law_overdamped}
    \bCcal^{k+1} = \bCcal^{k} + \Delta t \bGcal^k\bMcal^{k} \bFcal^{k}
\end{equation}
Unlike the inertial system, which requires approximations to produce a closed-form relation, no such approximations are necessary here, as the dynamics are already linear.

\subsubsection{Nonsmooth contact time-stepping as a constraint optimization problem}

Instead of treating the underlying contact dynamics as piecewise smooth by finding the exact times and locations of collision events\textemdash a task that is both numerically stiff and algorithmically complex\textemdash complementarity-based time-stepping localizes discontinuities induced by collision events to the discrete timestep locations, allowing any number of bodies to enter or leave contact in a single timestep. Central to this viewpoint is the reformulation of the discrete-time contact update as a constrained optimization problem with constraints between nearby bodies, which enforce that bodies cannot overlap, non-contacting bodies do not generate collision force, and constraint forces satisfy both Newton's laws and D'Alembert's principle. This approach necessitates consideration of three key concerns: (1) the formulation of collision constraints, (2) the generation of constraints between nearby bodies, and (3) the numerical methods used to solve the resulting complementarity problem. In the following discussion, attention is first restricted to \emph{how} the constraints affect the dynamics, postponing the discussion of constraint generation and solution until after the physics of these constraints and the assumptions they make are clear.

\subsubsection{Inelastic no-overlap constraints}

Multiple no-overlap constraints may exist between any pair of nearby bodies, each having the same functional form. Consider the \(\alpha\)\textsuperscript{th} constraint as defined between a pair of surface points \(\by_a^{\alpha}\) and \(\by_b^{\alpha}\) on bodies \(a\) and \(b\), respectively. These points are selected such that their surface normals satisfy \(\hat{\bn}_a^\alpha = -\hat{\bn}_b^\alpha\). When the points coincide (\(\|\by_a^{\alpha} - \by_b^{\alpha}\| = 0\)), this condition ensures that the resulting contact force\textemdash directed along the surface normal\textemdash acts equally and oppositely on the two bodies, in accordance with Newton’s third law. Because the force direction is fixed, the constraint is fully characterized by a single non-negative Lagrange multiplier $\gamma_\alpha$, which encodes the magnitude of the repulsive collision force.

The total force and torque acting on body $a$ from the set $\mathcal{A}_a$ of its associated constraints are then given by:
\begin{equation}\label{eq:collision_ft}
   \bF_a = -\sum_{\alpha\in \mathcal{A}_a} \hat{\bn}_\alpha^a \gamma_\alpha, \quad 
   \bT_a = -\sum_{\alpha\in \mathcal{A}_a} (\by_a - \bx_a) \times \left(\hat{\bn}_\alpha^a \gamma_\alpha\right).
\end{equation}
In contact dynamics, these center-of-mass forces and torques are often referred to as contact wrenches. In order to avoid explicit event detection or timestep subdivision, this formulation must remain valid whether or not particles are in contact. As such, a signed separation function $\Phi_\alpha$ is introduced between \(\by_a^{\alpha}\) and \(\by_b^{\alpha}\), to allow for the detection of contact and the enforcement of vanishing contact force for separated particles. By convention, 
$\Phi_\alpha < 0$ indicates overlap, 
$\Phi_\alpha = 0$ indicates contact, and 
$\Phi_\alpha > 0$ indicates separation. The constraint force magnitude $\gamma_\alpha$ can then be constrained such that it vanishes whenever 
$\Phi_\alpha > 0$.

The final form of the inelastic no-overlap constraints is thus:
\begin{equation}
    \begin{aligned}
    \text{No contact: }& \Phi_\alpha\geq 0, \quad \gamma_\alpha=0, \\ 
    \text{Contact: }& \Phi_\alpha=0, \quad \gamma_\alpha\geq0.
    \end{aligned}
   \label{eq:flip_flop}
\end{equation}
Formally, Eq.~\eqref{eq:flip_flop} is known as a nonlinear complementarity constraint and is often abbreviated as $0 \leq \bPhi(\bCcal(\bgamma)) \perp \bgamma \geq 0,$ where $\bgamma = \left[\ldots, \gamma_{\alpha}, \ldots\right]^\top$ and $\bPhi = \left[\ldots, \Phi_{\alpha}, \ldots\right]^\top$. In this notation, the inequalities are applied element-wise, and the perpendicular symbol emphasizes that $\bPhi$ and $\bgamma$ are orthogonal, i.e., $\bPhi^\top\bgamma = 0$. This formula is termed nonlinear due to the nonlinear dependence of $\bPhi$ on $\bgamma$.

While intuitive, Eq.~\eqref{eq:collision_ft} can be derived from first principles using D’Alembert’s principle, which asserts that the total virtual work done by applied and constraint forces vanishes for all virtual displacements consistent with the constraints. In the current context, this principle ensures that a scalar-valued constraint $\Psi(\bCcal) = 0$ will induce a force $\bF_\Psi = \bGcal^\top\left(\nabla_{\bCcal}\Psi\right)^\top\gamma_\Psi$, where $\nabla_{\bCcal}\Psi$ is the gradient of the constraint with respect to the system's configuration, $\gamma_\Psi$ is the Lagrange multiplier used to enforce the constraint, and $\bGcal^\top$ maps the force from generalized to Cartesian coordinates. Applying this principle to Eq.~\eqref{eq:flip_flop} by letting $\Psi(\bCcal)\coloneqq \Phi_\alpha$, gives $\bFcal = \bGcal^\top(\nabla_{\bCcal} \bPhi)^\top\bgamma$, which exactly recovers Eq.~\eqref{eq:collision_ft} \cite[Appendix 3]{alens_2022}. For notational convenience, we introduce the shorthand $\bDcal = \bGcal^\top(\nabla_{\bCcal} \bPhi)^\top$, and reduce this form to $\bFcal = \bDcal \bgamma$.

\subsubsection{Constraint resolution}
With the structure and interpretation of the no-overlap constraints established, the next step is to solve for the, yet unknown, Lagrange multipliers  $\bgamma =\left[\ldots, \gamma_{\alpha}, \ldots\right]^\top$ such that the system evolution respects both the equations of motion and the complementarity condition given in Eq.~\eqref{eq:flip_flop}. This leads naturally to a differential variational inequality problem (DVI), which generalizes constrained dynamical systems by combining variational inequalities with differential equations \cite{stewart_convergence_lcp_1998, stewart_dvi_2008}. For the overdamped law of motion in Eq.~\eqref{eq:law_overdamped}, the resulting discrete-time DVI with nonlinear complementarity constraints takes the form:
\begin{align}\label{cont_dvi}
\bCcal^{k+1} &= \bCcal^k + \Delta t\bGcal^k \bMcal^k \left(\bFcal_{\text{ext}}^k + \bDcal^k\bgamma^k\right), \\
        \quad\text{s.t. } &0 \leq \bPhi^{k+1}
        \perp \bgamma^k \geq 0. \nonumber
\end{align}

Solving the nonlinear complementarity problem (NCP) described in Eq.~\eqref{cont_dvi} directly\textemdash e.g., via Newton-type methods or fixed point iteration\textemdash can be computationally intensive and, outside special settings, lacks general guarantees that a solver will reach a solution of the full discrete-time rigid-body contact update in finite time or that the resulting update will be unique \cite{todorov_impl_nonlin_comp_2010, nvidia_nonsmooth_newton_2019, chakraborty_geom_implicit_2014, xie_multiple_contact_patches_2018}. To address this, one may linearize the time evolution of the constraint functions $\bPhi$ by approximating their discrete-time update using linearized left- or right-sided Riemann sums. This approximation replaces the nonlinear dependence of $\bPhi^{k+1}$ on the unknown configuration by an affine dependence on $\bgamma^k$, converting the timestep problem into an LCP.

The appropriate discretization depends on the governing equations of motion: left-sided Riemann sums for overdamped dynamics and linearized, right-sided Riemann sums for inertial systems:
\begin{align}\label{eq:linear_sep}
    \text{Left: }\, \bPhi^{k+1} \approx&\, \bPhi^{k} + \Delta t \left(\nabla_{\bCcal^{k}} \bPhi^{k}\right) \dot{\bCcal}^k,\\
    \text{Right: }\, \bPhi^{k+1} \approx&\, \bPhi^{k} + \Delta t \left(\nabla_{\bCcal^{k+1}} \bPhi^{k+1}\right) \dot{\bCcal}^{k+1}, \\
    \text{Right (linear): }\, \bPhi^{k+1} \approx&\, \bPhi^{k} + \Delta t \left(\nabla_{\bCcal^{k}} \bPhi^{k}\right) \dot{\bCcal}^{k+1}.
\end{align}
In the inertial case, the right-sided update depends implicitly on the unknown configuration \(\bCcal^{k+1}\), which prevents a closed-form solution. To address this, one typically approximates $\nabla_{\bCcal^{k+1}} \bPhi^{k+1} \approx \nabla_{\bCcal^{k}} \bPhi^{k}$, assuming that relative location and orientation of each collision constraint remains unchanged during the timestep. This yields a closed-form, semi-implicit scheme accurate to first order in $\Delta t$ and transforms the original nonlinear complementarity problem into a linear complementarity problem (LCP) in $\bgamma^k$.

Linearizing the dependence of $\bPhi^{k+1}$ on the configuration $\bCcal^{k+1}$, and thereby approximating its overall dependence on $\bgamma^k$, yields both mathematical and numerical advantages. First, it places each timestep in the classical LCP setting, where solvability and uniqueness of the induced velocity update can be established under the assumptions summarized in Section~\ref{sec:lcp_ccqpp}. Second, as $\Delta t \rightarrow 0$, the resulting discrete trajectories converge to the underlying DVI solution \cite{stewart_convergence_lcp_1998, anitescu_convergence_2008}. Numerically, the LCP admits an equivalent convex quadratic program (CQP):
\begin{align} \label{eq:cqpp}
    \text{Inertial: }&
        \bgamma^* = \argmin_{\bgamma \geq 0}\, \bgamma^\top\left(\bPhi^{k}
        + \Delta t \bDcal_k^\top\left(\bUcal^{k} + \Delta t M^{-1}_k\bFcal_{\text{ext}}^k\right)\right) \nonumber\\
        &\qquad+ \frac{\Delta t^2}{2} \bgamma^\top\bDcal_k^\top M^{-1}_k \bDcal^k\bgamma, \\
    \text{Overdamped: }&
        \bgamma^* = \argmin_{\bgamma \geq 0}\, \bgamma^\top\left(\bPhi^{k}
        + \Delta t\bDcal_k^\top \bMcal^k \bFcal_{\text{ext}}^k \right)  \nonumber \\
        &\qquad+ \frac{\Delta t}{2} \bgamma^\top\bDcal_k^\top\bMcal^k \bDcal^k\bgamma,
\end{align}
which admits efficient solution by a variety of modern optimization algorithms, including projected gradient, interior-point, and active-set methods \cite{pospisil_thesis_2015}. The equivalence between LCP, CQP, and the original DVI, along with conditions for existence and uniqueness of the solution $\bgamma^*$, is summarized in Section \ref{sec:lcp_ccqpp}.

The linearized constraint velocity  $\dot{\bPhi}^k = \bDcal_k^\top\bUcal^k$, while compact in its generalized form, admits a straightforward physical interpretation when expressed componentwise. For a constraint \(\alpha\) defined between surface points $\by_a^{\alpha}, \by_b^{\alpha}$ on bodies $a$ and $b$, with corresponding surface normals $\hat{\bn}_a^\alpha = -\hat{\bn}_b^\alpha$, the linearized separation rate is given by 
\begin{equation}
\dot{\Phi}^k_{\alpha} = -\dot{\by}_a^{\alpha,k} \cdot \hat{\bn}_a^{\alpha,k} - \dot{\by}_b^{\alpha,k} \cdot \hat{\bn}_b^{\alpha,k}.
\end{equation}
This expression assumes that the relative translational motion of the contact points along the current contact line is the only contributor to change in separation. All higher-order effects, such as tangential translation along the contact plane, relative rotation about the contact points, or change in contact locations, are neglected. Herein, such motion is termed \emph{orthogonal} motion, as it lies within the nullspace of $\bDcal_k^\top$.

\subsubsection{Surface representation, proxy surfaces, and the choice of constraint location}
\begin{figure}[htb]
\centering
\includegraphics[width=0.5\textwidth]{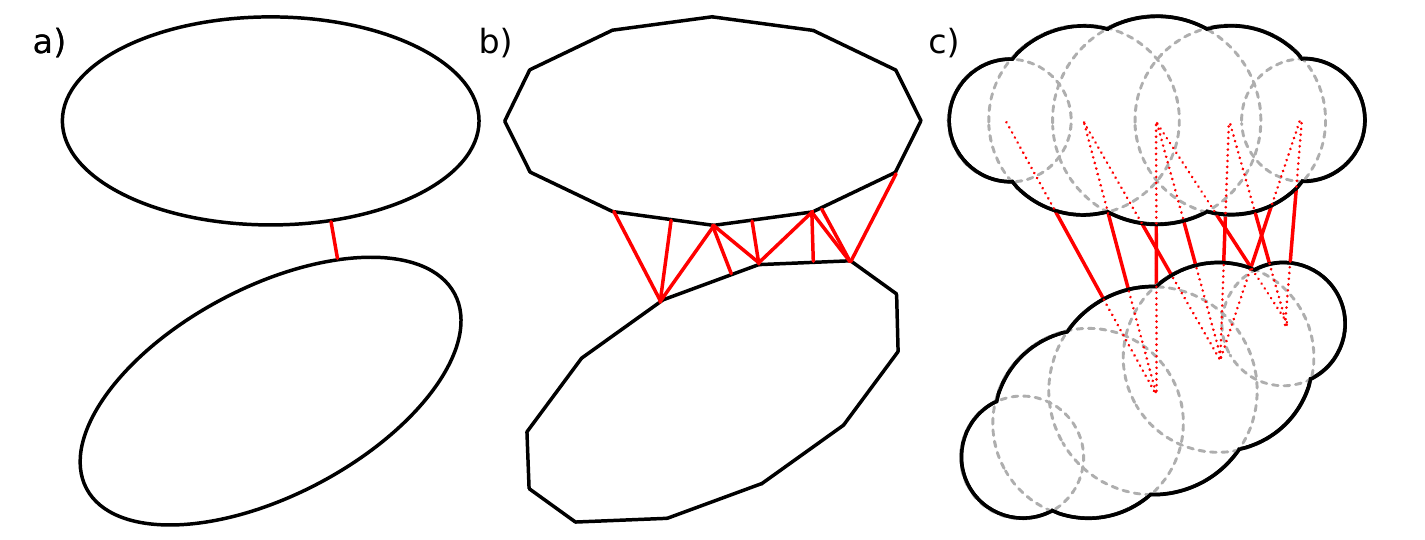}
\caption{
Two-dimensional schematic illustration of collision constraints (solid red lines) generated by three representative contact constructions for smooth bodies. 
\textbf{(a)}~Single-constraint shared-normal signed-distance method, which introduces one constraint between the pair of surface points minimizing the shared-normal signed separation distance. 
\textbf{(b)}~Discrete surface tesselation, which generates constraints between nearby nodes and edges on opposing surfaces. 
\textbf{(c)}~Multi-sphere representation, which places constraints between neighboring sub-spheres.
}
\label{fig:surface_representation}
\end{figure}
While linearizing the collision dynamics provides a mathematically and numerically tractable framework, it carries practical implications whose severity depends on the choice of \emph{proxy surface}—the geometric model used to approximate a body's true boundary for the purposes of constraint generation. Fig.~\ref{fig:surface_representation} illustrates three categories of proxy surfaces. The \emph{single-constraint shared-normal signed-distance method} illustrated in Fig.~\ref{fig:surface_representation}(a) places a single constraint between pairs of nearby bodies at the points that minimize their shared-normal signed separation distance. It is an appealing approach because it utilizes only one constraint for each interacting pair and can effectively prevent overlap, even for non-convex geometries—provided the system evolves linearly, meaning Eq.~\eqref{eq:linear_sep} holds exactly. This approach has been applied to large-scale systems containing spherical \cite{anitescu_tasora_2010a}, ellipsoidal \cite{arman_demp_vs_demc_2017}, and spherocylindrical bodies \cite{alens_2022, yan_lcp_col_stress_2019, palmer_dry_defects_2022}. Importantly, the single-constraint method can remain effective in certain nonlinear regimes as well. For instance, in collisions between spherical bodies, geometric symmetry ensures that orthogonal motions (those not captured by the constraint) cannot lead to overlap. However, for bodies that break this symmetry\textemdash particularly those that are highly elongated, possess complex geometries, or exhibit set-valued contacts\textemdash unconstrained motions are likely to induce overlaps. The severity of these overlaps is influenced by the timestep size, which may need to be prohibitively small to maintain acceptable overlap levels (as demonstrated quantitatively in Section~\ref{sec:examples}).

Contrary to recent claims, constraining orthogonal motions does not require switching to a nonlinear complementarity solver. Instead, one can simply add multiple linearized no-overlap constraints between the bodies, ensuring that orthogonal motion about one constraint induces motions governed by a different linear constraint. As illustrated in Fig.~\ref{fig:surface_representation}, there are two common approaches for generating these constraints: (panel b) tessellating object surfaces into discrete surface elements or (panel c) approximating object volumes as discrete unions of simple primitives such as overlapping spheres or capsules. These techniques are widely used in large-scale simulations involving dense, irregular bodies \cite{anitescu_tasora_2010a, tasora_matrix_free_2011, heyn_anitescu_negrut_kryloc_2013}. They are often combined with single-constraint representations for simpler shapes (e.g., spheres, ellipsoids, spherocylinders) within unified simulation infrastructures such as Project Chrono \cite{mazhar_apgd_2015, chrono_2016}. Both multi-sphere and surface tessellation function similarly: by discretizing the bodies into sub-elements, constraints can be established between pairs of sub-elements on opposing objects, allowing the location and maximum number of constraints to be known \emph{a priori}. No-overlap is then enforced to the resolution dictated by the discretization; the convergence of this approximation toward smooth-surface dynamics with increasing sub-element count is analyzed in Section~\ref{sec:examples}. Herein lies the central scalability challenge of discrete constraint generation: finer discretizations better constrain orthogonal motion, but the number of constraints—and thus computational cost—often grows super-linearly with surface resolution.

\subsection{Properties of linear complementarity and convex quadratic programs}\label{sec:lcp_ccqpp}

This subsection summarizes the relevant mathematical properties of linear complementarity problems (LCPs) and their equivalent convex quadratic programs (CQPs), providing their mathematical definition and the conditions for which existence and uniqueness of their solution can be proven. This presentation follows the lucid works of Lucas Pospíšil \cite{pospisil_thesis_2015} and Cottle-Pang-Stone \cite{CPS1992} and forms the foundation for our subsequent analysis. To start, a convex quadratic program is defined.
\begin{boxeddefinition}[Convex quadratic program] \label{def::ccqp}
    Let $\bx^* \in \mathbb{R}^N$ denote the solution to the constrained optimization problem
    \begin{equation}\label{eq:ccqp}
    \bx^* = \arg \min_{\bx \in \Omega} f(\bx).
    \end{equation}
    This problem is termed a \emph{quadratic program} if the objective function $f$ takes the form
    \begin{equation}
    f(\bx) = \frac{1}{2} \bx^\top \bAcal \bx - \bx^\top \bb,
    \end{equation}
    where $\bAcal \in \mathbb{R}^{N \times N}$ is symmetric and $\bb \in \mathbb{R}^N$. The function $f$ is \emph{convex} if its Hessian $\nabla^2 f(\bx)$ is symmetric positive semi-definite or symmetric positive definite. For quadratic $f$, the gradient and Hessian are given by
    \begin{equation}
    \nabla f(\bx) = \bAcal \bx - \bb, \quad \nabla^2 f(\bx) = \bAcal.
    \end{equation}
    The feasible set $\Omega \subset \mathbb{R}^N$ is convex if for all $x, y \in \Omega$ and all $\alpha \in [0, 1]$, the point $\alpha x + (1 - \alpha) y$ also belongs to $\Omega$, i.e.,
    \begin{equation}
    \forall x, y \in \Omega, \forall \alpha \in [0, 1] : \alpha x + (1 - \alpha) y \in \Omega.
    \end{equation}
    A \emph{convex quadratic program} (CQP) is thus defined as an optimization problem over a closed convex feasible set $\Omega$ with a convex quadratic objective function $f$.
\end{boxeddefinition}
The following classical results characterize the structure and solution conditions of convex quadratic programs, both in unconstrained and constrained forms.
\begin{boxedlemma} (Linear system equivalency) \label{lemma::lin_equiv}
    $\bx^*\in \mathbb{R}^N$ is a solution to the convex quadratic programming problem 
    \begin{equation}\label{eq:unconstrained_min}
    \bx^* = \arg \min_{\bx \in \mathbb{R}^N} f(\bx),
    \end{equation}
    where $f(\bx) = \frac{1}{2}\bx^\top\bAcal\bx - \bx^\top\bb$, if and only if $\bx^*$ is also a solution of the system of linear equations 
    \begin{equation}\label{eq:linear_system_equiv}
    \bAcal \bx = \bb.
    \end{equation}
\end{boxedlemma}
\begin{proof} 
    See Pospíšil \cite[Lem.~1.3.2]{pospisil_thesis_2015}.
\end{proof}
More generally, the solution to the CQP corresponds to a variational inequality over the feasible set.
\begin{boxedlemma} (Variational inequality equivalency) \label{lemma::variational_equiv}
    $\bx^*\in \Omega$ is a solution to the convex quadratic program 
    \begin{equation}\label{eq:constrained_min}
    \bx^* = \arg \min_{\bx \in \Omega} f(\bx),
    \end{equation}
    where $f(\bx)$ is a differentiable convex function and $\Omega$ is the closed convex feasible set, if and only if 
    \begin{equation}\label{eq:variational_ineq}
    \forall \bx \in \Omega : (\bx - \bx^*)^\top \nabla f(\bx^*) \geq 0.
    \end{equation}
\end{boxedlemma}
\begin{proof} 
    See Boyd and Vandenberghe \cite[Section.~4.2.3]{boyd_convex_optimization_2004}.
\end{proof}
Linear complementarity problems (LCPs) can be viewed as a special case of convex quadratic programs. This connection is made precise in the following classical result.
\begin{boxedlemma}[Equivalence of monotone LCP and CQP]
    \label{lemma::lcp_to_ccqp}
    Let $\bAcal\in\mathbb{R}^{N\times N}$ be symmetric positive-semidefinite and
    $\bb\in\mathbb{R}^N$.  The linear complementarity problem
    \begin{equation}
    \boldsymbol{0}\le \bAcal\bx-\bb \perp \bx\ge\boldsymbol{0},
    \end{equation}
    which can be broken into three conditions
    \begin{equation}
        \bx \geq 0, \quad
        \bg(\bx) = \bAcal \bx - \bb \geq 0, \quad
        \bx^\top \bg(\bx) = 0,
    \end{equation}
    is equivalent to the convex quadratic program
    \begin{equation}
    \bx^*=\arg\min_{\bx\ge 0}
    \frac12\bx^\top\bAcal\bx-\bb^\top\bx,
    \quad
    \text{with } \nabla f(\bx)=\bAcal\bx-\bb.
    \end{equation}
\end{boxedlemma}
\begin{proof}
    See Cottle–Pang–Stone \cite[Section 1.4]{CPS1992} or Pospíšil \cite[Lem.~3.3.1]{pospisil_thesis_2015}.
\end{proof}
Attention now turns to conditions ensuring the existence of a minimizer for a convex quadratic objective.
\begin{boxedlemma} (Solvability of monotone LCPs) \label{lemma::lcp_existence}
    Consider the linear complementarity problem
    \begin{equation}
        0 \le \bAcal \bx - \bb \perp \bx \ge 0,
    \end{equation}
    where $\bAcal \in \mathbb{R}^{N\times N}$ is symmetric. This problem always has a solution if either of the following conditions are met:
    \begin{enumerate}[wide=0pt]
        \item $\bAcal$ is symmetric positive definite
        \item $\bAcal$ is symmetric positive semi-definite and $\by^\top\bb \leq 0\, \forall \by\geq 0$ for which $\by \in \Ker \bAcal$.
    \end{enumerate}
\end{boxedlemma}
\begin{proof}
    \begin{enumerate}[wide=0pt]
        \item \textbf{Condition 1:} See Cottle–Pang–Stone \cite[Thm 3.1.6]{CPS1992}.
        \item \textbf{Condition 2:} Rewrite $0 \le \bAcal \bx - \bb \perp \bx \ge 0$ as
        $0 \le \bAcal \bx + \bq \perp \bx \ge 0$ with $\bq = -\bb$.
        An established existence theorem states that the LCP $0 \leq \bA \bx + \bq \perp \bx \geq 0$ with copositive $\bA$ admits a solution if $\by^\top \bq \geq 0$ for all $\by$ that solve the homogeneous LCP $0 \leq \bA \bx \perp \bx \geq 0$ \cite{stewart_frictional_review_2000}. If $\bA$ is symmetric positive semi-definite, then it is copositive. Moreover, if $\by$ solves the homogeneous LCP, then $\by^\top \bA \by = 0$, which for symmetric positive semi-definite $\bA$ implies $\by \in \Ker \bA$. Thus, for the sign convention used here, it suffices that $\by^\top \bb \le 0$ for all $\by \ge 0$ in $\Ker \bA$. In particular, since $\Ker \bA \perp \Im \bA$ when $\bA$ is symmetric, this condition holds whenever $\bb \in \Im \bA$; it also holds whenever $\bb \le 0$ componentwise.
    \end{enumerate}
\end{proof}
Finally, uniqueness is considered. The following classical result gives sufficient conditions under which the solution (or a transformation thereof) is unique.
\begin{boxedlemma} (Uniqueness criteria) \label{lemma::unique}
    Consider the optimization problem 
    \begin{equation}
    \bx^* = \arg \min_{\bx \in \Omega} f(\bx) 
    \end{equation}
    where $f : \mathbb{R}^N \rightarrow \mathbb{R}$ is a convex function that is twice continuously differentiable on $\Omega$ and $\Omega \subset \mathbb{R}^N$ is a non-empty, closed convex feasible set.
    \begin{enumerate}[wide=0pt]
        \item If $f$ has a symmetric positive definite Hessian matrix $\nabla^2f(\bx)\,  \forall \bx\in \Omega$, then \(\bx^*\) is unique.
        \item If $f$ has a symmetric positive semi-definite Hessian matrix
        \begin{equation}
            \nabla^2f(\bx) = \bDcal^\top \bMcal \bDcal
            \qquad \forall \bx\in \Omega,
        \end{equation}
        where $\bDcal\in \mathbb{R}^{r\times N}$ is constant and $\bMcal\in \mathbb{R}^{r\times r}$ is symmetric positive definite, then $\bDcal \bx^*$ is unique. 
    \end{enumerate}
\end{boxedlemma}
\begin{proof}
\begin{enumerate}[wide=0pt]
\item \textbf{Condition 1:} This is classical; see Cottle–Pang–Stone \cite[Thm 3.1.6]{CPS1992}.

\item \textbf{Condition 2:} This is a straightforward extension that seems less often stated in this general form. By Lemma \ref{lemma::variational_equiv}, any two minimizers $\bx_1^{*},\bx_2^{*}$ of $f$ on~$\Omega$ must satisfy the variational inequality conditions
\begin{equation}
    (\bx - \bx_1^*)^\top \nabla f(\bx_1^*) \geq 0, \text{ and } (\bx - \bx_2^*)^\top \nabla f(\bx_2^*) \geq 0, \forall \bx \in \Omega. 
\end{equation}
Setting \( \bx = \bx_2^* \) in the first inequality and \( \bx = \bx_1^* \) in the second inequality, rearranging, and subtracting the inequalities yields
\begin{equation}
    (\bx_2^* - \bx_1^*)^\top \left(\nabla f(\bx_2^*) - \nabla f(\bx_1^*)\right) \leq 0.
\end{equation}
Let $\Delta\bx \coloneqq \bx_2^* - \bx_1^*$. Since $f$ is twice continuously differentiable, the gradient difference admits the integral representation
\begin{equation}
    \nabla f(\bx_2^*) - \nabla f(\bx_1^*)
    = \int_0^1 \nabla^2 f\left(\bx_1^* + t\Delta\bx\right)\Delta\bx\, dt.
\end{equation}
Substituting $\nabla^2f(\bx)=\bDcal^\top\bMcal\bDcal$ and applying the preceding inequality gives
\begin{equation}
    0 \geq \int_0^1 \left(\bDcal\Delta\bx\right)^\top
    \bMcal
    \left(\bDcal\Delta\bx\right) \, dt.
\end{equation}
The integrand is non-negative for every $t\in[0,1]$ because $\bMcal$ is symmetric positive definite. Therefore the integral can vanish only if the integrand is identically zero. Writing $\by \coloneqq \bDcal\Delta\bx$ gives
\begin{equation}
0=\by^{T}\bMcal\by
\, \forall t\in[0,1]
   \,\Longrightarrow\,
\by=0
   \,\Longrightarrow\,
\bDcal\bx_1^*=\bDcal\bx_2^*.
\end{equation}
So while the minimizer itself need not be unique, its image under \(\bDcal\) is unique.
\end{enumerate}
The construction above depends only on the chosen compact regularity ball and on the fixed prefix length. In particular, if one starts from any prescribed compact ball $K^\sharp$ contained in the common regularity neighborhood of $\bCcal^k$, the same finite-prefix construction yields a threshold $\Delta t_{\mathrm{reg}}(K^\sharp,c)>0$ that keeps the first $c$ iterates in $K^\sharp$.
\end{proof}

\section{Smooth-surface contact dynamics by recursive constraint augmentation} \label{sec:augment}
This section reformulates complementarity-based time-stepping for smooth frictionless contact without pre-enumerating a proxy-surface contact set. This construction, originally presented without proof in the supplement of~\cite{weady_proliferating_collectives_2024}, asks whether one can perform LCP-based time-stepping directly on smooth surfaces without suffering the numerical instabilities of single-constraint formulations or the oversampling and poor scalability of discrete surface proxies, by replacing a fixed contact set with recursive constraint generation driven by the trial update itself.

The restriction to frictionless contact is deliberate: as a first step toward a broader formulation, it isolates the geometric difficulty introduced by proxy-surface representations and provides the analytical foundations on which frictional extensions can later build. Although the reformulation is presented here constructively, its structure was arrived at through sustained iteration between physical intuition and mathematical proof, searching for conditions under which an overlap-free update exists and the induced center-of-mass dynamics are unique. Earlier, more na\"ive strategies---including schemes that discard or re-linearize prior constraints at each iteration---failed to furnish these properties because they lose the monotone structure needed below. The retained-constraint construction developed here is one for which finite termination, existence of an overlap-free update to prescribed tolerance, uniqueness of the induced center-of-mass dynamics, and consistency with the continuous-time limit can be established under the assumptions stated in Section~\ref{sec:relcp}. Section~\ref{subsec:intuitive} gives the time-stepping procedure; Section~\ref{sec:relcp} abstracts the same recursive structure into a general complementarity object through which its mathematical properties can be studied.

\subsection{Recursive constraint augmentation for smooth-surface contact}\label{subsec:intuitive}

The construction replaces geometric oversampling by constraint augmentation: starting from a single-constraint shared-normal signed-distance LCP between the smooth surfaces, one solves for the trial configuration, identifies any residual overlaps, and adds constraints only for the violated pairs while retaining the effect of constraints introduced in earlier iterations. If the system were linear, this procedure would converge immediately. For smooth nonlinear geometry, the additional constraints should instead be introduced only as needed to suppress the dominant unconstrained modes of relative motion. The construction must therefore specify two operations:
\begin{enumerate}[wide=0pt]
    \item the retention and incorporation of constraints introduced in previous iterations;
    \item the physically consistent linearization of newly generated constraints.
\end{enumerate}
The remainder of this section develops these two operations explicitly.

The procedure begins with the classical single-constraint shared-normal signed-distance LCP: for each near-contacting pair, a single constraint is introduced between the surface points that minimize the shared-normal signed separation distance. After this LCP is solved, the resulting trial configuration is inspected for residual interpenetration. For each pair with unresolved overlap, a new shared-normal signed-distance constraint is generated and linearized about the trial configuration. The new constraints are then ``pulled back'' to the original configuration, while all previous constraints are retained in their existing linearized form. The updated LCP therefore contains both old and new constraints, and the recursion continues until no overlap exceeds the prescribed geometric tolerance.

\begin{figure}[htb]
\centering
\includegraphics[width=0.5\textwidth]{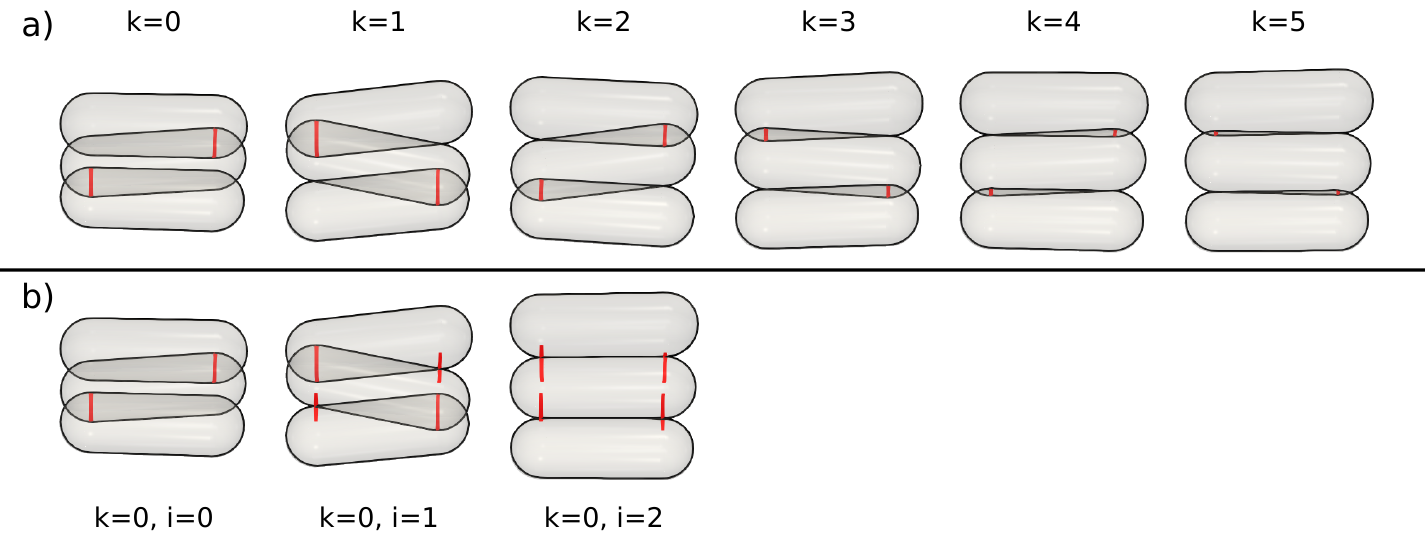}
\caption{
Toy problem with three initially overlapping capsules; active collision constraints are shown in red. 
Top row: the single-constraint SNSD formulation leaves orthogonal relative motion unconstrained, producing temporal jitter. 
Bottom row: successive recursive augmentation iterations within one timestep, showing how the contact set is augmented to resolve orthogonal motion.
}
\label{fig:jitter}
\end{figure}
Figure~\ref{fig:jitter} illustrates this constraint-generation procedure. In the top row, a single-constraint LCP controls only the most immediate component of interpenetration within the timestep, leaving orthogonal modes of motion unconstrained and producing the well-known discretization artifact referred to as \emph{jitter}. In contrast, the bottom row shows successive recursive augmentation iterations within a single timestep: each new constraint targets a previously unconstrained motion, while all prior constraints are retained. Discarding earlier constraints would reintroduce the same instability observed with single-constraint LCPs, manifesting per iteration rather than per timestep. This observation also clarifies why nonsmooth Newton or fixed-point iterations based on a single adaptively placed constraint can become unstable: the iteration may repeatedly move the constraint rather than accumulate the directions needed to suppress overlap.

The first and second overdamped iterations are written explicitly before the general recursion is stated. We begin with notation for the growing constraint space.

\subsubsection{Notation for a growing constraint space}

Each recursion consists of three phases: constraint generation, solution of an LCP, and a discrete-time configuration update. Let $\mathcal{A}_n$ denote the set of $N_C^n$ constraints at recursion iteration $n$, and define the increment $\Delta \mathcal{A}_{n} = \mathcal{A}_{n} \setminus \mathcal{A}_{n-1},$ which contains the $\Delta N_C^n = N_C^n - N_C^{n-1}$ new constraints generated that iteration. Without loss of generality, constraint indices are ordered such that each new block of constraints is appended to the end. The notation $\left.\bgamma\right|_{\mathcal{B}}$ denotes the restriction (or slice) of a vector $\bgamma$ to the subindices associated with constraint subset $\mathcal{B}$. Quantities decorated with a tilde (e.g., $\tilde{\bPhi}$) denote augmented vectors that concatenate information from previously existing and newly generated constraints.

Because the constraint space grows with each recursion iteration, several linear maps between spaces of different dimension are needed throughout. For any $p, q \in \mathbb{N}$ with $q > p$, define:
\begin{enumerate}[wide=0pt]
    \item $\iota_q$ is the canonical immersion into $\mathbb{R}^q$, taking any $\bv\in\mathbb{R}^p$ to $\mathbb{R}^q$ by appending $q-p$ zeros:
    \begin{equation}\label{eq:canonical_immersion}
    \iota_q\!(\mathbf{v}) = (v_1, v_2, \ldots, v_p, \underbrace{0, 0, \ldots, 0}_{q-p \text{ zeros}}).
    \end{equation}
    \item $\pi_p$ is the canonical projection onto $\mathbb{R}^p$, taking any $\bw\in\mathbb{R}^q$ to $\mathbb{R}^p$ by removing the last $q-p$ values:
    \begin{equation}\label{eq:canonical_projection}
    \pi_p(\mathbf{w}) = (w_1, w_2, \ldots, w_p).
    \end{equation} 
    \item $\bP_{i:j}:\mathbb{R}^p\rightarrow\mathbb{R}^p$ is an orthogonal window projection that takes $\bv\in\mathbb{R}^p$ and sets $v_\ell = 0\,  \forall \ell \notin [i, i+1, \dots, j]$. 
\end{enumerate}

\subsubsection{Iteration 0: the initial single-constraint SNSD LCP}

The zeroth iteration is simply the standard single-constraint SNSD LCP. For each near-contacting pair, a single constraint is placed at the pair of surface points that minimize the shared-normal signed separation distance, yielding a constraint set $\mathcal{A}_0$ of size $N_C^0$. The corresponding LCP updates the configuration from $\bCcal^k\in \mathbb{R}^{7N}$ to a trial configuration $\bCcal_1^{k+1}\in \mathbb{R}^{7N}$ and the true signed separation $\bPhi_0^k\in \mathbb{R}^{N_C^0}$ to a \emph{linearized} separation vector $\bPhi_1^{k+1}\in \mathbb{R}^{N_C^0}$:
\begin{equation}
    \begin{aligned}
    \bPhi_1^{k+1} &= \bPhi_0^{k} + \Delta t \bDcal_k^\top \bMcal^k \left(\bFcal_{\text{ext}}^k + \bDcal^k \bgamma_{0}^{k,*}\right),\\
    &\, 0\leq \bPhi_1^{k+1} \perp \bgamma_{0}^{k,*} \geq 0,
    \end{aligned}
\end{equation}
with trial configuration $\bCcal_1^{k+1} = \bCcal^{k} + \Delta t  \bGcal^k \bMcal^k\left(\bFcal_{\text{ext}}^k + \bDcal^k  \bgamma_{0}^{k,*}\right)$.

The trial configuration $\bCcal_1^{k+1}$ is accepted contingently: if no pair of bodies overlaps beyond a prescribed tolerance $\epsilon_{NCP}$, then $\bCcal_1^{k+1}$ is accepted as the discrete-time update for this timestep and the recursion terminates. Otherwise, new constraints are generated at updated points of minimum signed separation (with shared normals) between pairs that violate the threshold, the trial configuration is rejected, and a new LCP is formed and solved. Here, external forces and velocities appear only in the zeroth LCP, as their effect is already encoded in $\bCcal_1^{k+1}$.

\subsubsection{Iteration 1: augmenting the constraint set}\label{subsubsec:iteration1}

Assuming residual overlaps persist at $\bCcal_1^{k+1}$, the procedure must (a) carry forward the existing constraints and (b) incorporate freshly generated constraints in a manner consistent with the Riemann-sum-based time integration. The construction separates these two operations as follows.

\paragraph{Retaining previous constraints}
Previously existing constraints are not rebuilt or re-linearized; instead, they are carried forward exactly in the linearized form already used to produce the current trial configuration. For the original constraints $\alpha\in\mathcal{A}_0$, one keeps the unilateral model
\begin{equation}
\Psi_{0,\alpha}(\bCcal^k) \coloneqq \Phi_\alpha^k + \Delta t\,\dot{\Phi}_\alpha^k,
\end{equation}
whose evaluated linearized separations after the zeroth solve are the entries of $\bPhi_1^{k+1}$. For the newly generated constraints $\alpha\in\Delta\mathcal A_1$, no such linearized history yet exists, so the only information available at this stage is their true signed separation at the trial configuration,
\begin{equation}
\Phi_{1,\alpha}^{k+1} \coloneqq \Phi_\alpha\left(\bCcal_1^{k+1}\right).
\end{equation}
The augmented constraint vector is therefore formed by concatenating the already linearized separations from the old constraints with the freshly evaluated separations of the new constraints, $\left.\bPhi(\bCcal_1^{k+1})\right|_{\Delta \mathcal{A}_1}\in \mathbb{R}^{\Delta N_C^1}$:
\begin{align}
    \bPhi_1^{k+1} &=\bPhi_0^{k} + \Delta t \bDcal_k^\top \bMcal^k \left(\bFcal_{\text{ext}}^k + \bDcal^k \bgamma_{0}^{k,*}\right) \in \mathbb{R}^{N_C^0}, \\ 
    \tilde{\bPhi}_1^{k+1} &=
    \begin{bmatrix}
    \bPhi_1^{k+1} \\
    \left.\bPhi\left(\bCcal_1^{k+1}\right)\right|_{\Delta \mathcal{A}_1}
    \end{bmatrix}
    \in \mathbb{R}^{N_C^1}.
\end{align}
Thus the influence of the old constraints is preserved exactly through $\bPhi_1^{k+1}$: it already contains the full effect of the zeroth-iteration solve, so no re-evaluation or re-linearization of those constraints is performed.

\paragraph{Linearizing and pulling back new constraints}
The old constraints require no further modification: their force law is already fixed by the linearized unilateral model retained from the previous solve. In particular, for $\alpha\in\mathcal A_0$ one already has
$
\Psi_{0,\alpha}(\bCcal^k) \coloneqq \Phi_\alpha^k + \Delta t\,\dot{\Phi}_\alpha^k,
$
for which D'Alembert's principle gives
$
\bFcal^k = \bDcal^k \bgamma^k,
$
for $\left(\bDcal^k\right)^\top \coloneqq \nabla_{\bCcal^k} \bPhi^k\bGcal^k$. The only remaining task, then, is to define an analogous first-order model for the newly generated constraints. The difficulty is that for a new shared-normal signed-distance constraint, the signed-separation function and its gradient are only known analytically at the trial configuration $\bCcal_1^{k+1}$ where that constraint was discovered. Accordingly, for a newly added constraint one works with
\begin{equation}
\Psi_{1,\alpha}(\bCcal^k) \coloneqq \Phi_{1,\alpha}^{k+1} - \Delta t\,\dot{\Phi}_{1,\alpha}^{k} + \Delta t\,\dot{\Phi}_{\alpha}^{k}
\end{equation}
The first two terms ``pull back'' the separation distance in the trial configuration to $\bCcal^k$, and the final term pushes the system forward to the next trial configuration. Here $\Psi_{1,\alpha}$ should be understood as a \emph{first-order pulled-back model} for the new constraint, not as an exact constraint function whose full derivative is retained. Indeed, differentiating the exact pulled-back quantity would introduce higher-order terms from $\nabla_{\bCcal}\dot{\Phi}_{\alpha}$, including Hessian terms of $\Phi_{\alpha}$, changes in the contact-point map and contact normals, and variation of $\bGcal(\bCcal)$. Consistent with the linearized update in Eq.~\eqref{eq:linear_sep}, all such higher-order geometric terms are neglected and the contact geometry is frozen at the configuration where the new constraint was discovered; a constraint treated in this way is called a \emph{frozen-normal constraint}. Thus, to first order,
\begin{equation}
\nabla_{\bCcal^k} \Psi_{1,\alpha}
\approx
\nabla_{\bCcal_1^{k+1}} \Phi_{1,\alpha}^{k+1},
\end{equation}
and the D'Alembert force map for the augmented constraint set is taken to be:
\begin{align}
    \left( \bDcal_{0}^{k+1} \right)^\top &=
    \nabla_{\bCcal^{k}} \bPhi^{k}\bGcal^k \in \mathbb{R}^{N_C^0\times 6N}, \\
    \left( \bDcal_{1}^{k+1} \right)^\top &=
    \begin{bmatrix}
    \nabla_{\bCcal^{k}} \bPhi^{k}\bGcal^k \\
    \nabla_{\bCcal_{1}^{k+1}} \left.\bPhi\!\left(\bCcal_{1}^{k+1}\right)\right|_{\Delta \mathcal{A}_1} \bGcal^k
    \end{bmatrix}\in \mathbb{R}^{ N_C^1 \times 6N}.
\end{align}

\paragraph{The updated LCP}
With the augmented constraint vector and Jacobian in hand, the updated LCP is:
\begin{equation}
    \begin{aligned}
    \bPhi_2^{k+1} = \tilde{\bPhi}_1^{k+1} 
    &+ \Delta t \left(\bDcal_{1}^{k+1}\right)^\top \bMcal^k 
        \bDcal_{1}^{k+1} \bgamma_{1}^{k,*} \\
    &- \Delta t \left(\bDcal_{1}^{k+1}\right)^\top \bMcal^k \bDcal_{0}^{k+1} 
        \bgamma_{0}^{k,*}, \\
    \, 0\leq \bPhi_2^{k+1} &\perp \bgamma_{1}^{k,*} \geq 0.
    \end{aligned}
\end{equation}
The first term carries forward the augmented separations. The second applies the new constraint forces through the augmented Jacobian. The third subtracts the effect of the zeroth-iteration forces---already encoded in $\tilde{\bPhi}_1^{k+1}$---so that the new solve adjusts (rather than double-counts) their contribution.

\subsubsection{General iteration $n$: overdamped and inertial recursive updates}

The three-phase pattern of iteration~1---solve, inspect, augment---extends directly to arbitrary recursion iteration $n$, yielding the full overdamped recursive update:
\begin{equation}\label{eq:relcp}
    \begin{aligned}
    \bPhi_{n+1}^{k+1} &= \tilde{\bPhi}_n^{k+1} + \Delta t \left( \bDcal_n^{k+1} \right)^\top \bMcal^k \bDcal_n^{k+1} \left( \bgamma_{n}^{k,*} - \iota_{N_C^n}\!\left(\bgamma_{n-1}^{k,*}\right)\right), \\
    &\, 0\leq \bPhi_{n+1}^{k+1}\perp \bgamma_{n}^{k,*} \geq 0,
    \end{aligned}
\end{equation}
and the trial configuration is given by 
$
\bCcal_{n+1}^{k+1} = \bCcal_n^{k+1} + \Delta t  \bGcal^k \bMcal^k \bDcal_n^{k+1} \left(\bgamma_{n}^{k,*} - \iota_{N_C^n}\!\left(\bgamma_{n-1}^{k,*}\right) \right).
$
For the zeroth iteration, 
$\bDcal_0^{k+1} = \bDcal^{k}$ and $
\tilde{\bPhi}_0^{k+1} = \bPhi_0^k + \Delta t\bDcal_{k}^\top \bMcal^k\bFcal_{\text{ext}}^k. 
$
The complete contact-recursion procedure is summarized in Algorithm~\ref{alg:relcp}.

\begin{algorithm}[htb]
\caption{One overdamped recursive augmentation timestep.}
\label{alg:relcp}
\begin{algorithmic}[1]
\Require Configuration $\bCcal^k$, external forces $\bFcal_{\text{ext}}^k$, mobility $\bMcal^k$, tolerance $\epsilon_{NCP}$
\Ensure Updated configuration $\bCcal^{k+1}$
\State \textbf{Initialize:} Generate shared-normal signed distance constraints $\mathcal{A}_0$ between near-contacting pairs; set $n \gets 0$
\State Compute $\bPhi_0^k$, set $\bDcal_0^{k+1} \gets \bDcal^k$ and $\tilde{\bPhi}_0^{k+1} \gets \bPhi_0^k + \Delta t\bDcal_k^\top \bMcal^k \bFcal_{\text{ext}}^k$
\State Solve for $\bgamma_0^{k,*}$ such that $0\leq \bPhi_1^{k+1} \perp \bgamma_0^{k,*}\geq 0$
\Statex \hspace{\algorithmicindent}$\bPhi_1^{k+1} \coloneqq \tilde{\bPhi}_0^{k+1} + \Delta t \left(\bDcal_0^{k+1}\right)^\top\bMcal^k\bDcal_0^{k+1}\bgamma_0^{k,*}$
\State $\bCcal_1^{k+1} \gets \bCcal^k + \Delta t \bGcal^k\bMcal^k\left(\bFcal_{\text{ext}}^k + \bDcal^k\bgamma_0^{k,*}\right)$
\While{$\exists$ body pair with overlap $> \epsilon_{NCP}$ at $\bCcal_{n+1}^{k+1}$}
    \State $n \gets n + 1$
    \State Generate new shared-normal signed-distance constraints $\Delta\mathcal{A}_n$ from $\bCcal_n^{k+1}$ at overlapping pairs; $\mathcal{A}_n \gets \mathcal{A}_{n-1}\cup\Delta\mathcal{A}_n$
    \State $\tilde{\bPhi}_n^{k+1} \gets \text{append}\left(\bPhi_n^{k+1}, \left.\bPhi\left(\bCcal_n^{k+1}\right)\right|_{\Delta\mathcal{A}_n}\right)$
    \State $\bDcal_n^{k+1} \gets \text{append}\left(\bDcal_{n-1}^{k+1}, \left(\nabla_{\bCcal_{n}^{k+1}} \left.\bPhi\!\left(\bCcal_{n}^{k+1}\right)\right|_{\Delta \mathcal{A}_n} \bGcal^k\right)^\top \right)$
    \State Solve for $\bgamma_n^{k,*}$ such that $0\leq \bPhi_{n+1}^{k+1} \perp \bgamma_n^{k,*}\geq 0$
    \Statex \hspace{\algorithmicindent}$\bPhi_{n+1}^{k+1} \coloneqq \tilde{\bPhi}_n^{k+1} + \Delta t (\bDcal_n^{k+1})^\top\bMcal^k\bDcal_n^{k+1}(\bgamma_n^{k,*} - \iota_{N_C^n}\!(\bgamma_{n-1}^{k,*})) $
    \State $\bCcal_{n+1}^{k+1} \gets \bCcal_n^{k+1} + \Delta t\bGcal^k\bMcal^k\bDcal_n^{k+1}(\bgamma_n^{k,*} - \iota_{N_C^n}\!(\bgamma_{n-1}^{k,*}))$
\EndWhile
\State \Return $\bCcal^{k+1} \gets \bCcal_{n+1}^{k+1}$
\end{algorithmic}
\end{algorithm}

The inertial version takes the same form with $\Delta t \to \Delta t^2$ in the quadratic term:
\begin{equation}\label{eq:relcp_inertial}
    \begin{aligned}
    \bPhi_{n+1}^{k+1} &= \tilde{\bPhi}_n^{k+1} + \Delta t^2 \left( \bDcal_n^{k+1} \right)^\top Mcal_k^{-1} \bDcal_n^{k+1} \left( \bgamma_{n}^{k,*} - \iota_{N_C^n}\!\left(\bgamma_{n-1}^{k,*}\right)\right), \\
    &\quad\text{s.t. } 0\leq \bPhi_{n+1}^{k+1} \perp \bgamma_{n}^{k,*} \geq 0,
    \end{aligned}
\end{equation}
with the zeroth iteration instead using 
\begin{equation}
\tilde{\bPhi}_0^{k+1} = \bPhi_0^k + 
\Delta t \bDcal_{k}^\top\left(\bUcal^{k} + \Delta t M^{-1}_k\bFcal_{\text{ext}}^k\right).
\end{equation}

\section{ReLCP: A recursively generated linear complementarity problem}\label{sec:relcp}

The contact recursion above repeatedly solves LCPs whose dimension increases when the previous solution exposes additional constraints. Abstracting this structure from smooth-surface contact yields a complementarity problem with a recursively generated constraint space. We refer to this class as a \emph{recursively generated linear complementarity problem} (ReLCP). This section defines the ReLCP class, establishes sufficient conditions for solvability, finite termination, and uniqueness, and then verifies that the contact recursion of Section~\ref{sec:augment} is an instance of this class and the conditions under which it satisfies the ReLCP properties.

\subsection{Properties of the recursively generated linear complementarity problem} \label{subsubsec:relcp_properties}

In a ReLCP, each recursion augments the complementarity system itself: the unknown, the Hessian matrix, and the right-hand side all increase in dimension as new constraints are generated. This differs from a sequential linear complementarity problem (SLCP), which repeatedly solves updated LCPs of fixed size. Using the notation introduced in \S\ref{subsec:intuitive}, the ReLCP is defined as follows:
\begin{boxeddefinition} (Recursively generated linear complementarity problem) \label{def::relcp}
    A recursively generated linear complementarity problem (ReLCP) is a fixed point iteration of a standard LCP within which additional complementarity constraints are added to the system with each iteration. Hence, at recursion index $n$, the dimension $N_n$ of the unknown $\bx_n \in \mathbb{R}^{N_n}$\textemdash and therefore the matrix $\bAcal_n\in\mathbb{R}^{N_n \times N_n}$ and vector $\bb_n\in\mathbb{R}^{N_n}$\textemdash increases in size according to the solution at the previous recursion. Because constraints are strictly added and old constraints are not modified, $\bAcal_{n-1}$ is (without loss of generality) the upper-left block of $\bAcal_{n}$. Furthermore, because the solution $\bx_n^*$ to the $n$\textsuperscript{th} linear complementarity problem
    \begin{equation}\label{eq:relcp_lcp}
    0 \leq \bg_n(\bx_n) = \bAcal_n(\bx_{n-1}^*) \bx_n - \bb_n(\bx_{n-1}^*) \perp \bx_n \geq 0
    \end{equation}
    need not be unique (per Lemmas \ref{lemma::lcp_to_ccqp} and \ref{lemma::unique}), one of two sufficient but not necessary conditions should be met for every recursion $n$:
    \begin{enumerate}[wide=0pt]
        \item[Case 1] $\bAcal_n$ is symmetric positive definite, or
        \item[Case 2] $\bAcal_n$ is symmetric positive semi-definite and can be expressed as $\bAcal_n = \bDcal_n^\top\bMcal \bDcal_n$ for some rectangular matrix $\bDcal_n$ and fixed symmetric positive definite matrix $\bMcal$, and $\by^\top\bb_n \leq 0 \,\forall \by\geq 0$ for which $\by \in \Ker \bAcal_n$. In this case, $\bDcal_{n-1}$ is the leftmost block of $\bDcal_n$ and $\bAcal_{n-1}$ is the upper-left block of $\bAcal_{n}$.
    \end{enumerate}
    
    In the first case, the ReLCP can be deemed converged at iteration $c$ if $\|\bx_c^* - \iota_{N_c}\!(\bx_{c-1}^*)\|<\epsilon$ for a chosen positive tolerance $\epsilon$. On the other hand, to accommodate the lack of uniqueness of $\bx_n^*$ in the second case, convergence can be defined as $\|\bDcal_c\bx_c^* - \bDcal_{c-1}\bx_{c-1}^*\|<\epsilon$. 
\end{boxeddefinition}
The structure imposed on the ReLCP ensures that the dimensionality of the system grows monotonically: it increases whenever new constraints are appended and may plateau only once the optimality condition is met, meaning that no new constraints are added. Existence of a solution sequence for this iterative process depends on a boundedness condition and a specific relationship between successive right-hand sides. The following theorem gives sufficient conditions for existence of the sequence and for satisfaction of the convergence criterion in Definition \ref{def::relcp}.
\begin{boxedtheorem} (Solvability and finite termination of the ReLCP) \label{theorem::relcp_conv}
    The recursively generated linear complementarity problem admits a solution $\bx_n^*$ at every recursion if the conditions of Definition \ref{def::relcp} hold. Moreover, assume that $\left(\bx_n^*\right)^\top\bAcal_n(\bx_{n-1}^*) \bx_n^*$ is bounded from above and that there exists an augmented residual $\tilde{\bg}_{n-1}(\bx_{n-1}^*)\in\mathbb{R}^{N_n}$ satisfying $\pi_{N_{n-1}}\left(\tilde{\bg}_{n-1}(\bx_{n-1}^*)\right)=\bg_{n-1}(\bx_{n-1}^*)$ and
    \begin{equation}
        \bb_n(\bx_{n-1}^*) = \bAcal_n(\bx_{n-1}^*)\iota_{N_n}\!(\bx_{n-1}^*) - \tilde{\bg}_{n-1}(\bx_{n-1}^*),
    \end{equation}
    i.e., the iterative linear complementarity problem is Newton-esque:
    \begin{equation}\label{eq:relcp_newton}
    \begin{aligned}
    0 \leq \bg_n(\bx_n) &\perp \bx_n \geq 0, \\
    \bg_n(\bx_n) &\coloneqq \bAcal_{n}\left(\bx_{n-1}^*\right)\left( \bx_n - \iota_{N_n}\!(\bx_{n-1}^*)\right) + \tilde{\bg}_{n-1}(\bx_{n-1}^*).
    \end{aligned}
    \end{equation}
    Then in Case 2 the ReLCP satisfies the convergence criterion of Definition~\ref{def::relcp} after finitely many iterations. In Case 1, the same conclusion holds provided the matrices $\bAcal_n$ are uniformly positive definite, i.e., there exists a constant $\mu>0$ such that
    \begin{equation}
        \bv^\top\bAcal_n\bv \ge \mu\|\bv\|^2
        \qquad \forall \bv\in\mathbb{R}^{N_n},\ \forall n.
    \end{equation}
\end{boxedtheorem}
\begin{proof}
The proof is given in Appendix~\ref{app:proof:theorem_relcp_conv}.
\end{proof}
Because the ReLCP requires solving a series of repeated linear complementarity problems where the setup at recursion $n+1$ entirely depends on the solution at recursion $n$ (and each of these solutions is guaranteed to exist), uniqueness of the relevant converged quantity follows from uniqueness of this series of solutions.
\begin{boxedcorollary} (Uniqueness of the ReLCP) \label{lemma::relcp_uniqueness}
    Assume the hypotheses of Theorem \ref{theorem::relcp_conv}. Then the recursively generated linear complementarity problem has a unique converged result in the following cases:
    \begin{itemize}
        \item If Case 1 of Definition \ref{def::relcp} holds, the matrices $\bAcal_n$ are uniformly positive definite, and $\bAcal_n(\bx_{n-1}^*)$ and $\bb_n(\bx_{n-1}^*)$ are uniquely determined by $\bx_{n-1}^*$, then the converged solution $\bx_c^*$ is unique.
        \item If Case 2 of Definition \ref{def::relcp} holds and there exists an auxiliary state $\bs_n$ such that $\bs_n$ is uniquely determined by the previously unique quantity $\bDcal_{n-1}\bx_{n-1}^*$ and $\bDcal_n$ is uniquely determined by $\bs_n$, then the converged quantity $\bDcal_c\bx_c^*$ is unique.
\end{itemize}
\end{boxedcorollary}
\begin{proof}
The proof is given in Appendix~\ref{app:proof:lemma_relcp_uniqueness}.
\end{proof}

\subsection{Contact resolution between smooth, frictionless surfaces as a ReLCP} \label{sec:collision_relcp}
This subsection verifies that the contact recursion of Section~\ref{sec:augment} is an instance of the ReLCP defined above and identifies sufficient conditions for existence, finite termination, and uniqueness of the contact recursion solution. Following Anitescu \cite{anitescu_2006} and Anitescu--Cremer--Potra \cite{anitescu_1996}, we restrict attention to strictly convex smooth bodies in order to obtain explicit sufficient conditions for local regularity and exclusion of self-jamming. Under a small-timestep condition made precise in Corollary~\ref{cor:strictly_convex_finite_termination}, the recursion terminates in finitely many iterations with a unique, overlap-free solution. Outside the strictly convex regime, Eq.~\eqref{eq:bb-nonneg} remains the relevant solvability condition. We therefore rewrite the contact problem in the notation of Definition~\ref{def::relcp}, suppressing the physical timestep index $k$ and writing the generic ReLCP unknown $\bx_n$ for the collision force magnitude vector $\bgamma_n^{k,*}$. In the contact specialization, define
\begin{equation}
\bFcal_{\mathrm{c},n}^* \coloneqq \bDcal_n\bx_n^*,
\qquad
\bUcal_n^* \coloneqq \bMcal\bFcal_{\mathrm{c},n}^*,
\end{equation}
so $\bFcal_{\mathrm{c},n}^*$ is the contact wrench generated by the multipliers and $\bUcal_n^*$ is the induced rigid-body velocity. Thus $N_n=N_C^n$ in the contact problem, and
\begin{equation}
\begin{aligned}
\bAcal_n(\bx_{n-1}^*) &\coloneqq \Delta t\,\bDcal_n^\top\bMcal\bDcal_n, 
\,\,\tilde{\bg}_{n-1}(\bx_{n-1}^*) \coloneqq \tilde{\bPhi}_n, 
\,\,\bg_n(\bx_n) \coloneqq \bPhi_{n+1},\\
\bb_n(\bx_{n-1}^*) &\coloneqq
\bAcal_n(\bx_{n-1}^*)\iota_{N_C^n}\!\left(\bx_{n-1}^*\right)
-\tilde{\bg}_{n-1}(\bx_{n-1}^*).
\end{aligned}
\end{equation}
The matrix $\bDcal_n$ implicitly depends on $\bx_{n-1}^*$ through the newly appended rows $\nabla_{\bCcal_n}\left.\tilde{\bPhi}_n\right|_{\Delta\mathcal A_n}\bGcal$. Hence,
\begin{equation}
    \begin{aligned}
    \bCcal_{n+1} &= \bCcal_n + \Delta t  \bGcal \bMcal \bDcal_{n} \left( \bx_{n}^{*} - \iota_{N_C^n}\!\left(\bx_{n-1}^{*}\right) \right),
    \\
    \text{s.t. } &
    0 \leq \bg_n(\bx_n) \perp \bx_n \geq 0,
    \end{aligned}
\end{equation}
where $\bg_n$ has the Newton-esque residual form in Eq.~\eqref{eq:relcp_newton}.
New constraints are generated at each iteration if the true signed separation distance between any two body pairs is below some tolerance $-\epsilon_{NCP}<0$, i.e., $\left.\bPhi(\bCcal_n)\right|_{\Delta \mathcal{A}_n} < -\epsilon_{NCP}$ componentwise. Because the first $N_C^{n-1}$ components of $\tilde{\bg}_{n-1}(\bx_{n-1}^*)$ coincide with $\bg_{n-1}(\bx_{n-1}^*)$ and are therefore nonnegative, this is equivalent to
\begin{equation} \label{eq:gap_stopping_rule}
\left\|\min\!\left(\tilde{\bg}_{n-1}(\bx_{n-1}^*), 0\right)\right\|_\infty \le \epsilon_{NCP}.
\end{equation}
This geometric stopping rule is compatible with the ReLCP convergence criterion $\|\bDcal_c\bx_c^* - \bDcal_{c-1}\bx_{c-1}^*\|<\epsilon_f$ for sufficiently small $\epsilon_f$. In the contact interpretation, this criterion controls the contact wrench; since $\bMcal$ is fixed and symmetric positive definite over the timestep, it equivalently controls the induced velocities. Moreover, the definition of $\bAcal_n$ above has the Case~2 factorization in Definition~\ref{def::relcp}, with fixed symmetric positive definite matrix $\Delta t\,\bMcal$. Thus Eq.~\eqref{eq:relcp} is a ReLCP:
\begin{boxedtheorem} (Classifying the adaptive constraint generation scheme) \label{theorem::is_a_relcp}
Eq.~\eqref{eq:relcp} is a solvable ReLCP provided that, for every recursion index $n$, the constraint components used to form $\tilde{\bPhi}_n$ are twice continuously differentiable at their evaluation configurations (so $\bDcal_n$ exists and $\bAcal_n(\bx_{n-1}^*)=\Delta t\,\bDcal_n^\top\bMcal\bDcal_n$ is well-defined and symmetric positive semi-definite), and
\begin{equation}\label{eq:bb-nonneg}
\by^\top\bb_n(\bx_{n-1}^*) \le 0 
\quad\text{for every}\quad
\by\in\Ker\bAcal_n(\bx_{n-1}^*) \text{ with } \by\ge 0.
\end{equation}
\end{boxedtheorem}
\begin{proof}
The proof is given in Appendix~\ref{app:proof:theorem_is_a_relcp}.
\end{proof}

Following Anitescu \cite{anitescu_2006} and the shared-normal analysis of Anitescu, Cremer, and Potra \cite{anitescu_1996}, we now restrict attention to strictly convex smooth bodies. Strictly convex bodies have the property that every boundary point has a unique supporting hyperplane, so every normal direction corresponds to at most one boundary point. As proven by Anitescu, Cremer, and Potra \cite[Sec.~2 and Sec.~4.1.2]{anitescu_1996}, strictly convex smooth bodies also have the property that the shared-normal signed-separation map is locally single-valued and regular, as summarized in Lemma~\ref{lemma::local_regularity}. Finally, a pair of strictly convex bodies has the important property that their shared-normal signed separation distance is equivalent to a sphere-sphere signed separation distance between interior tangent balls, which gives the geometric control needed to exclude self-jamming from induced constraints.

\begin{boxedlemma} (Local regularity of the signed separation function for strictly convex smooth surfaces) \label{lemma::local_regularity}
Consider two strictly convex smooth bodies $a$ and $b$ at a configuration $\bCcal^\dagger$ for which the shared-normal signed separation function $\Phi_{ab}$ is well-defined and single-valued. Then there exist a neighborhood $U$ of $\bCcal^\dagger$ and a constant $\epsilon_{\mathrm{reg}}(\bCcal^\dagger)>0$ such that the mapping $\bCcal \mapsto \Phi_{ab}(\bCcal)$ is twice continuously differentiable on \(U \cap \left\{\bCcal : \Phi_{ab}(\bCcal)>-\epsilon_{\mathrm{reg}}(\bCcal^\dagger)\right\}\).
In particular, $\nabla_{\bCcal}\Phi_{ab}(\bCcal)$ is uniquely defined throughout this set.
\end{boxedlemma}
\begin{proof}
The proof is given in Appendix~\ref{app:proof:lemma_local_regularity}.
\end{proof}

We next replace the local shared-normal geometry by a uniform interior-ball construction valid on compact regularity sets. In this geometric discussion, $B(c,r)$ denotes the open Euclidean ball of radius $r$ centered at $c$, while $\overline{B_\rho(\bCcal)}$ denotes a closed configuration-space ball.

\begin{boxedlemma} (Uniform interior tangent-ball construction on a compact configuration set)
\label{lemma::uniform_interior_ball}
Let $K_a(\bC),K_b(\bC)\subset\mathbb R^d$ be strictly convex $C^2$ bodies moving by rigid motion with configuration $\bC$. Let $U$ be a compact set of configurations lying in a regime where Lemma~\ref{lemma::local_regularity} applies to the pair $(K_a,K_b)$, so that on $U$ the shared-normal signed-separation map is single-valued and its associated shared-normal points $(p_a(\bC),p_b(\bC))$ and unit normal $\bn(\bC)$ depend continuously on $\bC$. Here $\bn(\bC)$ is the outward unit normal of $K_a(\bC)$ at $p_a(\bC)$ and $-\bn(\bC)$ is the outward unit normal of $K_b(\bC)$ at $p_b(\bC)$. Write
\begin{equation}
\label{eq:rolling_ball_common_normal}
p_b(\bC)-p_a(\bC)=\Phi_{ab}(\bC)\,\bn(\bC).
\end{equation}
Then there exists $r_*>0$ such that, for every $\bC\in U$, the centers
\begin{equation}
\label{eq:rolling_ball_centers}
c_a(\bC):=p_a(\bC)-r_*\bn(\bC), \quad c_b(\bC):=p_b(\bC)+r_*\bn(\bC),
\end{equation}
define balls $B(c_a(\bC),r_*)\subset K_a(\bC)$ and $B(c_b(\bC),r_*)\subset K_b(\bC)$, each tangent at $p_a(\bC)$, respectively $p_b(\bC)$.
If
\begin{equation}
\label{eq:rolling_ball_cross_gap}
g(\bC):=\|c_b(\bC)-c_a(\bC)\|-2r_*,
\end{equation}
then, for every $\bC\in U$ satisfying $\Phi_{ab}(\bC)\ge -2r_*$, one has
\begin{equation}
\label{eq:rolling_ball_cross_gap_equals_phi}
g(\bC)=\Phi_{ab}(\bC).
\end{equation}
If $\Sigma_a$ and $\Sigma_b$ denote the sets of these centers in body-fixed coordinates, as $\bC$ ranges over $U$, then $\Sigma_a$ and $\Sigma_b$ are compact.
\end{boxedlemma}

\begin{proof}
The proof is given in Appendix~\ref{app:proof:lemma_uniform_interior_ball}.
\end{proof}

This representation supplies the body-fixed tangent-ball centers that the ReLCP freezes when a violated pair is appended. The next lemma shows that retaining the associated frozen-normal scalar constraint prevents the corresponding tangent balls from overlapping in later configurations.

\begin{boxedlemma} (Pairwise exclusion induced by retained frozen-normal constraints)
\label{lemma::pairwise_exclusion_from_relcp}
Assume the hypotheses of Lemma~\ref{lemma::uniform_interior_ball} hold for the ordered pair of bodies $(a_n,b_n)$ on a compact regularity set $U$, with tangent-ball radius $r_*$. Fix a recursion index $n$ at which this pair is appended at a configuration $\bCcal_n\in U$ satisfying
\[
-2r_*\le \Phi_n:=\Phi_{a_n b_n}(\bCcal_n)<0.
\]
Let $\bn_n$ be the shared-normal direction at $\bCcal_n$, and let $(\widehat c_{a_n}^n,\widehat c_{b_n}^n)$ be the body-fixed representations of the tangent-ball centers supplied by Lemma~\ref{lemma::uniform_interior_ball} at that configuration. For any configuration $\bC$, let $c_{a_n}^n(\bC)$ and $c_{b_n}^n(\bC)$ denote the spatial locations obtained by transporting these fixed body-frame centers with bodies $a_n$ and $b_n$, respectively.
Then Lemma~\ref{lemma::uniform_interior_ball} gives
\begin{equation} \label{eq:n_body_normal_diagonal_distance}
\|c_{b_n}^n(\bCcal_n)-c_{a_n}^n(\bCcal_n)\|=2r_*+\Phi_n<2r_*.
\end{equation}
Assume this frozen-normal constraint is retained for all $m>n$, and that at each such iteration it appears as a component of the ReLCP complementarity map, i.e., for some component index $\alpha(n)$,
\begin{equation}
\begin{aligned}
\bg_{m,\alpha(n)}\!(\bx_m^*)=
\bigl(c_{b_n}^n(\bCcal_m)&-c_{a_n}^n(\bCcal_m)\bigr)\cdot\bn_n-2r_*,
\\
0\le \bg_{m,\alpha(n)}\!(\bx_m^*)&\perp (\bx_m^*)_{\alpha(n)}\ge 0.
\end{aligned}
\end{equation}
Then
\begin{equation}
\label{eq:n_body_normal_exclusion}
\bigl(c_{b_n}^n(\bCcal_m)-c_{a_n}^n(\bCcal_m)\bigr)\cdot\bn_n\ge 2r_*
\,\,\forall m>n,
\end{equation}
which is the pairwise exclusion property in the frozen normal direction.
\end{boxedlemma}
\begin{proof}
The proof is given in Appendix~\ref{app:proof:lemma_pairwise_exclusion_from_relcp}.
\end{proof}

Condition \eqref{eq:bb-nonneg} has a natural mechanical interpretation: it excludes self-jamming for overlapping configurations, i.e., nonzero collision forces that produce zero net motion. For the recursively augmented ReLCP, one must verify that this exclusion persists after new constraints are appended and retained. The next lemma provides this recursion-level extension under strict convexity using only tangent-ball admissibility and frozen-normal exclusion.

\begin{boxedlemma} (Exclusion of self-jamming under strict convexity)
\label{lemma::self_jamming_exclusion}
Fix recursion index $n$ with
\begin{equation}
\mathcal A_n=\mathcal A_{n-1}\cup \Delta\mathcal A_n,\,\, N_C^n:=|\mathcal A_n|.
\end{equation}
Assume all bodies are strictly convex with $C^2$ boundaries, and fix $r_*>0$. For each retained constraint $\alpha\in\mathcal A_{n-1}$, let
$\hat{\bn}_\alpha$ be its frozen normal and
$\bar{\bc}_{a,\alpha}^n,\bar{\bc}_{b,\alpha}^n$
its transported frozen tangent-ball centers of radius $r_*$. Assume
\begin{equation}
\hat{\bn}_\alpha^\top(\bar{\bc}_{b,\alpha}^n-\bar{\bc}_{a,\alpha}^n)-2r_*\ge 0.
\end{equation}
For each body $\ell$ incident to $\mathcal A_n$, define the admissible interior reference region
\begin{equation}\label{eq:admissible_reference_region}
R_\ell^n:=\{\bx\in \operatorname{int}K_\ell(\bCcal_n): B(\bx,r_*)\subset K_\ell(\bCcal_n)\},
\end{equation}
Equivalently, $R_\ell^n$ is the radius-$r_*$ erosion of $K_\ell(\bCcal_n)$, i.e., the set of points in $K_\ell(\bCcal_n)$ whose distance to the boundary is at least $r_*$; thus the centers of radius-$r_*$ tangent balls lie in $R_\ell^n$. Choose $\mathbf r_\ell^n\in R_\ell^n$. For each new constraint $\alpha\in\Delta\mathcal A_n$, let
$\by_{a,\alpha}^n,\by_{b,\alpha}^n,\hat{\bn}_\alpha^n$
be the unique shared-normal data at $\bCcal_n$ (from strict convexity + local regularity), where $a(\alpha)$ and $b(\alpha)$ denote the two bodies incident to $\alpha$, and define
\begin{equation}
\eta_{\alpha,n}^{\mathrm{new}}
:=
(\hat{\bn}_\alpha^n)^\top(\by_{a,\alpha}^n-\mathbf r_{a(\alpha)}^n)
+
(\hat{\bn}_\alpha^n)^\top(\mathbf r_{b(\alpha)}^n-\by_{b,\alpha}^n)>0.
\end{equation}
Assume
\begin{equation}
\begin{aligned}
\Phi_\alpha(\bCcal_n) &\ge -\epsilon_n
\,\,(\alpha\in\Delta\mathcal A_n),\\
0\le \epsilon_n<\epsilon_{n,\mathrm{new}}^\star,&\,\,
\epsilon_{n,\mathrm{new}}^\star:=\min_{\alpha\in\Delta\mathcal A_n}\eta_{\alpha,n}^{\mathrm{new}}.
\end{aligned}
\end{equation}

Define, for every $\alpha\in\mathcal A_n$,
\(
q_{\alpha,n}:=\hat{\bn}_\alpha^\top(\mathbf r_{b(\alpha)}^n-\mathbf r_{a(\alpha)}^n),
\)
where $\hat{\bn}_\alpha$ denotes $\hat{\bn}_\alpha^n$ for $\alpha\in\Delta\mathcal A_n$ and denotes the retained frozen normal for $\alpha\in\mathcal A_{n-1}$.
Then $q_{\alpha,n}>0$ for all $\alpha$, and therefore
\begin{equation}
\Ker \bDcal_n\cap \mathbb R_+^{N_C^n}=\{\mathbf 0\}.
\end{equation}
Equivalently, self-jamming is excluded at recursion $n$.
\end{boxedlemma}

\begin{proof}
The proof is given in Appendix~\ref{app:proof:lemma_self_jamming_exclusion}.
\end{proof}

When the hypotheses of Lemma~\ref{lemma::self_jamming_exclusion} hold at every recursion, the lemma gives $\Ker\bDcal_n\cap\mathbb R_+^{N_C^n}=\{\mathbf 0\}$ for every $n$. Lemma 1.3.1 of Posp\'{i}\v{s}il \cite{pospisil_thesis_2015} gives
\begin{equation}\label{eq:kernel_equality}
    \Ker \bAcal_{n} = \Ker \bDcal_{n} =\left\{\bv \in \mathbb{R}^{N_C^n}: \bDcal_n \bv = 0 \right\}.
\end{equation}
Therefore $\Ker \bAcal_n\cap\mathbb{R}_+^{N_C^n}=\{\mathbf{0}\}$ at every recursion $n$, so condition~\eqref{eq:bb-nonneg} holds automatically throughout recursion. This recovers the classical overlap-free solvability regime and extends it to finite overlaps through tangent-ball clearance for newly created constraints together with frozen-normal exclusion for retained constraints. For larger overlaps, or outside the strictly convex regime, condition~\eqref{eq:bb-nonneg} must be checked directly.

In the absence of self-jamming and under the assumption of local regularity, the $n$\textsuperscript{th} ReLCP iteration is guaranteed to exist. We now turn this solvability criterion into a finite-termination argument using a packing argument applied to the interior tangent-ball construction.

\begin{boxedtheorem} (Finite $\epsilon$-termination from pairwise exclusion in the normal direction for finitely many bodies)
\label{theorem::finite_epsilon_normal_exclusion_n_body}
Let $\{K_\ell(\bC)\}_{\ell=1}^N\subset\mathbb R^d$ be strictly convex $C^2$ bodies moving by rigid motion with configuration $\bC$, where $N<\infty$. For each ordered pair of body labels $(a,b)$ with $a\neq b$, let $U_{ab}$ be a compact set of configurations such that every iterate $\bCcal_n$ whose active pair satisfies $a_n=a$ and $b_n=b$ lies in $U_{ab}$, and $U_{ab}$ lies in a regime where Lemma~\ref{lemma::local_regularity} applies to the pair $(K_a,K_b)$, so that on $U_{ab}$ the shared-normal signed-separation map is single-valued and its associated shared-normal points and unit normal depend continuously on $\bC$.

Applying Lemma~\ref{lemma::uniform_interior_ball} pairwise and shrinking radii if necessary, fix a common radius $r_*>0$ such that, for each $a\neq b$, there are compact sets $\Sigma_{ab}^{a}$ and $\Sigma_{ab}^{b}$ of the corresponding radius-$r_*$ centers in body-fixed coordinates. Assume, in addition, that each pairwise compact regime satisfies
\begin{equation}
\label{eq:n_body_pairwise_overlap_regime}
\Phi_{ab}(\bC)\ge -2r_*
\qquad\text{for all }\bC\in U_{ab},\ a\neq b.
\end{equation}

Let $\{\bCcal_n\}_{n\ge1}$ be a sequence of configurations such that, for each $n$, one ordered pair $(a_n,b_n)$ with $a_n\neq b_n$ is designated active and satisfies $\Phi_{a_n b_n}(\bCcal_n)<0$. For each $n$, use the construction in Lemma~\ref{lemma::pairwise_exclusion_from_relcp} to define $\bn_n$, $\widehat c_{a_n}^n$, $\widehat c_{b_n}^n$, $c_{a_n}^n(\cdot)$, and $c_{b_n}^n(\cdot)$. In particular,
\(
\widehat c_{a_n}^n\in\Sigma_{a_n b_n}^{a_n},
\widehat c_{b_n}^n\in\Sigma_{a_n b_n}^{b_n}.
\)
Assume that the retained ReLCP component associated with each frozen-normal constraint satisfies the retention and complementarity hypothesis of Lemma~\ref{lemma::pairwise_exclusion_from_relcp} for all later recursions. By that lemma, whenever $m>n$, the pairwise exclusion property \eqref{eq:n_body_normal_exclusion} holds.

For each ordered pair $(a,b)$ with $a\neq b$, define on $\Sigma_{ab}^{a}\times\Sigma_{ab}^{b}$ the metric
\begin{equation}
\label{eq:n_body_normal_product_metric}
d_\times((x,y),(x',y')):=\|x-x'\|+\|y-y'\|.
\end{equation}
Then for every $\epsilon>0$, only finitely many recursion indices satisfy $-\Phi_{a_n b_n}(\bCcal_n)\ge\epsilon$. More precisely,
\begin{equation}
\label{eq:n_body_normal_packing_bound}
\#\{n\ge1:-\Phi_{a_n b_n}(\bCcal_n)\ge\epsilon\}
\le
\sum_{a\neq b}
P(\Sigma_{ab}^{a}\times\Sigma_{ab}^{b},\epsilon;d_\times),
\end{equation}
where $P(X,\epsilon;d)$ denotes the maximum cardinality of an $\epsilon$-separated subset of the metric space $(X,d)$. Consequently, there exists $n_\epsilon<\infty$ with
\begin{equation}
0<-\Phi_{a_n b_n}(\bCcal_n)<\epsilon
\qquad
\text{for all } n\ge n_\epsilon.
\end{equation}
\end{boxedtheorem}

\begin{proof}
The proof is given in Appendix~\ref{app:proof:theorem_finite_epsilon_normal_exclusion_n_body}.
\end{proof}

The theorem gives finite geometric termination. For the ReLCP construction, two points remain: the terminal recursion must have bounded collision work, and the iterates must stay within the regularity regime on which the preceding geometry rests.

\begin{boxedcorollary} (Finite $\epsilon$-normal exclusion implies bounded collision work) \label{cor::epsilon_normal_exclusion_implies_bounded_work}
Assume the hypotheses of Theorem~\ref{theorem::finite_epsilon_normal_exclusion_n_body}. Fix a recursion tolerance $\epsilon_{NCP}>0$ and run the adaptive scheme of Eq.~\eqref{eq:relcp} with stopping rule \eqref{eq:gap_stopping_rule}. For each iteration, define the collision work by
\begin{equation}
\begin{aligned}
W_n &\coloneqq \left(\bx_n^*\right)^\top\bAcal_n\bx_n^*
= \Delta t\,\left(\bFcal_{\mathrm{c},n}^*\right)^\top\bMcal\,\bFcal_{\mathrm{c},n}^*
= \Delta t\,\left(\bUcal_n^*\right)^\top\bMcal^{-1}\bUcal_n^*.
\end{aligned}
\end{equation}
Then:
\begin{enumerate}[wide=0pt]
    \item $\{W_n\}$ is bounded from above (indeed eventually constant);
    \item $\sup_n\|\bUcal_n^*\|<\infty$.
\end{enumerate}
\end{boxedcorollary}
\begin{proof}
The proof is given in Appendix~\ref{app:proof:cor_epsilon_normal_exclusion_implies_bounded_work}.
\end{proof}

It remains to justify the regularity point, namely that sufficiently small timesteps keep the recursion inside the local regularity regime.

\begin{boxedlemma} (Small-timestep invariance of the local regularity regime) \label{lemma::timestep_regularity}
Fix one physical timestep $k\to k+1$ and consider the ReLCP recursion \eqref{eq:relcp} for strictly convex $C^2$ bodies. Assume the starting configuration is overlap-free, i.e., $\Phi_{ab}(\bCcal^k)\ge 0$ for all interacting body pairs $(a,b)$.
Then, for any fixed finite recursion index $c$, there exists $\Delta t_{\mathrm{reg}}(c)>0$ such that for every $0<\Delta t\le\Delta t_{\mathrm{reg}}(c)$, every ReLCP iterate $\bCcal_n$ with $n=0,\ldots,c$ remains in the local regularity regime; equivalently, $\Phi_{ab}(\bCcal_n)>-\epsilon_{\mathrm{reg}}$ for every active body pair $(a,b)$ and every $n\le c$.
\end{boxedlemma}
\begin{proof}
The proof is given in Appendix~\ref{app:proof:lemma_timestep_regularity}.
\end{proof}

With these ingredients in place, the strictly convex regime can be summarized in a single existence--uniqueness corollary.

\begin{boxedcorollary} (Strictly convex finite-termination existence--uniqueness regime for ReLCP contact resolution) \label{cor:strictly_convex_finite_termination}
Consider Eq.~\eqref{eq:relcp} for strictly convex smooth $C^2$ bodies with an overlap-free starting configuration $\Phi_{ab}(\bCcal^k)\ge0$ for all body pairs. Assume:
\begin{enumerate}[wide=0pt]
    \item \label{assume:small-dt} the timestep $\Delta t$ is sufficiently small (in the sense of Lemma~\ref{lemma::timestep_regularity}),
    \item \label{assume:retention} all violating pairs are augmented and retained as frozen-normal constraints at each iteration.
\end{enumerate}
By Lemma~\ref{lemma::timestep_regularity}, Assumption~\ref{assume:small-dt} confines every iterate within any prescribed finite prefix to a compact ball $K$ on which $\Phi_{ab}>-\epsilon_{\mathrm{reg}}$ and each active pair lies in the local regularity regime (Lemma~\ref{lemma::local_regularity}). By Lemma~\ref{lemma::uniform_interior_ball} applied on $K$, there exists a uniform interior-ball radius $r_*>0$ such that the admissible interior reference regions are given by~\eqref{eq:admissible_reference_region}. Shrinking $\epsilon_{\mathrm{reg}}$ (and $\rho_*$) if necessary so that $\epsilon_{\mathrm{reg}}\le 2r_*$, the tangent-ball clearance condition $\Phi_\alpha(\bCcal_n)+\eta_{\alpha,n}^{\mathrm{new}}>0$ follows: $\eta_{\alpha,n}^{\mathrm{new}}\ge 2r_*$ from the interior-ball construction and $\Phi_\alpha>-\epsilon_{\mathrm{reg}}\ge-2r_*$ from the regularity bound on $K$. These facts, together with Assumption~\ref{assume:retention} and Lemma~\ref{lemma::pairwise_exclusion_from_relcp}, satisfy all hypotheses of Lemma~\ref{lemma::self_jamming_exclusion}.
Then, for any prescribed overlap tolerance $\epsilon_{NCP}>0$, the adaptive constraint generation scheme terminates after finitely many ReLCP iterations at a discrete-time update for which no pair of smooth bodies overlaps by more than $\epsilon_{NCP}$. The resulting contact wrench $\bFcal_{\mathrm{c},c}^*=\bDcal_c\bx_c^*$ is unique, and therefore the induced rigid-body velocity $\bUcal_c^*=\bMcal\bFcal_{\mathrm{c},c}^*$ is unique.
\end{boxedcorollary}
\begin{proof}
The proof is given in Appendix~\ref{app:proof:cor_strictly_convex_finite_termination}.
\end{proof}

This uniqueness result should be read with three qualifications. First, the assumption that \emph{all} violating pairs are augmented at every iteration is algorithmic rather than physical; it is the default implementation used in all numerical experiments reported here. Second, numerically, two identical ellipsoids retain a unique contact pair for all overlaps short of center coincidence, suggesting that the regularity regime can be quite large for some strictly convex families and allowing for a unique configurational update even when reasonably large unconstrained motions are present. For such families, the regime treated by the corollary need not be especially restrictive in practice. Third, outside that regime, or for non-strictly convex bodies, the constraint-generation map may be set-valued. This does not preclude practical applicability: one may still impose a deterministic selection rule. Section~\ref{sec:examples} includes such a simulation for non-convex interlaced rings. As before, the uniqueness result concerns the induced rigid-body velocity $\bUcal_c^*=\bMcal\bDcal_c\bx_c^*$, equivalently the contact wrench $\bDcal_c\bx_c^*$, rather than the Lagrange multipliers $\bx_c^*$, which may remain non-unique when $\bAcal_n$ is only positive semi-definite.

\begin{boxedcorollary} (Reduction of the adaptive constraint generation scheme to the classical single-constraint SNSD LCP) \label{cor:relcp_reduction_to_snsd_lcp}
    Fix a timestep $k$ in the strictly convex smooth contact setting of Corollary~\ref{cor:strictly_convex_finite_termination}, and fix a geometric stopping tolerance $\epsilon_{NCP}>0$. Then there exists a threshold
    \begin{equation}
        \Delta t_{\mathrm{red}}(\bCcal^k,\epsilon_{NCP})>0
    \end{equation}
    such that, for every $0<\Delta t\le\Delta t_{\mathrm{red}}(\bCcal^k,\epsilon_{NCP})$, the trial configuration produced by the initial shared-normal signed-distance LCP solve already satisfies the stopping rule~\eqref{eq:gap_stopping_rule}. Consequently, no additional constraints are appended, the adaptive recursion of Eq.~\eqref{eq:relcp} terminates after that initial solve, and the converged ReLCP solution coincides exactly with the classical single-constraint shared-normal signed-distance LCP formulation of Stewart and Anitescu \cite{stewart_first_lcp_1996, anitescu_2006}.
\end{boxedcorollary}
\begin{proof}
The proof is given in Appendix~\ref{app:proof:cor_relcp_reduction_to_snsd_lcp}.
\end{proof}

\section{Numerical examples} \label{sec:examples}

This section presents a series of numerical examples that test the accuracy, convergence, and computational performance of the ReLCP time-integration scheme. These examples are designed to validate the theoretical properties established above and to highlight the practical advantages of ReLCP over traditional discrete-surface complementarity strategies. Throughout this section, the multi-sphere LCP formulation is denoted LCP-MS and the classical single-constraint shared-normal signed-distance LCP formulation is denoted LCP-SNSD. Three representative test cases are considered: (1) a convergence study of ellipsoid-ellipsoid contact, used to compare the accuracy of ReLCP to the LCP-MS approximation; (2) a performance benchmark in which a suspension of ellipsoids condenses under an external force field; and (3) a large-scale application simulating a growing bacterial colony. Together, these examples show that ReLCP accurately captures smooth surface interactions and provide numerical evidence consistent with the local reduction theorem above: under timestep refinement and at fixed stopping tolerance, the number of augmentation steps beyond the initial SNSD-LCP solve collapses to zero in the small-timestep regime. They also show that substantial performance increases are obtained in regimes where artificial surface roughness would otherwise introduce unphysical behavior.

\subsection{On performance metrics}
All numerical experiments presented in this work were conducted using modifications to a high-performance simulation library SimToolBox~\cite{SimToolbox,yan_lcp_col_stress_2019}, which serves as the backend for domain-specific applications such as aLENS~\cite{alens_2022} and SphereSimulator~\cite{yan_scalable_platform_2020}. SimToolBox is a hybrid MPI+OpenMP C\texttt{++} framework that leverages FDPS~\cite{fdps_2015} for particle management, Zoltan2~\cite{zoltan2_2012} for dynamic load balancing and sparse communication, and Tpetra/Belos~\cite{tpetra_2012, belos_2012} for distributed sparse linear algebra, most of which are subpackages within the Trilinos framework~\cite{trilinos_2025}.

To ensure fair and unbiased comparisons between contact time-integration strategies, each constraint generation method was implemented within the same codebase, modifying only the constraint generation and solution schemes, with all other components unchanged across tests. Executables were compiled using identical dependencies and compiler optimization flags, and all simulations were performed on dual-socket, 64-core Intel IceLake CPUs using two MPI processes (one per socket) and 32 OpenMP threads per process. Across all examples, the only free numerical parameter was the overlap tolerance, and the same value $10^{-5}$ was used for every method. For ReLCP, the recursive augmentation was terminated when the maximum true overlap of the underlying smooth surfaces fell below this same tolerance; no separate recursion tolerance was introduced. Across all numerical experiments, the ReLCP recursion typically converged in fewer than 10 iterations per timestep. This setup enables reporting of relative wall-clock performance with respect to a consistent baseline implementation. However, reported runtimes may vary across architectures and implementations, and thus a software-agnostic performance metric is also presented: the number of collision constraints generated and resolved at each timestep. This metric provides a hardware-independent proxy for algorithmic cost, offering insight into the scalability and efficiency of each method—independent of low-level implementation details. Other reported metrics such as convergence, constraint violation, and numerical stability will similarly remain software-agnostic.

\subsection{Two-ellipsoid refinement study}
\begin{figure*}[ht]
\centering
\includegraphics[width=0.8\textwidth]{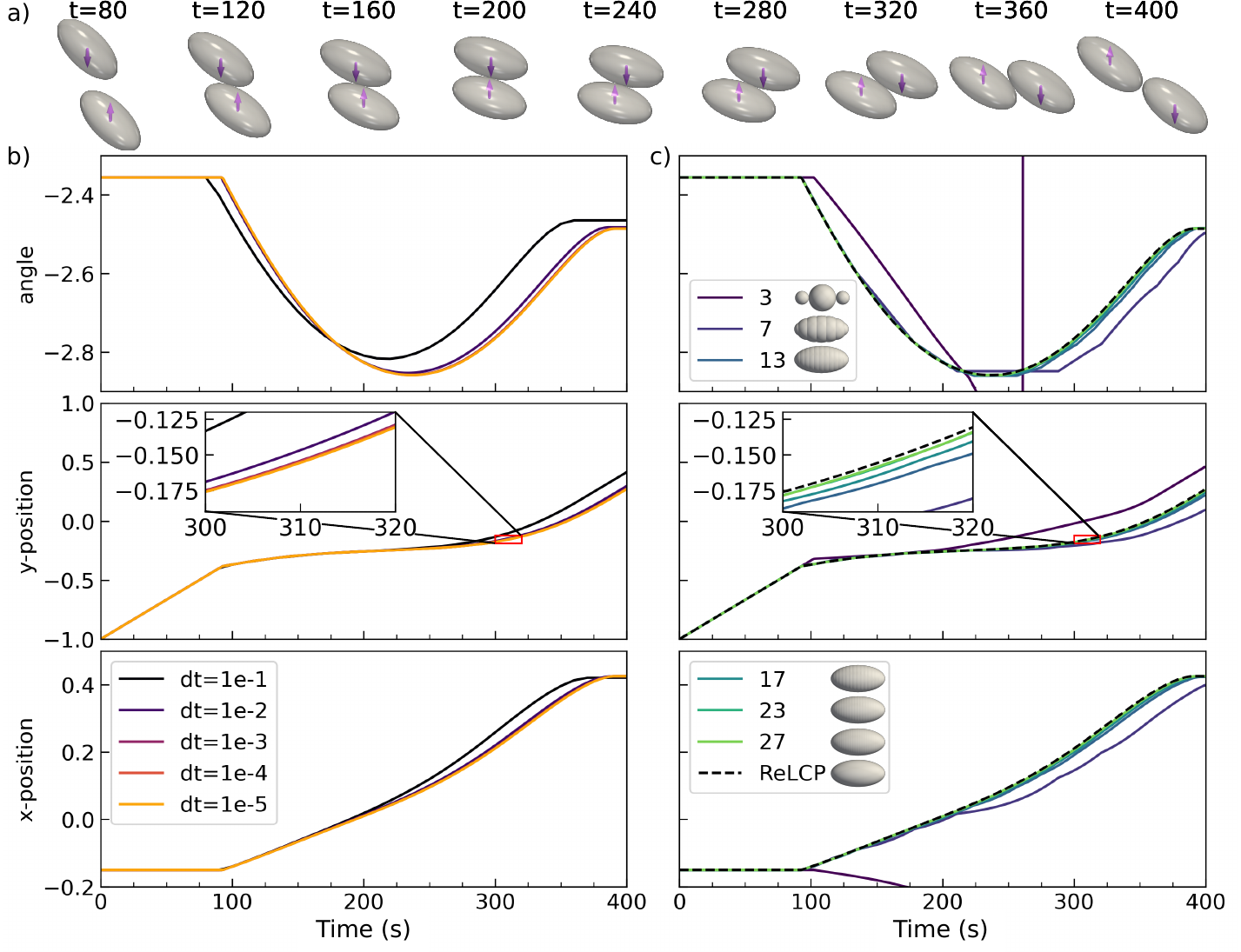}
\caption{
Collision and sliding dynamics of two identical ellipsoids driven by constant, equal-and-opposite external forces. 
\textbf{(a)}~Representative configurations spanning the interaction from initial contact to final separation. 
\textbf{(b)}~ReLCP trajectories obtained under timestep refinement, demonstrating self-convergence and reduction the classical LCP-SNSD formulation in the small-timestep regime. At the smallest timestep size, no more than one ReLCP iteration per timestep was required to satisfy an overlap tolerance of $10^{-5}$ diameters. 
\textbf{(c)}~Corresponding trajectories for LCP-MS under shape refinement, showing comparatively slow convergence toward the ReLCP reference solution as the number of sub-spheres increases.
}
\label{fig:two_ellipsoids}
\end{figure*}
The first test involves two identical ellipsoids of principal radii \( r_1 = r_2 = 1 \), \( r_3 = 2 \), lying within the \( x\text{-}y \) plane. Each ellipsoid is initially rotated by 45 degrees about the \( z \)-axis and positioned such that their centers are offset by a unit distance along $x$, ensuring that they collide and rotate as they slide past one another. A constant unit external force is applied to each ellipsoid: downward on the upper body and upward on the lower. For this test, the mobility of each ellipsoid is governed by overdamped dynamics according to a local, linear drag model:
\begin{equation}\label{eq:ellipsoid_mobility}
\bMcal_\ell = 
    \frac{1}{\xi}
    \begin{bmatrix}
      \dfrac{1}{L_\ell}\bI & \mathbf{0} \\ 
      \mathbf{0} & \dfrac{12}{L_\ell^3}\bI
    \end{bmatrix},
    \quad    
    \begin{bmatrix} 
        \bU_\ell \\ 
        \bOmega_\ell
    \end{bmatrix} = \bMcal_\ell
    \begin{bmatrix}
        \bF_\ell \\ \bT_\ell
    \end{bmatrix},
\end{equation}
where \( L_\ell = 2 r_3 \) denotes the major axis length of ellipsoid \( \ell \). The resulting interaction produces a glancing collision with rich rotational, translational, and sliding dynamics. For this system, the shared-normal signed-distance constraints were computed using the formulation of Wellmann and Wriggers~\cite{wellman_common_normal_2008}.

To evaluate the convergence behavior of the ReLCP formulation, a timestep refinement study was conducted, systematically decreasing the timestep size by four orders of magnitude while enforcing a stringent overlap tolerance of $10^{-5}$. The first column of Fig.~\ref{fig:two_ellipsoids} illustrates the temporal evolution of the two ellipsoids for a range of timestep sizes, capturing their interaction from initial contact (time $\sim100$) to final separation (time $\sim400$). As the timestep size is progressively reduced, the computed center-of-mass trajectories and orientations converge, in agreement with the classical consistency theory for complementarity time-stepping summarized in Section~\ref{sec:preliminaries}. Moreover, at the smallest timestep size tested, the solver converged after the initial SNSD-LCP solve, with no additional constraints appended. This behavior is consistent with the reduction result of Corollary~\ref{cor:relcp_reduction_to_snsd_lcp}, which predicts that, in the sufficiently-small-timestep regime, the adaptive constraint generation scheme collapses to the classical single-constraint SNSD-LCP formulation.

For comparative purposes, the aforementioned test was replicated using the LCP-MS approximation, whereby each ellipsoid is modeled as a sequence of overlapping spheres aligned along the body’s major axis. The selection of sphere positions and radii adheres to the methodology proposed by Markauskas et al.~\cite{markauskas_multisphere_adequacy_2010}, employing an odd number of sub-spheres to ensure the largest sphere remains centered. Each simulation was executed at a fixed timestep size of \( \Delta t = 10^{-5} \), and variations in the number of sub-spheres were introduced to facilitate an assessment of convergence. As depicted in the second column of Fig.~\ref{fig:two_ellipsoids}, coarse discretizations (for instance, 3 and 7 sub-spheres) yield dynamics that diverge significantly from the smooth ReLCP solution due to the introduction of artificial surface roughness. Notice that the 3 sub-sphere velocity extends outside of the graphical limits as a result of the two ellipsoids contacting and then rotating the opposite direction of all other tests. By 13 sub-spheres, the gross sliding trajectory is recovered much more faithfully, although visible discrepancies in both translation and rotation remain. Further refinement to 17 and 23 sub-spheres produces trajectories that are visually close to the ReLCP reference, yet the convergence reported in Table~\ref{tab:two_ellipsoids} remains slow in the \( L_2 \)-norm: even with 27 sub-spheres, the positional errors remain on the order of \(10^{-2}\). From a practical standpoint, these results suggest that, while the precise resolution requirement is application dependent, at least 13 sub-spheres are needed before the LCP-MS approximation begins to capture the sliding dynamics of smooth ellipsoids with reasonable fidelity. This interpretation is consistent with the findings of Markauskas et al.~\cite{markauskas_multisphere_adequacy_2010}, who studied the shape sensitivity of potential-based multi-sphere representations in a classical piling experiment: ellipsoids of aspect ratio 2.35 were allowed to form a static pile against a planar wall under gravity, and the resulting angle of repose was measured—the angle formed between the pile surface and the wall. They found that at least 13 sub-spheres per ellipsoid were required for the LCP-MS method to reproduce the angle of repose within 4\% of that produced by smooth ellipsoids, and 17 sub-spheres were needed to reach 1\% agreement—underscoring the geometric approximation error introduced by surface discretizations and the slow convergence of LCP-MS to the smooth-body limit.
\begin{table}[ht]
  \centering
    \caption{ReLCP timestep-refinement and LCP-MS shape-refinement errors for the two-ellipsoid problem. Errors are $L_2$ norms relative to the $\Delta t = 10^{-5}$ ReLCP trajectories.}
  \label{tab:two_ellipsoids}
  \begin{tabular}{llccc}
    \hline
    Study & Refinement & $x$-position & $y$-position & angle \\
    \hline
    \multirow{5}{*}{\shortstack[l]{ReLCP}} 
      & $\Delta t = 10^{-1}$ & 2.4530 & 1.0111 & 0.1078 \\
      & $\Delta t = 10^{-2}$ & 0.0256 & 0.0243 & 0.0041 \\
      & $\Delta t = 10^{-3}$ & 0.0048 & 0.0044 & 0.0006 \\
      & $\Delta t = 10^{-4}$ & 0.0026 & 0.0024 & 0.0003 \\
      & $\Delta t = 10^{-5}$ & 0.0000 & 0.0000 & 0.0000 \\
    \hline
    \multirow{6}{*}{\shortstack[l]{LCP-MS}} 
      & 3 sub-spheres  & 2.1326 & 0.2473 & 1.2034 \\
      & 7 sub-spheres  & 0.2086 & 0.1361 & 0.0145 \\
      & 13 sub-spheres & 0.0574 & 0.0422 & 0.0044 \\
      & 17 sub-spheres & 0.0309 & 0.0230 & 0.0023 \\
      & 23 sub-spheres & 0.0120 & 0.0087 & 0.0008 \\
      & 27 sub-spheres & 0.0103 & 0.0077 & 0.0007 \\
    \hline
  \end{tabular}
\end{table}

\subsection{Compacting ellipsoid suspension}

\begin{figure*}[ht]
\centering
\includegraphics[width=0.8\textwidth]{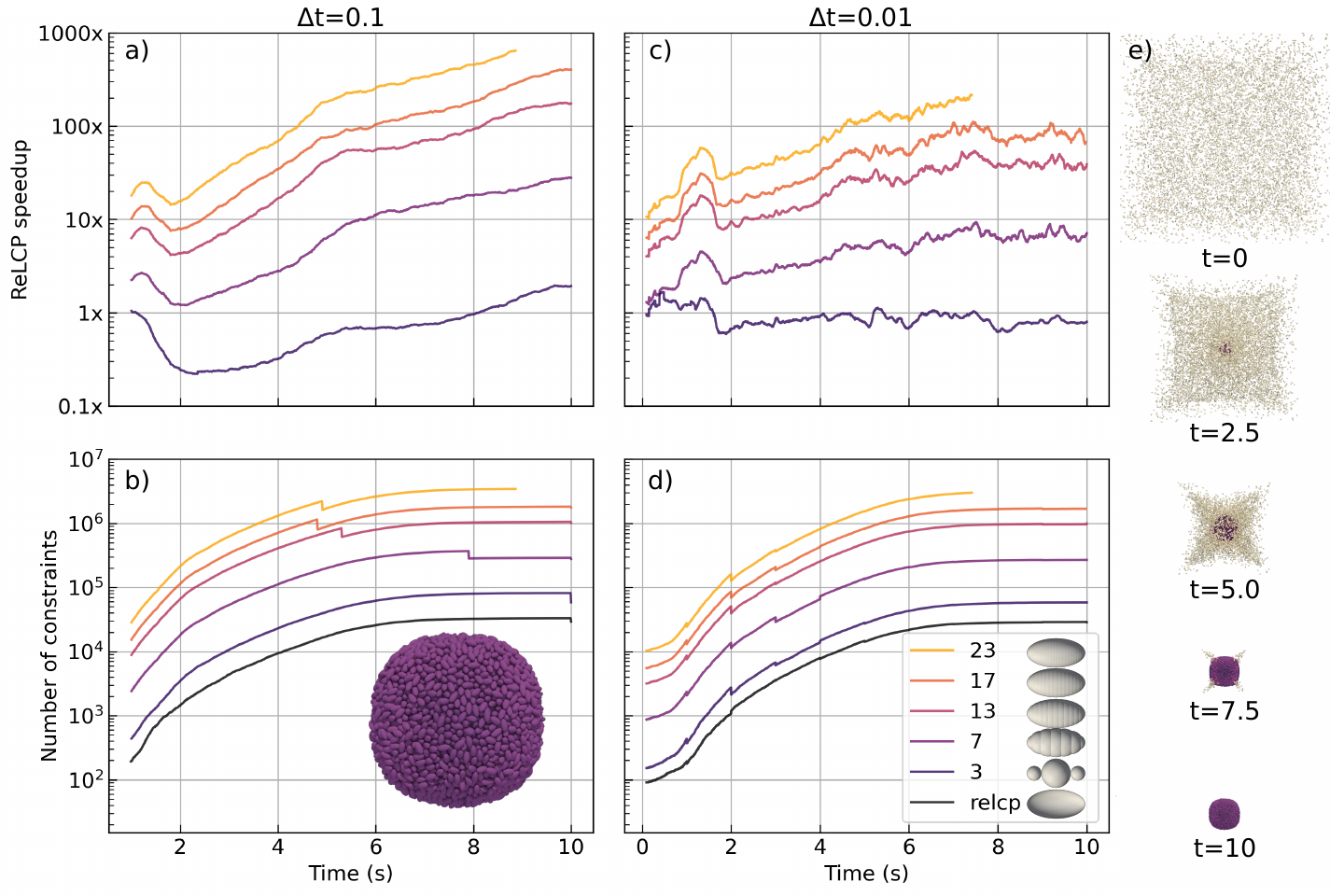}
\caption{
Performance comparison of ReLCP and LCP-MS contact time-stepping for a suspension of 10,000 aspect-ratio-2 ellipsoids compacting under the smooth radial force field
$
\bF(\bx) = -\bx (\|\bx\| - \sin\|\bx\|) / \|\bx\|^2
$
at two timestep sizes, $\Delta t = 0.1$ and $\Delta t = 0.01$, from an initially dilute state to a densely packed configuration. 
\textbf{(a,c)}~Runtime of LCP--MS relative to the ReLCP baseline, shown as a rolling average over 100 timesteps for several sub-sphere resolutions. The 27-sphere case was terminated early due to excessive runtime. 
\textbf{(b,d)}~Evolution of the total number of active collision constraints generated by each method. 
\textbf{(e)}~Representative snapshots from the ReLCP simulation illustrating the progression from a dilute suspension to a compact aggregate; the final state, shown at higher magnification, is inset in panel (b).
}
\label{fig:compression}
\end{figure*}

\begin{table}[ht] 
  \centering
    \caption{Final-state constraint counts and ReLCP speedups for the compacting ellipsoid suspension at two timestep sizes.}
  \label{tab:compression}
        \begin{tabular}{l l c c}
    \toprule
                $\Delta t$ & Method & Constraint Count & ReLCP Speedup \\
    \midrule
                \multirow{6}{*}{0.01}
                & LCP–MS 23 & $3.0 \times 10^6$ & 215 \\
                & LCP–MS 17 & $1.7 \times 10^6$ &  67 \\
                & LCP–MS 13 & $9.9 \times 10^5$ &  39 \\
                & LCP–MS 7  & $2.7 \times 10^5$ &   7 \\
                & LCP–MS 3  & $5.8 \times 10^4$ & 0.8 \\
                & ReLCP     & $2.9 \times 10^4$ & 1.0 \\
        \midrule
                \multirow{6}{*}{0.1}
                & LCP–MS 23 & $3.5 \times 10^6$ & 642 \\
                & LCP–MS 17 & $1.8 \times 10^6$ & 402 \\
                & LCP–MS 13 & $1.0 \times 10^6$ & 174 \\
                & LCP–MS 7  & $2.8 \times 10^5$ &  28 \\
                & LCP–MS 3  & $5.9 \times 10^4$ & 1.9 \\
                & ReLCP     & $2.9 \times 10^4$ & 1.0 \\
    \bottomrule
  \end{tabular}
\end{table}

To examine the relation between surface resolution and computational cost, we next consider a suspension of 10,000 ellipsoids compressed from a dilute state to a dense packed configuration under a smooth, radially inward force field,
\( \bF(\bx) = -\bx \frac{ \|\bx\| - \sin(\|\bx\|) }{ \|\bx\|^2 } \). The ellipsoids are initialized isotropically within a cubic domain at a volume fraction of 0.25\% and evolve under the same overdamped dynamics and local drag model used in the two-ellipsoid refinement study (Eq.~\eqref{eq:ellipsoid_mobility}). These simulations were performed for two timestep sizes: $\Delta t = 0.01$, chosen small enough that the initial SNSD-LCP solve was accepted without further augmentation, and $\Delta t = 0.1$, which lies well within the nonlinear regime.

As the suspension becomes denser, the cost of contact time integration rises sharply, particularly for LCP-MS, for which the number of constraints scales quadratically with the number of sub-spheres. Figure~\ref{fig:compression}(a--d) shows that this scaling leads to severe performance degradation at high volume fractions, and Table~\ref{tab:compression} summarizes the late-time cost in the compact configuration. A central point is that ReLCP achieves a computational cost comparable to the coarsest admissible multi-sphere proxy with 3 sub-spheres, while avoiding the substantial geometric error associated with so coarse a representation. By contrast, the preceding refinement study—consistent with the findings of Markauskas et al.~\cite{markauskas_multisphere_adequacy_2010}—indicated that approximately 13--17 sub-spheres per ellipsoid are needed before LCP-MS begins to reproduce the sliding behavior of smooth ellipsoids with reasonable fidelity. Relative to those practically relevant resolutions, the performance advantage of ReLCP is substantial: at $\Delta t = 10^{-2}$, ReLCP is $39\times$ faster than LCP-MS with 13 sub-spheres and $67\times$ faster than the 17-sphere case, while using $34\times$ and $58\times$ fewer constraints, respectively. This disparity widens at larger timestep size. At $\Delta t = 10^{-1}$, ReLCP is $174\times$ faster than the 13-sphere approximation and $402\times$ faster than the 17-sphere approximation. Taken together, these results show that once the multi-sphere resolution is made fine enough to recover smooth-surface sliding with acceptable fidelity, the computational cost of LCP-MS increases dramatically, whereas ReLCP attains comparable or better physical fidelity without the corresponding growth in constraint count.

\subsection{Bacterial colony growth}
\begin{figure*}[ht]
\centering
\includegraphics[width=0.8\textwidth]{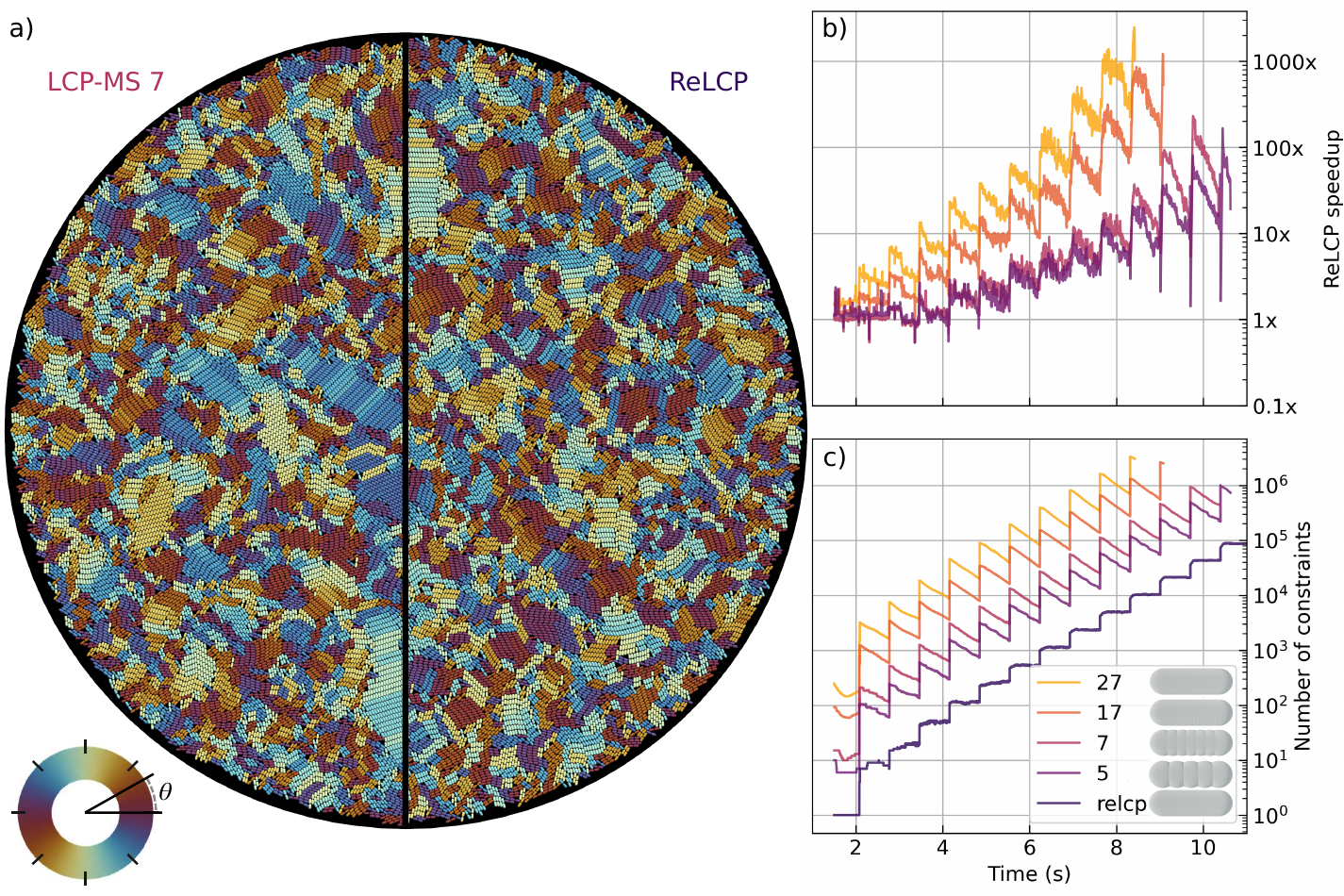}
\caption{
Growth simulation of a bacterial colony modeled as spherocylinders of length \(\ell\) and diameter \(b=0.5\), elongating at rate \(\dot{\ell} = \ell\). 
Each spherocylinder starts at \(\ell_0=2\) and divides into two upon reaching \(2\ell_0\). 
The colony doubles in cell count every \(\ln(2)\) seconds, starting from one cell and ending with \(2^{17}=131{,}072\) rods. 
\textbf{(a)}~Final colony configuration, with the left half shown for a 7-sub-sphere LCP-MS discretization and the right half shown for ReLCP; the full colony is approximately circular. 
\textbf{(b)}~Time evolution of the runtime of each LCP-MS simulation relative to the ReLCP baseline, shown as a rolling average over 100 timesteps, including both constraint generation and solution. The 17- and 27-sphere cases were terminated early because of excessive runtime.
\textbf{(c)}~Time evolution of the total number of active collision constraints for ReLCP and LCP-MS.
}
\label{fig:colony}
\end{figure*}
\begin{table}[ht]
  \centering
  \caption{Constraint statistics for the bacterial colony simulations (mean $\pm$ standard deviation).}
  \label{tab:compression_statistics}
  \begin{tabular}{lcc}
    \toprule
        Method & \shortstack{Relative \# Constraints\\vs. ReLCP} & \shortstack{Compute Time per\\Constraint [\textmu s]} \\
    \midrule
    LCP–MS 27  & $252.0 \pm 78.0$ & $15.3 \pm 20.1$ \\
    LCP–MS 17  & $101.0 \pm 32.8$ & $20.6 \pm 30.8$ \\
    LCP–MS 7   & $17.0 \pm 5.1$   & $45.9 \pm 101.0$ \\
    LCP–MS 5   & $8.77 \pm 2.47$  & $82.2 \pm 251.0$ \\
    ReLCP      & $1.00 \pm 0.00$  & $341.0 \pm 1010.0$ \\
    \bottomrule
  \end{tabular}
\end{table}

To illustrate the applicability of the ReLCP time-integration scheme in realistic, large-scale simulations, it is applied to a biologically motivated system: the simulation of a growing bacterial colony. Each bacterium is modeled as a growing spherocylinder (a cylinder with spherical end-caps) with a diameter of \( b = 0.5 \) and an initial length of \( \ell_0 = 2 \). The cells grow at an exponential rate with an elongation defined by \( \dot{\ell} = \ell \) and divide into two daughter cells upon reaching a length of \( 2\ell_0 \). To break symmetry and prevent purely one-dimensional growth, a slight random reorientation of \(\pm0.01\) degrees is applied to the daughter cells at the moment of division. The simulation begins with a single cell and progresses through 17 generations, culminating in a final colony of \( 2^17 = 131{,}072 \) cells. All cells interact solely through steric collision forces, governed by overdamped dynamics consistent with the local drag model used in previous sections. For spherocylinders, the shared-normal signed-distance constraints were computed from the minimum-distance problem between the two centerline segments using the standard line-segment distance algorithm of Eberly~\cite{eberly_distance_line3_line3}, after which the cap radii were subtracted to obtain the signed surface separation. As in the other examples, all methods used the same overlap tolerance of $10^{-5}$, and the ReLCP recursion was terminated once the true overlap of the underlying spherocylinder surfaces dropped below this value.

The performance of ReLCP is assessed against LCP-MS over a range of sub-sphere resolutions, with emphasis on both computational cost and the colony-scale structures that emerge from the underlying contact model. Figure~\ref{fig:colony} summarizes these results. Panel (a) shows the final colony configuration, with the left half obtained using 7 sub-spheres and the right half using ReLCP. Although both simulations reach the same colony size and overall volume fraction, they differ markedly in their internal organization. The LCP-MS approximation produces extended regions of local alignment, consistent with artificial surface roughness impeding relative sliding, whereas the ReLCP colony remains more isotropic and is organized into smaller aligned groups. This difference is directly attributable to the proxy-surface discretization: the bumpy multi-sphere boundary resists tangential rearrangement, effectively locking neighboring cells into extended nematic domains that would not form between the smooth spherocylinders. Thus, even at modest multi-sphere resolution, the proxy surface affects not only computational cost but also the emergent colony morphology.

Panel (b) reports the runtime of each LCP-MS simulation relative to the ReLCP baseline. Because cell division is synchronized, the population doubles every $\ln 2 \approx 0.69$ seconds, producing abrupt decreases in cell length and corresponding increases in the number of near neighbors; as the cells regrow, these neighbor counts relax, giving rise to the sawtooth structure visible in the performance curves. At early times the methods perform comparably, but the gap widens rapidly as the colony expands. For the largest systems, ReLCP is faster than LCP-MS with 5 or 7 sub-spheres by roughly one to two orders of magnitude, and faster than the 17- and 27-sphere approximations by more than two to three orders of magnitude. The precise sub-sphere resolution needed to reproduce smooth-surface dynamics is application dependent, but panel (a) already shows that 7 sub-spheres produces visibly different alignment behavior, at which point the computational penalty is already substantial. Panel (c) shows the corresponding number of active collision constraints over time: LCP-MS generates one to two orders of magnitude more constraints than ReLCP, and that increase directly translates into substantially larger complementarity problems and the observed loss of efficiency.

Table~\ref{tab:compression_statistics} quantifies these trends through simulation-averaged statistics. On average, ReLCP required approximately 9$\times$ fewer constraints than LCP-MS with 5 sub-spheres and 17$\times$ fewer than LCP-MS with 7 sub-spheres; at higher resolutions, the ratio increases to approximately 100--250$\times$. Although ReLCP incurs a higher cost per constraint because of its recursive iterative solve, that overhead is decisively offset by the much smaller number of constraints required.

\subsection{Non-convex objects: Rings}
\begin{figure*}[ht]
\centering
\includegraphics[width=0.8\textwidth]{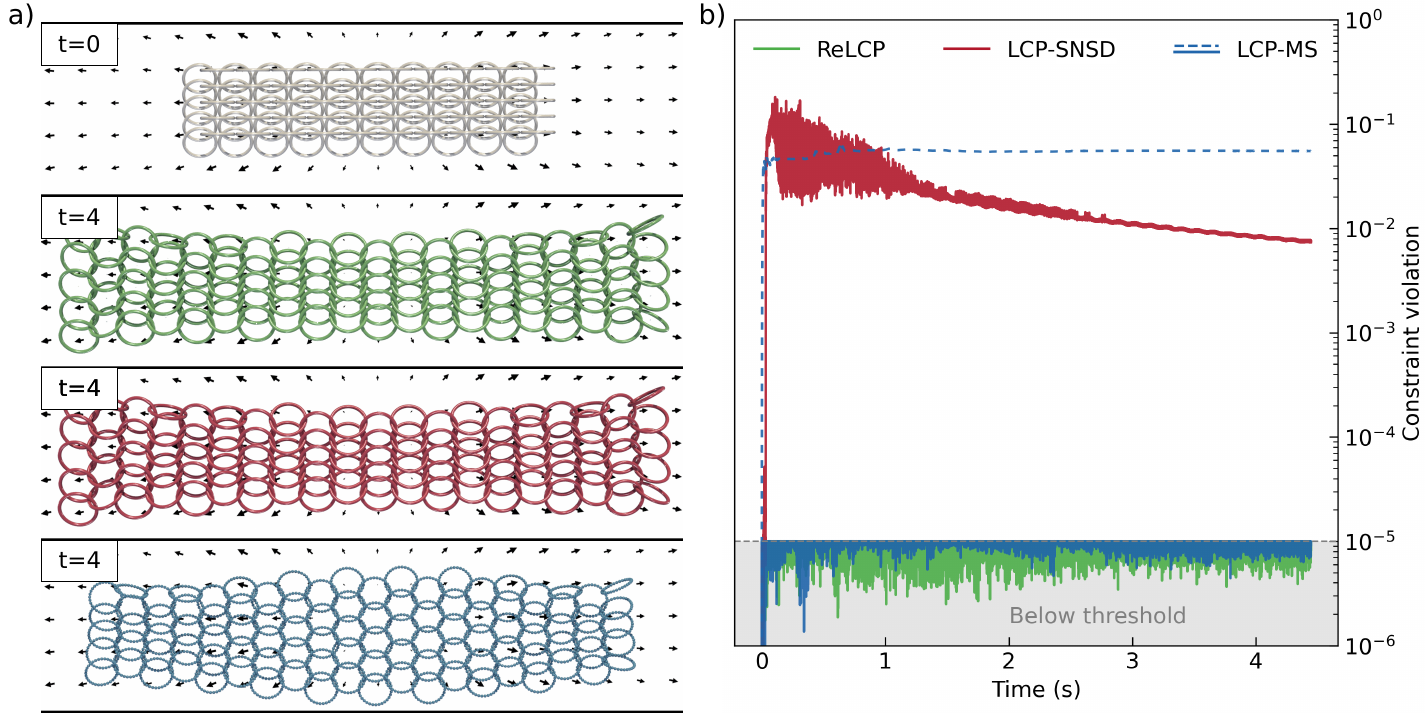}
\caption{
Chainmail-like network of interlaced rigid rings on a two-dimensional lattice subjected to a smooth extensional force field. The system evolves under overdamped dynamics from an initially relaxed configuration to a taut state. 
\textbf{(a)}~The initial overlap-free configuration of the chainmail network and the final taut configuration for each method. 
\textbf{(b)}~Time evolution of the maximum constraint violation for each method: ReLCP, LCP-MS evaluated with respect to both sub-sphere overlaps and the underlying ring surfaces, and LCP-SNSD. 
}
\label{fig:chainmail}
\end{figure*}

As a final example, the applicability of ReLCP to non-convex rigid bodies is examined through the simulation of a chainmail-like network of interlaced rings. In contrast to the preceding examples, which emphasize convergence and scalability, this case is intended primarily as a test of stability and numerical fidelity in a multiply connected system with geometrically intricate contacts. The results are compared against both LCP-MS and LCP-SNSD.

The chainmail network consists of rigid rings arranged on a two-dimensional lattice and initialized in a non-overlapping configuration, as shown in Fig.~\ref{fig:chainmail}(a). A smooth extensile force field,
$\bF(\bx) = \bx (\|\bx\| - \sin\|\bx\|) / \|\bx\|^2,$
is then applied so that the network expands toward a taut configuration, shown in Fig.~\ref{fig:chainmail}(b). Each ring is modeled as a torus with major radius $1.0$ and minor radius $0.1$ and evolves under overdamped dynamics. Here, the shared-normal signed distance between two rings was computed by solving the corresponding two-dimensional unconstrained minimization problem for the Euclidean distance between the two ring centerlines and then subtracting the two minor radii. Contact time integration is performed using three methods: ReLCP, LCP-MS with 32 spheres arranged along each ring centerline, and LCP-SNSD. Note, 32 sub-spheres is the smallest number of spheres that can be arranged along the ring centerline without producing gaps.

To assess numerical fidelity, the maximum absolute constraint violation is monitored over time. For LCP-MS, this quantity admits two distinct interpretations: the violation of the LCP constraints between sub-spheres, which defines the discrete solver tolerance (between the proxy surfaces), and the true overlap between the underlying continuous ring surfaces, which measures the geometric error introduced by the proxy surface. As shown in Fig.~\ref{fig:chainmail}(c), both ReLCP and LCP-MS maintain solver-level constraint violations below the prescribed tolerance of $10^{-5}$. When evaluated on the true ring geometry, however, the LCP-MS approximation exhibits a maximum overlap of approximately $60\%$ of the minor radius, consistent with the coarse resolution of the proxy surface. LCP-SNSD behaves in a different manner. Because orthogonal motions are left unconstrained and the timestep $\Delta t = 10^{-3}$ lies outside the linear regime, the method does not adequately suppress overlap-inducing modes of relative motion. The resulting constraint violation fluctuates substantially and at times exceeds the minor radius, producing visually apparent oscillatory motion.

These differences are evident in Supplemental Movie S1, which shows the time evolution of the chainmail network with rings colored by collision force magnitude. LCP-SNSD exhibits pronounced temporal fluctuations in the contact forces, with amplitudes substantially larger than those produced by ReLCP or LCP-MS. Although its large-scale deformation remains qualitatively similar to that of ReLCP, including an initial horizontal extension followed by vertical contraction, this behavior is accompanied by persistent overlap and force oscillations. LCP-MS, by contrast, produces stable force profiles but fails to reproduce the contractile rearrangements observed in the other two simulations, owing to geometric interlocking induced by surface discretization and the resulting non-physical resistance to sliding. Taken together, these results show that ReLCP can maintain stable and geometrically faithful contact dynamics even in complex non-convex systems, whereas alternative formulations either inherit substantial proxy-surface error or fail to constrain the dynamics adequately at practical timestep sizes.

\section{Conclusions}\label{sec:conclusion}
This work has presented a reformulation of complementarity-based time-stepping for nonsmooth contact between smooth rigid bodies as a recursively generated linear complementarity problem. The central contribution is to replace a fixed, pre-enumerated set of proxy-surface constraints with a recursive augmentation strategy in which additional unilateral constraints are introduced only when the discrete-time update predicted by the current contact set would violate nonpenetration for the underlying smooth surfaces. In this way, the method avoids the oversampling inherent to discrete proxy surfaces while preserving the complementarity structure of the time-integration scheme.

At the analytical level, the ReLCP formulation was placed on rigorous footing. Under explicit assumptions, the recursive augmentation is shown to terminate after finitely many LCP solves, thereby providing a constructive existence result for a converged discrete-time smooth-geometry contact update rather than only a conditional consistency statement about hypothetical solutions. Conditions were also established under which the resulting contact-induced rigid-body velocity is unique. In addition, for each fixed overlap-free discrete state and each fixed geometric stopping tolerance, we proved the reduction of the ReLCP to the classical single-constraint SNSD-LCP formulation in the limit of sufficiently small timestep size, thereby recovering the standard consistency properties of complementarity time-stepping under the stated regularity assumptions.

The numerical studies corroborate these analytical results and clarify the practical implications of surface representation in nonsmooth dynamics. The two-ellipsoid tests show convergence under timestep refinement and recovers the classical LCP-SNSD behavior in the small timestep regime, while multi-sphere approximations converge only slowly with increasing proxy-surface resolution. The compacting suspension and bacterial-colony simulations demonstrate that slow geometric convergence of proxy-surface models carries a substantial computational penalty: once the proxy surface is refined enough to reproduce smooth-surface sliding with acceptable fidelity, the number of collision constraints and the associated runtime increase dramatically. The chainmail example further shows that ReLCP remains stable in geometrically complex settings where single-constraint formulations exhibit persistent overlap and oscillatory behavior.

Taken together, these results show that the principal limitation is not complementarity-based time-stepping itself, but the reliance on discrete proxy surfaces to represent fundamentally smooth bodies. ReLCP addresses this limitation by generating only those constraints required by the predicted smooth-surface dynamics, thereby combining geometric fidelity with substantially improved scalability relative to high-resolution proxy-surface LCP formulations. More broadly, the formulation provides a systematic path for extending nonsmooth contact dynamics beyond oversampled geometric proxies while retaining the mathematical structure and robustness that make complementarity-based methods attractive for large-scale simulations.

The formal presentation of ReLCPs as a general mathematical object in Section~\ref{sec:relcp} was deliberately chosen to facilitate investigation of a broader question: what other problems can be cast as recursively generated linear complementarity problems for the sake of numerical and analytical treatment? A particularly promising direction is frictional contact. In standard complementarity-based formulations, the Coulomb friction cone is approximated by a discrete polyhedral cone, introducing a fixed set of tangential constraints whose number grows with the desired angular resolution of the friction law---a discretization issue that closely parallels the proxy-surface problem addressed here. One may ask whether friction can instead be treated through recursive augmentation: starting from a single tangential constraint along the current direction of sliding and introducing additional friction directions only when the trial update would violate the friction law. Such an approach would mirror the conceptual shift from discrete surface proxies to smooth-surface constraints that motivates the present work. Whether the resulting recursion inherits the monotone-decrease and finite-termination properties established here remains an open problem. Investigating this and other ReLCP formulations constitutes a natural and, in our view, compelling line of future research.

\section*{Declaration of competing interest}
The authors declare that they have no known competing financial interests or personal relationships that could have appeared to influence the work reported in this paper.

\section*{Acknowledgments}
We thank Wen Yan for developing the algorithmic and software framework upon which this paper extends. BP thanks the Flatiron Institute for hospitality while this research was carried out and all numerical computations for this work were performed at facilities supported by the Scientific Computing Core at the Flatiron Institute. TG is supported by NSF CAREER grant no. 2601612.

\appendix
\section{Proofs}\label{app:proofs}
\subsection{Proof of Theorem~\ref{theorem::relcp_conv}: Solvability and finite termination of the ReLCP}\label{app:proof:theorem_relcp_conv}
\begin{proof} The proof proceeds in five parts. 

\begin{enumerate}[wide=0pt]
    \item \textbf{Quadratic program equivalence:} By the equivalence of linear complementarity problems and convex quadratic programs (Lemma \ref{lemma::lcp_to_ccqp}), the ReLCP can be written as the fixed point iteration of the following minimization problem:
\begin{equation}
\bx_n^* = \argmin_{\bx \geq 0} f_n(\bx), \quad f_n(\bx) = \frac{1}{2} \bx^\top \bAcal_n(\bx_{n-1}^*) \bx - \bx^\top \bb_n(\bx_{n-1}^*).
\end{equation}

    \item \textbf{Existence of $\bx_n^*$:} By Lemma \ref{lemma::lcp_existence}, if either of the conditions specified in Definition \ref{def::relcp} are met, then $\bx_n^*$ is guaranteed to exist.
    \item \textbf{$f_n(\bx_n^*)$ decreases monotonically with $n$:} At optimality, 
    \begin{equation}\label{eq:optimality_complementarity}
    \begin{aligned}
    \left(\bx_n^*\right)^\top\left( \bAcal_n(\bx_{n-1}^*) \bx_n^* - \bb_n(\bx_{n-1}^*) \right) = 0 \Rightarrow \\
    \left(\bx_n^*\right)^\top \bb_n(\bx_{n-1}^*) = \left(\bx_n^*\right)^\top \bAcal_n(\bx_{n-1}^*) \bx_n^* .
    \end{aligned}
    \end{equation}
    Hence,
    \begin{equation}\label{eq:fi_simplified}
    \begin{aligned}
        f_n(\bx_n^*) 
        &= \frac{1}{2} \left(\bx_n^*\right)^\top \bAcal_n(\bx_{n-1}^*) \bx_n^* - \left(\bx_n^*\right)^\top \bb_n(\bx_{n-1}^*) \\
        &= -\frac{1}{2} \left(\bx_n^*\right)^\top \bAcal_n(\bx_{n-1}^*) \bx_n^* .
    \end{aligned}
    \end{equation}
    In the case that $\left(\bx_n^*\right)^\top\bAcal_n(\bx_{n-1}^*) \bx_n^*$ is bounded from above, $f_n(\bx_n^*)$ is thus bounded from below. By the variational inequality equivalence of convex quadratic programs (Lemma \ref{lemma::variational_equiv}), it follows that 
    \begin{equation}\label{eq:vi_applied}
    \begin{aligned}
    \forall \bx \geq 0 &: (\bx - \bx_n^*)^\top \left( \bAcal_n(\bx_{n-1}^*) \bx_n^* - \bb_n(\bx_{n-1}^*) \right) \geq 0 .
    \end{aligned}
    \end{equation}
    Let $\bx = \iota_{N_n}\!\left(\bx_{n-1}^*\right)$ and simplify:
    \begin{equation}\label{eq:vi_simplified}
    \begin{aligned}
    \iota_{N_n}\!\left(\bx_{n-1}^*\right)^\top \left( \bAcal_n(\bx_{n-1}^*) \bx_{n}^* - \bb_n(\bx_{n-1}^*) \right) &\geq 0.
    \end{aligned}
    \end{equation}
    If it additionally holds that the problem is Newton-esque,
    \begin{equation}
    \begin{aligned}
        \bb_n(\bx_{n-1}^*) &= \bAcal_n(\bx_{n-1}^*)\iota_{N_n}\!(\bx_{n-1}^*) - \tilde{\bg}_{n-1}(\bx_{n-1}^*), \\
        \pi_{N_{n-1}}\left(\tilde{\bg}_{n-1}(\bx_{n-1}^*)\right)&=\bg_{n-1}(\bx_{n-1}^*),
    \end{aligned}
    \end{equation}
    then the above simplifies to 
    \begin{equation}\label{eq:vi_newton_simplified}
    \begin{aligned}
     \iota_{N_n}\!\left(\bx_{n-1}^*\right)^\top\bAcal_n(\bx_{n-1}^*) \bx_{n}^* - \left(\bx_{n-1}^*\right)^\top\bAcal_{n-1}(\bx_{n-2}^*)\bx_{n-1}^* &\geq 0.
    \end{aligned}
    \end{equation}
    Indeed,
    \begin{equation}
    \begin{aligned}
    &\iota_{N_n}\!\left(\bx_{n-1}^*\right)^\top \bb_n(\bx_{n-1}^*) \\
    &\,= \iota_{N_n}\!\left(\bx_{n-1}^*\right)^\top\bAcal_n(\bx_{n-1}^*)\iota_{N_n}\!\left(\bx_{n-1}^*\right)
        - \iota_{N_n}\!\left(\bx_{n-1}^*\right)^\top\tilde{\bg}_{n-1}(\bx_{n-1}^*) \\
    &\,= \left(\bx_{n-1}^*\right)^\top\bAcal_{n-1}(\bx_{n-2}^*)\bx_{n-1}^*
        - \left(\bx_{n-1}^*\right)^\top\bg_{n-1}(\bx_{n-1}^*) \\
    &\,= \left(\bx_{n-1}^*\right)^\top\bAcal_{n-1}(\bx_{n-2}^*)\bx_{n-1}^*,
    \end{aligned}
    \end{equation}
    where the second line uses that $\bAcal_{n-1}$ is the upper-left block of $\bAcal_n$, the third line uses that $\iota_{N_n}\!\left(\bx_{n-1}^*\right)$ has zeros in the newly appended coordinates, and the last line follows from complementarity at recursion $n-1$.  
    To turn Eq.~\eqref{eq:vi_newton_simplified} into a monotonicity statement, we now distinguish the two cases.
    
    In Case~1, let $\bLcal_n$ be the Cholesky factor of $\bAcal_n$, so $\bAcal_n = \bLcal_n \bLcal_n^\top$. Because $\bAcal_{n-1}$ is the upper-left block of $\bAcal_n$, $\bLcal_{n-1}$ is the upper-left block of $\bLcal_n$, and therefore
    \begin{equation}
        \bLcal_n^\top\,\iota_{N_n}\!(\bx_{n-1}^*)
        =
        \iota_{N_n}\!\left(\bLcal_{n-1}^\top \bx_{n-1}^*\right).
    \end{equation}
    Denoting $\bz_n^* = \bLcal_n^\top \bx_{n}^*$ and $\hat{\bz}_{n-1}^* \coloneqq \iota_{N_n}\!\left(\bz_{n-1}^*\right)$,
    \begin{equation}\label{eq:vi_cholesky}
    \begin{aligned}
    \iota_{N_n}\!\left(\bx_{n-1}^*\right)^\top\bLcal_n \bLcal_n^\top \bx_{n}^* &\geq \left(\bx_{n-1}^*\right)^\top\bLcal_{n-1} \bLcal_{n-1}^\top\bx_{n-1}^*  \\
    \left(\hat{\bz}_{n-1}^*\right)^\top \bz_{n}^* &\geq \left(\hat{\bz}_{n-1}^*\right)^\top \hat{\bz}_{n-1}^* .
    \end{aligned}
    \end{equation}
    Applying the Cauchy--Schwarz inequality then gives
    \begin{equation}\label{eq:cauchy_schwarz_monotone}
    \|\hat{\bz}_{n-1}^*\| \|\bz_{n}^*\|
    \geq \left(\hat{\bz}_{n-1}^*\right)^\top \bz_{n}^* 
    \geq \left(\hat{\bz}_{n-1}^*\right)^\top \hat{\bz}_{n-1}^*
    = \|\hat{\bz}_{n-1}^*\|^2,
    \end{equation}
    which implies that $\|\bz_{n}^*\| \geq \|\hat{\bz}_{n-1}^*\| = \|\bz_{n-1}^*\|$.
    
    In Case~2, factorize $\bMcal = \bJcal\bJcal^\top$ with $\bJcal$ invertible and define $\bZ_n^* \coloneqq \bJcal^\top\bDcal_n \bx_n^*$. Since $\bDcal_{n-1}$ is the leftmost block of $\bDcal_n$,
    \begin{equation}
        \bJcal^\top\bDcal_n\iota_{N_n}\!\left(\bx_{n-1}^*\right)
        =
        \bJcal^\top\bDcal_{n-1}\bx_{n-1}^*
        =
        \bZ_{n-1}^*.
    \end{equation}
    Hence Eq.~\eqref{eq:vi_newton_simplified} yields
    \begin{equation}\label{eq:vi_case2}
    \begin{aligned}
        \left(\bZ_{n-1}^*\right)^\top\bZ_n^*
        \geq
        \left(\bZ_{n-1}^*\right)^\top\bZ_{n-1}^*.
    \end{aligned}
    \end{equation}
    Applying the Cauchy--Schwarz inequality gives
    \begin{equation}\label{eq:cauchy_schwarz_case2}
        \|\bZ_n^*\| \geq \|\bZ_{n-1}^*\|.
    \end{equation}
    By Eq.~\eqref{eq:fi_simplified}, $f_n(\bx_n^*) = -\frac12\|\bz_n^*\|^2$ in Case~1. In Case~2,
    \begin{equation}
        \left(\bx_n^*\right)^\top\bAcal_n(\bx_{n-1}^*)\bx_n^*
        =
        \left(\bDcal_n\bx_n^*\right)^\top\bMcal\left(\bDcal_n\bx_n^*\right)
        =
        \left(\bZ_n^*\right)^\top\bZ_n^*,
    \end{equation}
    so $f_n(\bx_n^*) = -\frac12\|\bZ_n^*\|^2$ there as well. Since the relevant norm is nondecreasing from one iteration to the next in either case, $f_n(\bx_n^*) \le f_{n-1}(\bx_{n-1}^*)$, so $f_n(\bx_n^*)$ decreases monotonically.
    
    \item \textbf{Finite convergence of $f_n(\bx_n^*)$ to within $\eta$:} $f_n(\bx_n^*)$ monotonically decreases and, by assumption, $\left(\bx_n^*\right)^\top\bAcal_n(\bx_{n-1}^*) \bx_n^*$ is bounded from above and (hence) $f_n(\bx_n^*)$ is bounded from below. By the monotone convergence theorem, $f_n(\bx_n^*)$ must converge, i.e., $f_n(\bx_n^*) \rightarrow F$ for some $F\in\mathbb{R}$ as $n\rightarrow \infty$. Equivalently, for every $\eta>0$ there exists a finite index $c_\eta$ such that
    \begin{equation}\label{eq:objective_gap_bound}
        0 \le f_n(\bx_n^*) - f_{n+1}(\bx_{n+1}^*) < \eta
        \qquad \forall n \ge c_\eta.
    \end{equation}
	
    \item \textbf{Convergence of $f_n(\bx_n^*)$ to within $\eta$ implies convergence of the ReLCP problem:} 
    Let $\epsilon_f>0$ be an arbitrary convergence tolerance. From here the proof branches based on which of the two cases in Definition~\ref{def::relcp} is satisfied by the ReLCP:
    \begin{enumerate}[label=(\alph*),wide=0pt]
        \item \textbf{Case 1:} Choose $\eta < \frac12 \mu\epsilon_f^2$ and $n\ge c_\eta$ as in Eq.~\eqref{eq:objective_gap_bound}. Since
        \begin{equation}
            f_n(\bx_n^*) = -\frac12\|\bz_n^*\|^2,
        \end{equation}
        Eq.~\eqref{eq:objective_gap_bound} yields
        \begin{equation}\label{eq:ui_gap_bound}
            0 \le \|\bz_{n+1}^*\|^2 - \|\bz_n^*\|^2
            = 2\left(f_n(\bx_n^*) - f_{n+1}(\bx_{n+1}^*)\right) < 2\eta.
        \end{equation}
        Combining Eq.~\eqref{eq:vi_cholesky} with Eq.~\eqref{eq:ui_gap_bound}, with $\hat{\bz}_{n}^* \coloneqq \iota_{N_{n+1}}\!\left(\bz_n^*\right)$, gives
        \begin{equation}\label{eq:ui_difference_bound}
        \begin{aligned}
            \left\|\bz_{n+1}^* - \hat{\bz}_n^*\right\|^2
            &= \|\bz_{n+1}^*\|^2 + \|\hat{\bz}_n^*\|^2
                - 2\left(\hat{\bz}_n^*\right)^\top\bz_{n+1}^* \\
            &\le \|\bz_{n+1}^*\|^2 - \|\hat{\bz}_n^*\|^2
            < 2\eta.
        \end{aligned}
        \end{equation}
        If $\bAcal_n$ is symmetric positive definite, then $\bLcal_n$ has full column rank and
        \begin{equation}
            \bz_{n+1}^* - \hat{\bz}_n^*
            = \bLcal_{n+1}^\top\left(\bx_{n+1}^* - \iota_{N_{n+1}}\!\left(\bx_n^*\right)\right).
        \end{equation}
        Therefore,
        \begin{equation}
            \left\|\bx_{n+1}^* - \iota_{N_{n+1}}\!\left(\bx_n^*\right)\right\|
            \le \frac{1}{\sigma_{\min}(\bLcal_{n+1})}
            \left\|\bz_{n+1}^* - \hat{\bz}_n^*\right\|.
        \end{equation}
        By assumption, there exists $\mu>0$ such that $\bv^\top\bAcal_n\bv \ge \mu\|\bv\|^2$ for all $\bv\in\mathbb{R}^{N_n}$ and all $n$, then $\lambda_{\min}(\bAcal_n)\ge \mu$ for all $n$. Since $\bAcal_n = \bLcal_n\bLcal_n^\top$, the smallest singular value of $\bLcal_n$ satisfies
        \begin{equation}
            \sigma_{\min}(\bLcal_n) = \sqrt{\lambda_{\min}(\bAcal_n)} \ge \sqrt{\mu}
            \qquad \forall n.
        \end{equation}
        Therefore,
        \begin{equation}
            \left\|\bx_{n+1}^* - \iota_{N_{n+1}}\!\left(\bx_n^*\right)\right\|
            < \sqrt{\frac{2\eta}{\mu}}
            < \epsilon_f,
        \end{equation}
        which is the convergence criterion of Definition~\ref{def::relcp} in Case~1.
        \item \textbf{Case 2:} Choose $\eta < \frac12\lambda_{\min}(\bMcal)\epsilon_f^2$ and $n\ge c_\eta$ as in Eq.~\eqref{eq:objective_gap_bound}. If $\bAcal_n$ is symmetric positive semi-definite and can be expressed as $\bAcal_n = \bDcal_n^\top\bMcal \bDcal_n$ for some rectangular matrix $\bDcal_n$ and symmetric positive definite matrix $\bMcal$, factorize $\bMcal = \bJcal \bJcal^\top$ with $\bJcal$ invertible and define $\bZ_n^* = \bJcal^\top\bDcal_n \bx_n^*$. Then
        \begin{equation}
            f_n(\bx_n^*) = -\frac12 (\bZ_n^*)^\top\bZ_n^*,
        \end{equation}
        so Eq.~\eqref{eq:objective_gap_bound} gives
        \begin{equation}
            0 \le \|\bZ_{n+1}^*\|^2 - \|\bZ_n^*\|^2 < 2\eta.
        \end{equation}
        Moreover, Eq.~\eqref{eq:vi_case2} with $n+1$ in place of $n$ yields
        \begin{equation}
        \begin{aligned}
            \left(\bZ_n^*\right)^\top\bZ_{n+1}^*
            \ge \left(\bZ_n^*\right)^\top\bZ_n^*.
        \end{aligned}
        \end{equation}
        Therefore,
        \begin{equation}
        \begin{aligned}
            \left\|\bZ_{n+1}^* - \bZ_n^*\right\|^2
            &= \|\bZ_{n+1}^*\|^2 + \|\bZ_n^*\|^2 - 2\left(\bZ_n^*\right)^\top\bZ_{n+1}^* \\
            &\le \|\bZ_{n+1}^*\|^2 - \|\bZ_n^*\|^2 < 2\eta.
        \end{aligned}
        \end{equation}
        Since
        \begin{equation}
            \bZ_{n+1}^* - \bZ_n^*
            = \bJcal^\top\left(\bDcal_{n+1}\bx_{n+1}^* - \bDcal_n\bx_n^*\right),
        \end{equation}
        the smallest eigenvalue of $\bMcal$ gives
        \begin{equation}
            \lambda_{\min}(\bMcal)
            \left\|\bDcal_{n+1}\bx_{n+1}^* - \bDcal_n\bx_n^*\right\|^2
            < 2\eta.
        \end{equation}
        Hence
        \begin{equation}
            \left\|\bDcal_{n+1}\bx_{n+1}^* - \bDcal_n\bx_n^*\right\| < \epsilon_f,
        \end{equation}
        which is the convergence criterion of Definition~\ref{def::relcp} in Case~2.
    \end{enumerate}
\end{enumerate}

\end{proof}

\subsection{Proof of Corollary~\ref{lemma::relcp_uniqueness}: Uniqueness of the ReLCP}\label{app:proof:lemma_relcp_uniqueness}
\begin{proof} The proof of this corollary follows from Lemma \ref{lemma::lcp_to_ccqp} and \ref{lemma::unique}:
\begin{enumerate}[wide=0pt]
    \item \textbf{Case 1:} Under Case 1 of Definition \ref{def::relcp}, Lemmas \ref{lemma::lcp_to_ccqp} and \ref{lemma::unique} imply that the solution $\bx_n^*$ to the $n$\textsuperscript{th} linear complementarity problem is unique. The additional uniform positive-definiteness assumption is exactly the hypothesis required in Theorem \ref{theorem::relcp_conv} to ensure that the sequence satisfies the Case 1 convergence criterion. Furthermore, because $\bAcal_n(\bx_{n-1}^*)$ and $\bb_n(\bx_{n-1}^*)$ are uniquely determined by $\bx_{n-1}^*$, the $n$\textsuperscript{th} linear complementarity problem (and therefore its solution $\bx_n^*$) is uniquely determined by $\bx_{n-1}^*$. Hence, the entire sequence $\{\bx_n^*\}$ up to convergence is deterministic, and so is its converged value $\bx_c^*$.
    \item \textbf{Case 2:} Under Case 2 of Definition \ref{def::relcp}, Lemmas \ref{lemma::lcp_to_ccqp} and \ref{lemma::unique} imply that the solution $\bx_n^*$ to the $n$\textsuperscript{th} linear complementarity problem need not be unique, but $\bDcal_n\bx_n^*$ is unique once $\bDcal_n$ is fixed. By hypothesis, the previously unique quantity $\bDcal_{n-1}\bx_{n-1}^*$ uniquely determines the auxiliary state $\bs_n$, which in turn uniquely determines $\bDcal_n$. Hence $\bDcal_n\bx_n^*$ is uniquely determined by the previous iteration. Inducting on $n$, the entire sequence $\{\bDcal_n\bx_n^*\}$ up to convergence is uniquely determined, and so is its converged value $\bDcal_c\bx_c^*$.
\end{enumerate}
\end{proof}

\subsection{Proof of Theorem~\ref{theorem::is_a_relcp}: Classifying the adaptive constraint generation scheme}\label{app:proof:theorem_is_a_relcp}
\begin{proof} By hypothesis, the constraint components used to form $\tilde{\bPhi}_n$ are twice continuously differentiable (hence differentiable) at their evaluation configurations. Therefore $\bDcal_n$ is well-defined at each recursion, and
\begin{equation}
\bAcal_n(\bx_{n-1}^*)=\Delta t\,\bDcal_n^\top\bMcal\bDcal_n
\end{equation}
is well-defined and symmetric positive semi-definite because $\bMcal$ is symmetric positive definite. It remains to verify that the stopping condition in Eq.~\eqref{eq:gap_stopping_rule} is compatible with the convergence argument of Thm.~\ref{theorem::relcp_conv}:
\begin{enumerate}[wide=0pt]
\item \textbf{Convergence of Eq.~\eqref{eq:gap_stopping_rule} implies configurational convergence: } Assume that Eq.~\eqref{eq:gap_stopping_rule} holds. Then no pair of bodies violates the geometric overlap tolerance, so no new constraints are added in the next recursion and $\Delta\mathcal{A}_n=\varnothing$. Hence the constraint set is unchanged, which gives
\begin{equation}
N_C^n=N_C^{n-1},\qquad
\bDcal_n=\bDcal_{n-1},\qquad
\bAcal_n=\bAcal_{n-1}.
\end{equation}
To show that the right-hand side is also unchanged, use the Newton-esque relation
\(
\bb_n=\bAcal_n\iota_{N_C^n}\!(\bx_{n-1}^*)-\tilde{\bg}_{n-1}(\bx_{n-1}^*).
\)
Because no components are appended, $\iota_{N_C^n}\!(\bx_{n-1}^*)=\bx_{n-1}^*$ and $\tilde{\bg}_{n-1}(\bx_{n-1}^*)=\bg_{n-1}(\bx_{n-1}^*)$, so
\begin{equation}
\bb_n
=\bAcal_{n-1}\bx_{n-1}^*-\bg_{n-1}(\bx_{n-1}^*)
=\bb_{n-1}.
\end{equation}
Thus the subsequent LCP has exactly the same data $(\bAcal,\bb)$ as the preceding one. By Case 2 of Lem.~\ref{lemma::unique}, although repeated solutions $\bx_n^*$ need not be unique, the quantity $\bDcal_n\bx_n^*$ is unique for that fixed LCP data. Consequently,
\(
\bDcal_n\bx_n^* = \bDcal_{n-1}\bx_{n-1}^*,
\)
which establishes the configurational convergence criterion exactly.
\item \textbf{Configurational convergence implies convergence of Eq.~\eqref{eq:gap_stopping_rule}: } Assume
\(
\|\bDcal_n\bx_n^* - \bDcal_{n-1}\bx_{n-1}^*\| < \epsilon_f.
\)
Since $\|\bv\|_\infty \le \|\bv\|$, this implies
\(
\|\bDcal_n\bx_n^* - \bDcal_{n-1}\bx_{n-1}^*\|_\infty < \epsilon_f.
\)
Let
\(
\Delta\bx_n \coloneqq \bx_n^* - \iota_{N_C^n}\!\left(\bx_{n-1}^{*}\right).
\)
Because $\bx_n^*$ solves the $n$\textsuperscript{th} LCP,
\begin{equation}
   \bg_{n}(\bx_{n}^*) =
   \tilde{\bg}_{n-1}(\bx_{n-1}^*) + \bAcal_{n}(\bx_{n-1}^*) \Delta\bx_n \ge 0.
\end{equation}
Hence, componentwise,
\begin{equation}
\begin{aligned}
    -\min\!\left(\tilde{\bg}_{n-1}(\bx_{n-1}^*), 0\right)
    \le
    \left|\bAcal_n(\bx_{n-1}^*)\Delta\bx_n\right|.
\end{aligned}
\end{equation}
Taking infinity norms and using $\bAcal_n = \Delta t \, \bDcal_n^\top \bMcal \, \bDcal_n$ gives
\begin{equation}
\begin{aligned}
\left\|\min\!\left(\tilde{\bg}_{n-1}(\bx_{n-1}^*), 0\right)\right\|_\infty
&\le
\left\|\bAcal_n(\bx_{n-1}^*)\Delta\bx_n\right\|_\infty \\
&\le
\Delta t \, \left\|\bDcal_n^\top \bMcal\right\|_\infty
\left\|\bDcal_n\Delta\bx_n\right\|_\infty.
\end{aligned}
\end{equation}
Because $\bDcal_n$ contains $\bDcal_{n-1}$ as its leftmost block,
\begin{equation}
\bDcal_n\iota_{N_C^n}\!\left(\bx_{n-1}^{*}\right)=\bDcal_{n-1}\bx_{n-1}^{*},
\end{equation}
so
\(
\left\|\bDcal_n\Delta\bx_n\right\|_\infty
=
\left\|\bDcal_n\bx_n^* - \bDcal_{n-1}\bx_{n-1}^*\right\|_\infty
< \epsilon_f.
\)
Therefore,
\begin{equation}
\left\|\min\!\left(\tilde{\bg}_{n-1}(\bx_{n-1}^*), 0\right)\right\|_\infty
\le
\Delta t \, \left\|\bDcal_n^\top \bMcal\right\|_\infty \epsilon_f.
\end{equation}
If the current ReLCP instance has no constraints, then Eq.~\eqref{eq:gap_stopping_rule} is trivial. Otherwise, $\bDcal_n$ contains the Jacobian of at least one unilateral contact constraint and is therefore nonzero. Since $\Delta t>0$, the factor $\Delta t \, \left\|\bDcal_n^\top \bMcal\right\|_\infty$ can vanish only if $\left\|\bDcal_n^\top \bMcal\right\|_\infty = 0$. But the infinity norm of a matrix is zero if and only if the matrix itself is zero, so this would imply $\bDcal_n^\top \bMcal = \mathbf 0$. Because $\bMcal$ is positive definite it would follow that $\bDcal_n = \mathbf 0$, a contradiction. Hence $\Delta t \, \left\|\bDcal_n^\top \bMcal\right\|_\infty > 0$, and since $\epsilon_{NCP}>0$ Eq.~\eqref{eq:gap_stopping_rule} holds whenever
\begin{equation}
\epsilon_f \leq \frac{\epsilon_{NCP}}{\Delta t \, \left\|\bDcal_n^\top \bMcal\right\|_{\infty}}.
\end{equation}
Moreover, $\left\|\bDcal_n^\top \bMcal\right\|_\infty$ is uniformly bounded in $n$. Because $\|\cdot\|_\infty$ is the maximum absolute row sum, it suffices to bound each contact row of $\bDcal_n^\top$ uniformly. Each row of $\bDcal_n^\top$ corresponds to one shared-normal unilateral constraint and has the rigid-body wrench form (for some body pair $(a,b)$)
\begin{equation}
\left[-\hat{\bn}^\top,\;-(\br_a\times\hat{\bn})^\top,\;\hat{\bn}^\top,\;(\br_b\times\hat{\bn})^\top\right]
\end{equation}
with zeros on all other body blocks, where $\|\hat{\bn}\|=1$ and $\br_a,\br_b$ are vectors from fixed body reference points to the contact points. Because each rigid body is compact, there is a finite geometric bound $R_{\max}$ such that $\|\br_a\|,\|\br_b\|\le R_{\max}$ for every possible contact row. Hence the absolute row sum of every such row is bounded by a constant $C_{\mathrm{row}}<\infty$ independent of $n$. Therefore
\begin{equation}
\|\bDcal_n^\top\|_\infty\le C_{\mathrm{row}}
\qquad \forall n.
\end{equation}
Importantly, this bound is independent of the number of retained constraints because $\|\bDcal_n^\top\|_\infty$ is a maximum over row sums, not a sum over all rows. Since $\bMcal$ is fixed,
\begin{equation}
\left\|\bDcal_n^\top \bMcal\right\|_\infty
\le
\left\|\bDcal_n^\top\right\|_\infty\left\|\bMcal\right\|_\infty
\le
C_{\mathrm{row}}\left\|\bMcal\right\|_\infty
\quad\forall n,
\end{equation}
hence
\(
\sup_n \left\|\bDcal_n^\top \bMcal\right\|_\infty < \infty.
\)

\item \textbf{Solvability: } The condition $\by^\top\bb_n(\bx_{n-1}^*) \leq 0$ for all $\by\geq 0$ with $\by \in \Ker \bAcal_n(\bx_{n-1}^*)$ is exactly the solvability condition in Definition~\ref{def::relcp}.
\end{enumerate}

Hence Eq.~\eqref{eq:relcp} is a solvable ReLCP, and Eq.~\eqref{eq:gap_stopping_rule} is compatible with the configurational convergence criterion.
\end{proof}

\subsection{Proof of Lemma~\ref{lemma::local_regularity}: Local regularity of the signed separation function for strictly convex smooth surfaces}\label{app:proof:lemma_local_regularity}
\begin{proof}
Anitescu \cite{anitescu_2006} states that the local regularity assumption used there holds for strictly convex smooth bodies, referring to the shared-normal analysis of Anitescu, Cremer, and Potra \cite[Sec.~2 and Sec.~4.1.2]{anitescu_1996}. The latter gives an open neighborhood $V$ of $\bCcal^\dagger$ in which the defining shared-normal pair is unique and the map $\Phi_{ab}$ is smooth. Choose an open neighborhood $U$ with $\bCcal^\dagger\in U$ and compact closure $\overline U\subset V$. Since $\Phi_{ab}$ is continuous on $\overline U$, the minimum
\begin{equation}
\phi_{\min}\coloneqq \min_{\bCcal\in\overline U}\Phi_{ab}(\bCcal)
\end{equation}
is finite. Define
\begin{equation}
\epsilon_{\mathrm{reg}}(\bCcal^\dagger)\coloneqq 1+\max(0,-\phi_{\min})>0.
\end{equation}
Then $\Phi_{ab}(\bCcal)\ge \phi_{\min}>-\epsilon_{\mathrm{reg}}(\bCcal^\dagger)$ for every $\bCcal\in U$, so \(U\cap\{\bCcal:\Phi_{ab}(\bCcal)>-\epsilon_{\mathrm{reg}}(\bCcal^\dagger)\}=U\).
Because $U\subset V$ and $\Phi_{ab}$ is smooth on $V$, $\Phi_{ab}$ is $C^2$ on the stated set, and $\nabla_{\bCcal}\Phi_{ab}$ is uniquely defined there.
\end{proof}

\subsection{Proof of Lemma~\ref{lemma::uniform_interior_ball}: Uniform interior tangent-ball construction on a compact configuration set}\label{app:proof:lemma_uniform_interior_ball}
\begin{proof}
Let $P_a$ and $P_b$ be the sets of shared-normal points realized on the two bodies (written in body-fixed coordinates) as $\bC$ ranges over $U$. Because $U$ is compact and the maps $\bC\mapsto p_a(\bC)$ and $\bC\mapsto p_b(\bC)$ are continuous on $U$, the sets $P_a$ and $P_b$ are compact.

Fix $p\in P_a$. Since $\partial K_a$ is $C^2$ and $K_a$ is strictly convex, the boundary has the local interior sphere property at $p$: there exist $r_p>0$ and a neighborhood $W_p\subset\partial K_a$ of $p$ such that, for every $q\in W_p$, the ball $B(q-r_p\hat n_a(q),r_p)$ lies in $K_a$ and is tangent to $\partial K_a$ at $q$, where $\hat n_a(q)$ is the outward unit normal of $K_a$ at $q$; see, for example, Schneider \cite{schneider_convex_bodies_2014}. The neighborhoods $W_p$ cover the compact set $P_a$, so choose a finite subcover $\{W_{p_\ell}\}_{\ell=1}^J$ and set $r_{*,a}:=\min_\ell r_{p_\ell}>0$. If $q\in P_a$, pick $\ell$ with $q\in W_{p_\ell}$. Then $B(q-r_{p_\ell}\hat n_a(q),r_{p_\ell})\subset K_a$, and because $r_{*,a}\le r_{p_\ell}$ one has
\begin{equation}
B(q-r_{*,a}\hat n_a(q),r_{*,a})\subset B(q-r_{p_\ell}\hat n_a(q),r_{p_\ell})\subset K_a,
\end{equation}
with tangency at $q$. Thus $r_{*,a}$ works for every point of $P_a$. The same argument applied to $K_b$ yields $r_{*,b}>0$ for every point of $P_b$. Set
\(
r_*:=\min\{r_{*,a},r_{*,b}\}>0.
\)
Then for every $\bC\in U$, the centers defined by \eqref{eq:rolling_ball_centers} give interior tangent balls of radius $r_*$ at $p_a(\bC)$ and $p_b(\bC)$.

By \eqref{eq:rolling_ball_common_normal} and \eqref{eq:rolling_ball_centers},
\begin{equation}
c_b(\bC)-c_a(\bC)=(\Phi_{ab}(\bC)+2r_*)\,\bn(\bC).
\end{equation}
Hence
\begin{equation}
g(\bC)=\|c_b(\bC)-c_a(\bC)\|-2r_*=|\Phi_{ab}(\bC)+2r_*|-2r_*.
\end{equation}
If $\Phi_{ab}(\bC)\ge -2r_*$, then $|\Phi_{ab}(\bC)+2r_*|=\Phi_{ab}(\bC)+2r_*$, so \eqref{eq:rolling_ball_cross_gap_equals_phi} follows.

The compactness of $\Sigma_a$ and $\Sigma_b$ follows because, in body-fixed coordinates, they are the images of the compact sets $P_a$ and $P_b$ under the inward-normal offset maps $p\mapsto p-r_*\hat n_a(p)$ and $p\mapsto p+r_*\hat n_b(p)$, respectively; these maps are continuous since the unit normal fields on $\partial K_a$ and $\partial K_b$ are continuous for $C^2$ boundaries.
\qedhere
\end{proof}

\subsection{Proof of Lemma~\ref{lemma::pairwise_exclusion_from_relcp}: Pairwise exclusion induced by retained frozen-normal constraints}\label{app:proof:lemma_pairwise_exclusion_from_relcp}
\begin{proof}
At the discovery configuration $\bCcal_n$, the transported fixed body-frame centers coincide with the instantaneous tangent-ball centers supplied by Lemma~\ref{lemma::uniform_interior_ball}. Therefore \eqref{eq:rolling_ball_cross_gap_equals_phi} gives \eqref{eq:n_body_normal_diagonal_distance}.

For any later recursion $m>n$, nonnegativity in complementarity gives
\(
\bg_{m,\alpha(n)}(\bx_m^*)\ge0.
\)
By identification with the retained scalar constraint,
\begin{equation}
\bigl(c_{b_n}^n(\bCcal_m)-c_{a_n}^n(\bCcal_m)\bigr)\cdot\bn_n-2r_*
=
\bg_{m,\alpha(n)}(\bx_m^*)
\ge0,
\end{equation}
which is equivalent to the pairwise exclusion inequality \eqref{eq:n_body_normal_exclusion}.
\end{proof}

\subsection{Proof of Lemma~\ref{lemma::self_jamming_exclusion}: Exclusion of self-jamming under strict convexity}\label{app:proof:lemma_self_jamming_exclusion}
\begin{proof}
Let $\bgamma_n\in \Ker \bDcal_n\cap \mathbb R_+^{N_C^n}$.
Bodywise translational balance from $\bDcal_n\bgamma_n=\mathbf 0$ gives
\begin{equation}
\sum_{\alpha=1}^{N_C^n}(\bgamma_n)_\alpha q_{\alpha,n}=0.
\end{equation}
For $\alpha\in\Delta\mathcal A_n$:
\begin{equation}
\left(\hat{\bn}_\alpha^n\right)^\top\!\left(\by_{a,\alpha}^n-\mathbf r_{a(\alpha)}^n\right)\ge r_*,
\qquad
\left(\hat{\bn}_\alpha^n\right)^\top\!\left(\mathbf r_{b(\alpha)}^n-\by_{b,\alpha}^n\right)\ge r_*.
\end{equation}
Indeed, $B(\mathbf r_{a(\alpha)}^n,r_*)\subset K_{a(\alpha)}(\bCcal_n)$ by definition of $R_{a(\alpha)}^n$, so \(\mathbf r_{a(\alpha)}^n+r_*\hat{\bn}_\alpha^n\in K_{a(\alpha)}(\bCcal_n)\). Since $\by_{a,\alpha}^n$ is the support point with outward normal $\hat{\bn}_\alpha^n$,
\begin{equation}
\left(\hat{\bn}_\alpha^n\right)^\top\!\left(\mathbf r_{a(\alpha)}^n+r_*\hat{\bn}_\alpha^n-\by_{a,\alpha}^n\right)\le 0,
\end{equation}
which gives the first inequality. The second follows analogously from \(\mathbf r_{b(\alpha)}^n-r_*\hat{\bn}_\alpha^n\in K_{b(\alpha)}(\bCcal_n)\) and the fact that $-\hat{\bn}_\alpha^n$ is the outward normal of $K_{b(\alpha)}(\bCcal_n)$ at $\by_{b,\alpha}^n$. Therefore
\begin{equation}
q_{\alpha,n}
=
\Phi_\alpha(\bCcal_n)+\eta_{\alpha,n}^{\mathrm{new}}
\ge -\epsilon_n+\eta_{\alpha,n}^{\mathrm{new}}
>0.
\end{equation}
For $\alpha\in\mathcal A_{n-1}$:
let
\begin{equation}
\by_{a,\alpha}^{\mathrm{ret}}:=\bar{\bc}_{a,\alpha}^n+r_*\hat{\bn}_\alpha,
\qquad
\by_{b,\alpha}^{\mathrm{ret}}:=\bar{\bc}_{b,\alpha}^n-r_*\hat{\bn}_\alpha,
\end{equation}
so that $B(\bar{\bc}_{a,\alpha}^n,r_*)\subset K_{a(\alpha)}(\bCcal_n)$ is tangent at $\by_{a,\alpha}^{\mathrm{ret}}$ with outward normal $\hat{\bn}_\alpha$, and $B(\bar{\bc}_{b,\alpha}^n,r_*)\subset K_{b(\alpha)}(\bCcal_n)$ is tangent at $\by_{b,\alpha}^{\mathrm{ret}}$ with outward normal $-\hat{\bn}_\alpha$.

We claim that the hyperplanes orthogonal to $\hat{\bn}_\alpha$ through the frozen tangent-ball centers support the eroded regions $R_{a(\alpha)}^n$ and $R_{b(\alpha)}^n$. Indeed, if $\bx\in R_{a(\alpha)}^n$, then $\bx+r_*\hat{\bn}_\alpha\in K_{a(\alpha)}(\bCcal_n)$ by definition of $R_{a(\alpha)}^n$. Since $\by_{a,\alpha}^{\mathrm{ret}}$ is the support point of $K_{a(\alpha)}(\bCcal_n)$ in direction $\hat{\bn}_\alpha$,
\begin{equation}
\hat{\bn}_\alpha^\top\bigl(\bx+r_*\hat{\bn}_\alpha-\by_{a,\alpha}^{\mathrm{ret}}\bigr)\le 0,
\end{equation}
hence
\begin{equation}
\hat{\bn}_\alpha^\top\bx\le \hat{\bn}_\alpha^\top\bar{\bc}_{a,\alpha}^n.
\end{equation}
Likewise, if $\bx\in R_{b(\alpha)}^n$, then $\bx-r_*\hat{\bn}_\alpha\in K_{b(\alpha)}(\bCcal_n)$, and because $-\hat{\bn}_\alpha$ is the outward normal at $\by_{b,\alpha}^{\mathrm{ret}}$,
\begin{equation}
(-\hat{\bn}_\alpha)^\top\bigl(\bx-r_*\hat{\bn}_\alpha-\by_{b,\alpha}^{\mathrm{ret}}\bigr)\le 0,
\end{equation}
which is equivalent to
\begin{equation}
\hat{\bn}_\alpha^\top\bx\ge \hat{\bn}_\alpha^\top\bar{\bc}_{b,\alpha}^n.
\end{equation}
Applying these bounds to the chosen reference points $\mathbf r_{a(\alpha)}^n\in R_{a(\alpha)}^n$ and $\mathbf r_{b(\alpha)}^n\in R_{b(\alpha)}^n$ gives
\begin{equation}
\begin{aligned}
q_{\alpha,n}
&=\hat{\bn}_\alpha^\top(\mathbf r_{b(\alpha)}^n-\mathbf r_{a(\alpha)}^n)\\
&\ge
\hat{\bn}_\alpha^\top(\bar{\bc}_{b,\alpha}^n-\bar{\bc}_{a,\alpha}^n)
\ge 2r_*>0.
\end{aligned}
\end{equation}
Thus $q_{\alpha,n}>0$ for every $\alpha$. Therefore
\begin{equation}
0=\sum_\alpha (\bgamma_n)_\alpha q_{\alpha,n},
\quad (\bgamma_n)_\alpha\ge 0,\ q_{\alpha,n}>0
\ \Longrightarrow\ 
(\bgamma_n)_\alpha=0\ \forall\alpha,
\end{equation}
i.e. $\bgamma_n=\mathbf 0$.
\end{proof}

\subsection{Proof of Theorem~\ref{theorem::finite_epsilon_normal_exclusion_n_body}: Finite $\epsilon$-termination from pairwise exclusion in the normal direction for finitely many bodies}\label{app:proof:theorem_finite_epsilon_normal_exclusion_n_body}
\begin{proof}
Fix ordered body labels $a,b$ with $a\neq b$, and let
\begin{equation}
I_{ab}:=\{n\ge1:a_n=a,\ b_n=b\}.
\end{equation}
For $j\in I_{ab}$, write 
\(\Phi_j:=\Phi_{ab}(\bCcal_j),\) \(\boldsymbol{\rho}_j(\bC):=c_{b_j}^j(\bC)-c_{a_j}^j(\bC),\) \( \widehat z_j:=(\widehat c_{a_j}^j,\widehat c_{b_j}^j)\in\Sigma_{ab}^{a}\times\Sigma_{ab}^{b}.\)
We first prove finite $\epsilon$-termination on this subsequence.

Fix $n,m\in I_{ab}$ with $n<m$. By Lemma~\ref{lemma::pairwise_exclusion_from_relcp}, the retained frozen-normal constraint discovered at recursion $n$ gives, at the later configuration $\bCcal_m$,
\begin{equation}
\label{eq:n_body_normal_exclusion_applied}
\boldsymbol{\rho}_n(\bCcal_m)\cdot\bn_n\ge2r_*.
\end{equation}
By \eqref{eq:n_body_normal_diagonal_distance},
\begin{equation}
\label{eq:n_body_normal_diagonal_distance_m}
\|\boldsymbol{\rho}_m(\bCcal_m)\|=2r_*+\Phi_m.
\end{equation}
Since $\|\bn_n\|=1$,
\begin{equation}
\boldsymbol{\rho}_m(\bCcal_m)\cdot\bn_n
\le
\|\boldsymbol{\rho}_m(\bCcal_m)\|
=
2r_*+\Phi_m.
\end{equation}
Subtracting this inequality from \eqref{eq:n_body_normal_exclusion_applied} yields
\begin{equation}
\label{eq:n_body_normal_metric_bound_body_fixed}
\begin{aligned}
-\Phi_m
&\le
\bigl(\boldsymbol{\rho}_n(\bCcal_m)-\boldsymbol{\rho}_m(\bCcal_m)\bigr)\cdot\bn_n \\
&\le
\|c_{b_n}^n(\bCcal_m)-c_{b_m}^m(\bCcal_m)\|
+
\|c_{a_n}^n(\bCcal_m)-c_{a_m}^m(\bCcal_m)\| \\
&=
d_\times(\widehat z_n,\widehat z_m),
\end{aligned}
\end{equation}
where the last equality follows by passing to body-fixed coordinates and using rigid-motion invariance. Therefore, if $-\Phi_m\ge\epsilon$, then
\begin{equation}
\label{eq:n_body_normal_epsilon_separated}
d_\times(\widehat z_n,\widehat z_m)\ge\epsilon
\qquad
\forall\, n<m \text{ in } I_{ab} \text{ with } -\Phi_m\ge\epsilon.
\end{equation}

Define $I_{ab,\epsilon}:=\{j\in I_{ab}:-\Phi_j\ge\epsilon\}$. By \eqref{eq:n_body_normal_epsilon_separated}, the set $\{\widehat z_j\}_{j\in I_{ab,\epsilon}}$ is an $\epsilon$-separated subset of the compact metric space $(\Sigma_{ab}^{a}\times\Sigma_{ab}^{b},d_\times)$.
Therefore
\begin{equation}
\label{eq:n_body_normal_single_pair_packing}
|I_{ab,\epsilon}|
\le
P(\Sigma_{ab}^{a}\times\Sigma_{ab}^{b},\epsilon;d_\times).
\end{equation}

Finally, the index set $I_\epsilon:=\{n\ge1:-\Phi_{a_n b_n}(\bCcal_n)\ge\epsilon\}$ is the disjoint union of the finitely many sets $I_{ab,\epsilon}$ over ordered pairs $(a,b)$ with $a\neq b$. Summing \eqref{eq:n_body_normal_single_pair_packing} over all such pairs yields
\begin{equation}
|I_\epsilon|
=
\sum_{a\neq b}|I_{ab,\epsilon}|
\le
\sum_{a\neq b}
P(\Sigma_{ab}^{a}\times\Sigma_{ab}^{b},\epsilon;d_\times)
<
\infty.
\end{equation}
Hence only finitely many $n$ satisfy $-\Phi_{a_n b_n}(\bCcal_n)\ge\epsilon$, and the stated finite $\epsilon$-termination follows.
\end{proof}

\subsection{Proof of Corollary~\ref{cor::epsilon_normal_exclusion_implies_bounded_work}: Finite $\epsilon$-normal exclusion implies bounded collision work}\label{app:proof:cor_epsilon_normal_exclusion_implies_bounded_work}
\begin{proof}
Apply Theorem~\ref{theorem::finite_epsilon_normal_exclusion_n_body} with $\epsilon=\epsilon_{NCP}$. Then there exists $n_{\epsilon_{NCP}}<\infty$ such that for all $n\ge n_{\epsilon_{NCP}}$, every active geometric violation satisfies $-\Phi_{a_n b_n}(\bCcal_n)<\epsilon_{NCP}$. Equivalently, from that index onward there are no pairs violating the augmentation threshold in \eqref{eq:gap_stopping_rule}. Hence no new constraints are appended, i.e., eventually $\Delta\mathcal A_n=\varnothing$. By the first implication proved in Theorem~\ref{theorem::is_a_relcp} (``Convergence of Eq.~\eqref{eq:gap_stopping_rule} implies configurational convergence''), once $\Delta\mathcal A_n=\varnothing$ the next LCP has identical data and therefore $\bFcal_{\mathrm{c},n}^*=\bFcal_{\mathrm{c},n-1}^*$.
Let $c$ denote the first index at which the stopping rule is satisfied, whose finiteness is supplied by Theorem~\ref{theorem::finite_epsilon_normal_exclusion_n_body} together with the stopping-rule compatibility in Theorem~\ref{theorem::is_a_relcp}. Extending the sequence after this index by the terminal fixed point gives $\bFcal_{\mathrm{c},n}^*=\bFcal_{\mathrm{c},c}^*$ for all $n\ge c$.

For $n\le c$, the values $W_n$ form a finite set, hence are bounded above. For $n\ge c$, the fixed-point identity gives $\bFcal_{\mathrm{c},n}^*=\bFcal_{\mathrm{c},c}^*$, and therefore also $\bUcal_n^*=\bUcal_c^*$ and $W_n=W_c$. Thus $\{W_n\}$ is bounded from above (in fact, eventually constant).

Finally, since $\bMcal$ is symmetric positive definite, so is $\bMcal^{-1}$, and
\begin{equation}
W_n=\Delta t\,\left(\bUcal_n^*\right)^\top\bMcal^{-1}\bUcal_n^*
\ge
\Delta t\,\lambda_{\min}(\bMcal^{-1})\,\|\bUcal_n^*\|^2,
\end{equation}
and therefore
\begin{equation}
\|\bUcal_n^*\|^2
\le
\frac{W_n}{\Delta t\,\lambda_{\min}(\bMcal^{-1})}
\le
\frac{\sup_m W_m}{\Delta t\,\lambda_{\min}(\bMcal^{-1})}<\infty.
\end{equation}
Therefore $\sup_n\|\bUcal_n^*\|<\infty$.
\end{proof}

\subsection{Proof of Lemma~\ref{lemma::timestep_regularity}: Small-timestep invariance of the local regularity regime}\label{app:proof:lemma_timestep_regularity}
\begin{proof}
Suppress the $k$ superscript and write 
$\bCcal_n:=\bCcal_n^{k+1}$, $\bgamma_n^*:=\bgamma_n^{k,*}$, $\bDcal_n:=\bDcal_n^{k+1}$, $\tilde{\bPhi}_n:=\tilde{\bPhi}_n^{k+1}$, $\bUcal_{\mathrm{free}}:=\bMcal^k\bFcal_{\mathrm{ext}}^k$. The case $c=0$ is immediate; fix $c\ge1$. The proof proceeds in three parts.

\begin{enumerate}[wide=0pt]

\item \textbf{Finite-prefix constants:}
Because the body set is finite and $\bCcal^k$ is overlap-free, Lemma~\ref{lemma::local_regularity} gives, for each ordered body pair $(a,b)$, a neighborhood of $\bCcal^k$ on which $\Phi_{ab}$ is single-valued and $C^2$. Intersecting these finitely many neighborhoods and shrinking $\rho_*>0$ if necessary yields $\epsilon_{\mathrm{reg}}>0$ and a compact ball $K:=\overline{B_{\rho_*}(\bCcal^k)}$ such that $\Phi_{ab}>-\epsilon_{\mathrm{reg}}/2$ and $\Phi_{ab}\in C^2(K)$ for every ordered pair $(a,b)$.

Since $c$ is fixed, only finitely many nonempty constraint patterns can occur in the first $c$ recursions: for each constraint one records its ordered body pair, the trial step at which it is first appended, and whether it appears at the current step as new or retained. Fix one such pattern, and let $\bDcal^0$ be the matrix obtained by evaluating its rows at the overlap-free base configuration $\bCcal^k$. At $\bCcal^k$, every new constraint has zero overlap depth, while every retained constraint satisfies the frozen-normal geometric condition there. Hence Lemma~\ref{lemma::self_jamming_exclusion} gives $\Ker\bDcal^0\cap\mathbb R_+^{N_C}=\{\mathbf 0\}$, where $N_C$ is the number of constraints in the pattern, and therefore $\min_{\|y\|=1,\,y\ge0}\|\bDcal^0 y\|>0$. For this fixed pattern, the associated Jacobian rows and the new-constraint clearances $\eta_{\alpha,n}^{\mathrm{new}}$ vary continuously with the trial configurations on $K$ at which the constraints are appended. After shrinking $\rho_*$ if necessary, the same pattern therefore admits positive lower bounds throughout the corresponding subset of $K$. Taking the minimum over the finitely many possible patterns yields constants $\mu_{R0}(c)>0$ and $\eta_*(c)>0$ such that every nonempty constraint set generated in the $c$-prefix with all relevant trial configurations in $K$ satisfies $\min_{\|y\|=1,\,y\ge0}\|\bDcal y\|\ge\mu_{R0}(c)$, and every newly appended constraint satisfies $\eta_{\alpha,n}^{\mathrm{new}}\ge\eta_*(c)$.
Set
\begin{equation}\label{eq:K_bounds_proof}
\begin{aligned}
L_K &:=\max_{(a,b)}\sup_{\bCcal\in K}\|\nabla_{\bCcal}\Phi_{ab}\|,\\ H_K&:=\max_{(a,b)}\sup_{\bCcal\in K}\|\nabla_{\bCcal}^2\Phi_{ab}\|,\\ D_K&:=\sup_{\bDcal\text{ in }c\text{-prefix on }K}\|\bDcal\|,
\end{aligned}
\end{equation}
where $L_K$ and $H_K$ are finite by compactness and $C^2$ regularity on $K$, and $D_K$ is finite because only finitely many such generated matrices can occur in the $c$-prefix and the relevant Jacobian blocks depend continuously on configuration.

\item \textbf{Finite-prefix induction:}
Scaling the $n$-th LCP of Eq.~\eqref{eq:relcp} by $1/\Delta t$ gives $0\le q_n+\bDcal_n^\top\bMcal^k\bDcal_n\bgamma_n^*\perp\bgamma_n^*\ge0$, with
\begin{equation}\label{eq:qn_scaled}
q_0:=\frac{\bPhi_0^k}{\Delta t}+\bDcal_0^\top\bUcal_{\mathrm{free}}, \,\,q_n:=\frac{\tilde{\bPhi}_n}{\Delta t}-\bDcal_n^\top\bMcal^k\bDcal_n\iota_{N_C^n}(\bgamma_{n-1}^*)\quad(n\ge1).
\end{equation}
Since $\bCcal^k$ is overlap-free, $\bPhi_0^k\ge0$ and
\begin{equation}
\left\|\min\!\left(q_0,0\right)\right\|\le\|\bDcal_0^\top\bUcal_{\mathrm{free}}\|=:Q_0.
\end{equation}
At step~$0$, $\bCcal_0=\bCcal^k\in K$ by construction; there are no retained constraints, so the clearance conditions of Lemma~\ref{lemma::self_jamming_exclusion} hold at $\bCcal^k$ and step~$0$ is solvable with $\bgamma_0^*\ge0$. Since $\bgamma_0^*\ge0$, only the negative entries of $q_0$ contribute to the complementarity inner product:
\begin{equation}
\begin{aligned}
\bgamma_0^{*\top}\bDcal_0^\top\bMcal^k\bDcal_0\bgamma_0^* = -\bgamma_0^{*\top}q_0 &= -\bgamma_0^{*\top}\min\!\left(q_0,0\right) \\
&\le \|\bgamma_0^*\|\left\|\min\!\left(q_0,0\right)\right\| \le \|\bgamma_0^*\|Q_0.
\end{aligned}
\end{equation}
If $\bDcal_0$ is empty, then $\bgamma_0^*=\mathbf 0$ and the estimate is trivial. Otherwise the left side is at least $\lambda_{\min}(\bMcal^k)\mu_{R0}(c)^2\|\bgamma_0^*\|^2$. If also $\bgamma_0^*=\mathbf 0$, then the desired bound already holds, since $\|\bgamma_0^*\|=0\le\Gamma_0$ for any choice of positive constant $\Gamma_0$; in particular, the subsequent definitions of $Z_0$ and $B_0$ remain valid. If instead $\bgamma_0^*\neq\mathbf 0$, dividing by $\|\bgamma_0^*\|$ gives
\begin{equation}
\|\bgamma_0^*\|\le\Gamma_0:=\frac{Q_0}{\lambda_{\min}(\bMcal^k)\mu_{R0}(c)^2},
\end{equation}
independently of $\Delta t$. Set
\begin{equation}
Z_0:=D_K\Gamma_0,
\qquad
B_0:=\|\bGcal^k\|\|\bMcal^k\|Z_0,
\qquad
S_0:=B_0,
\end{equation}
and choose $\delta_0>0$ so that $S_0\delta_0<\rho_*$ and $L_KS_0\delta_0<\eta_*(c)$. Assume inductively that for some $0\le m\le c-1$ there are constants $Q_j,Z_j,B_j$ ($0\le j\le m$) and $\delta_m>0$ such that, with $S_m:=\sum_{j=0}^m B_j$, every $0<\Delta t\le\delta_m$ satisfies, for each $0\le j\le m$,
\begin{equation}\label{eq:ind_hyp}
\bCcal_j\in K, \qquad \|\bCcal_{j+1}-\bCcal_j\|\le B_j\Delta t, \qquad \|\bDcal_j\bgamma_j^*\|\le Z_j,
\end{equation}
together with $\left\|\min\!\left(q_m,0\right)\right\|\le Q_m$, $S_m\delta_m<\rho_*$, and $L_KS_m\delta_m<\eta_*(c)$. Summing the increment bounds in~\eqref{eq:ind_hyp} gives
\begin{equation}
\|\bCcal_{m+1}-\bCcal^k\|
\le
\sum_{j=0}^m \|\bCcal_{j+1}-\bCcal_j\|
\le
S_m\Delta t
\le
S_m\delta_m
<
\rho_*,
\end{equation}
so $\bCcal_{m+1}\in K$. Hence every newly exposed overlap at step $m+1$ satisfies
\begin{equation}\label{eq:new_overlap_clearance_bound}
\begin{aligned}
-\Phi_\alpha(\bCcal_{m+1})
&\le
L_K\|\bCcal_{m+1}-\bCcal^k\|
\le
L_KS_m\Delta t \\
&<
L_KS_m\delta_m
<
\eta_*(c) \quad
(\alpha\in\Delta\mathcal A_{m+1}).
\end{aligned}
\end{equation}

\begin{enumerate}[label=(\alph*),wide=0pt]
\item \textbf{Solvability at step $m+1$:} For each $\alpha\in\Delta\mathcal A_{m+1}$, Eq.~\eqref{eq:new_overlap_clearance_bound} and the definition of $\eta_*(c)$ give $\Phi_\alpha(\bCcal_{m+1})+\eta_{\alpha,m+1}^{\mathrm{new}}>0$. The eroded-body regions~\eqref{eq:admissible_reference_region} are nonempty for the incident bodies by construction of the finite-prefix constants. Now let $\alpha\in\mathcal A_m$ be retained, and let $j(\alpha)\le m$ denote the recursion at which $\alpha$ was first appended. By construction, $\alpha$ has been retained at every recursion from $j(\alpha)+1$ through $m$, and at each such recursion it appears as a frozen-normal component of the ReLCP complementarity map. Therefore Lemma~\ref{lemma::pairwise_exclusion_from_relcp} gives the retained-constraint geometric condition at the current configuration $\bCcal_{m+1}$. Moreover, the carried scalar $\iota_{N_C^{m+1}}(\bgamma_m^*)_\alpha$ is nonnegative because $\bgamma_m^*\ge0$. Thus all hypotheses of Lemma~\ref{lemma::self_jamming_exclusion} hold at step $m+1$, so $\Ker\bDcal_{m+1}\cap\mathbb{R}_+^{N_C^{m+1}}=\{\mathbf{0}\}$. Theorem~\ref{theorem::is_a_relcp} then gives solvability.

\item \textbf{Bounding $\left\|\min\!\left(q_{m+1},0\right)\right\|$:} The update formula and append construction of Eq.~\eqref{eq:relcp} give
\begin{equation}
\tilde{\bPhi}_{m+1}|_{\mathcal{A}_m}=\tilde{\bPhi}_m+\Delta t\bDcal_m^\top\bMcal^k\bDcal_m\bigl(\bgamma_m^*-\iota(\bgamma_{m-1}^*)\bigr).
\end{equation}
Using the frozen-normal identity $\bDcal_{m+1}|_{\mathcal{A}_m}=\bDcal_m$ from Eq.~\eqref{eq:relcp}, substitution into~\eqref{eq:qn_scaled} gives
\begin{equation}
\left.q_{m+1}\right|_{\mathcal A_m}=q_m.
\end{equation}
Hence
\begin{equation}
\left\|\min\!\left(\left.q_{m+1}\right|_{\mathcal A_m},0\right)\right\|=\left\|\min\!\left(q_m,0\right)\right\|\le Q_m.
\end{equation}
Now let $\alpha\in\Delta\mathcal A_{m+1}$. Since $\alpha$ is newly appended at step $m+1$, Eq.~\eqref{eq:qn_scaled} gives
\begin{equation}
q_{m+1,\alpha}=\frac{\Phi_\alpha(\bCcal_{m+1})}{\Delta t}-\bigl(\bDcal_{m+1}^\top\bMcal^k\bDcal_m\bgamma_m^*\bigr)_\alpha.
\end{equation}
By Eq.~\eqref{eq:new_overlap_clearance_bound},
\begin{equation}
-\frac{\Phi_\alpha(\bCcal_{m+1})}{\Delta t}\le L_KS_m,
\end{equation}
and by the definition of $D_K$ together with $\|\bDcal_m\bgamma_m^*\|\le Z_m$,
\begin{equation}
\left|\bigl(\bDcal_{m+1}^\top\bMcal^k\bDcal_m\bgamma_m^*\bigr)_\alpha\right|
\le
\|\bDcal_{m+1}\|\,\|\bMcal^k\|\,\|\bDcal_m\bgamma_m^*\|
\le
D_K\|\bMcal^k\|Z_m.
\end{equation}
Therefore every new coordinate satisfies
\begin{equation}
\left|\min\!\left(q_{m+1,\alpha},0\right)\right|
\le
L_KS_m+D_K\|\bMcal^k\|Z_m.
\end{equation}
Summing over the new block gives
\begin{equation}
\left\|\min\!\left(\left.q_{m+1}\right|_{\Delta\mathcal A_{m+1}},0\right)\right\|
\le
\sqrt{|\Delta\mathcal A_{m+1}|}\bigl(L_KS_m+D_K\|\bMcal^k\|Z_m\bigr).
\end{equation}
Combining the old and new blocks,
\begin{equation}\label{eq:Qn_recursion}
Q_{m+1}:=Q_m+\sqrt{|\Delta\mathcal{A}_{m+1}|}\bigl(L_KS_m+D_K\|\bMcal^k\|Z_m\bigr),
\end{equation}
and hence $\left\|\min\!\left(q_{m+1},0\right)\right\|\le Q_{m+1}$. The constant $Q_{m+1}$ depends only on previously constructed bounds and is therefore independent of $\Delta t$.

\item \textbf{Multiplier and increment bounds:} Complementarity gives $\bgamma_{m+1}^{*\top}\bDcal_{m+1}^\top\bMcal^k\bDcal_{m+1}\bgamma_{m+1}^*=-\bgamma_{m+1}^{*\top}q_{m+1}$. Since $\bgamma_{m+1}^*\ge0$, only $q_{m+1}^-$ contributes, so
\begin{equation}
\begin{aligned}
\lambda_{\min}(\bMcal^k)\mu_{R0}(c)^2\|\bgamma_{m+1}^*\|^2 
&\le \bgamma_{m+1}^{*\top}\bDcal_{m+1}^\top\bMcal^k\bDcal_{m+1}\bgamma_{m+1}^* \\
&\le \|\bgamma_{m+1}^*\|Q_{m+1}.
\end{aligned}
\end{equation}
If $\bDcal_{m+1}$ is empty, then $\bgamma_{m+1}^*=\mathbf 0$ and the estimate is trivial. Otherwise, if $\bgamma_{m+1}^*=\mathbf 0$ there is nothing to prove; if not, dividing by $\|\bgamma_{m+1}^*\|$ yields
\begin{equation}\label{eq:gamma_Z_bounds}
\|\bgamma_{m+1}^*\|\le\Gamma_{m+1}:=\frac{Q_{m+1}}{\lambda_{\min}(\bMcal^k)\,\mu_{R0}(c)^2}, \quad Z_{m+1}:=D_K\Gamma_{m+1}.
\end{equation}
Since $Q_{m+1}$ is independent of $\Delta t$, Eq.~\eqref{eq:gamma_Z_bounds} shows that $\Gamma_{m+1}$ is also independent of $\Delta t$, and hence so is $Z_{m+1}$. By the configuration update in Eq.~\eqref{eq:relcp},
\begin{equation}
\bCcal_{m+2}-\bCcal_{m+1}
=
\Delta t\,\bGcal^k\bMcal^k\Bigl(\bDcal_{m+1}\bgamma_{m+1}^*-\bDcal_m\iota_{N_C^{m+1}}(\bgamma_m^*)\Bigr).
\end{equation}
Therefore, using the triangle inequality together with $\|\bDcal_{m+1}\bgamma_{m+1}^*\|\le Z_{m+1}$ and $\|\bDcal_m\bgamma_m^*\|\le Z_m$,
\begin{equation}
\|\bCcal_{m+2}-\bCcal_{m+1}\|\le\Delta t\,\|\bGcal^k\|\|\bMcal^k\|(Z_{m+1}+Z_m)=:B_{m+1}\Delta t.
\end{equation}
Set $S_{m+1}:=S_m+B_{m+1}$ and choose $\delta_{m+1}\le\delta_m$ so that $S_{m+1}\delta_{m+1}<\rho_*$ and $L_KS_{m+1}\delta_{m+1}<\eta_*(c)$. This is exactly the inductive statement with $m+1$ in place of $m$, so the induction closes. Finally, setting $\Delta t_{\mathrm{reg}}(c):=\delta_c$, every $0<\Delta t\le\Delta t_{\mathrm{reg}}(c)$ keeps $\bCcal_n\in K$ for $n=0,\ldots,c$, and therefore $\Phi_{ab}(\bCcal_n)>-\epsilon_{\mathrm{reg}}$ for every pair appearing in the $c$-prefix.
This construction depends only on the chosen compact regularity ball and on the fixed prefix length. Therefore, for any prescribed compact ball $K^\sharp$ contained in the common regularity neighborhood of $\bCcal^k$, the same finite-prefix argument produces a threshold $\Delta t_{\mathrm{reg}}(K^\sharp,c)>0$ that keeps the first $c$ iterates in $K^\sharp$.
\end{enumerate}

\item \textbf{First-order pulled-back model error:}
For constraint $\alpha$ discovered at recursion $j(\alpha)\le m\le c$, define the first-order pulled-back model
\begin{equation}
L_{\alpha,m}:=\Phi_{a_\alpha b_\alpha}(\bCcal_{j(\alpha)})+\nabla_{\bCcal}\Phi_{a_\alpha b_\alpha}(\bCcal_{j(\alpha)})^\top(\bCcal_m-\bCcal_{j(\alpha)}).
\end{equation}
By Taylor's theorem and~\eqref{eq:K_bounds_proof},
\begin{equation}
\left|\Phi_{a_\alpha b_\alpha}(\bCcal_m)-L_{\alpha,m}\right|\le\tfrac{1}{2}H_K\Bigl(\sum_{\ell=j(\alpha)}^{m-1}B_\ell\Bigr)^{\!2}\Delta t^2,
\end{equation}
so the true signed separation and the first-order pulled-back model differ by $O(\Delta t^2)$ on any fixed finite prefix.

\end{enumerate}
\end{proof}

\subsection{Proof of Corollary~\ref{cor:strictly_convex_finite_termination}: Strictly convex finite-termination existence--uniqueness regime for ReLCP contact resolution}\label{app:proof:cor_strictly_convex_finite_termination}
\begin{proof}
Because the body set is finite and $\bCcal^k$ is overlap-free, Lemma~\ref{lemma::local_regularity} gives, for each ordered pair $(a,b)$, an open neighborhood $V_{ab}$ of $\bCcal^k$ and a constant $\epsilon_{ab}>0$ such that $\Phi_{ab}$ is single-valued and $C^2$ on
\begin{equation}
V_{ab}\cap\{\bCcal:\Phi_{ab}(\bCcal)>-\epsilon_{ab}\}.
\end{equation}
Intersect these finitely many neighborhoods and choose $\rho_0>0$ and $\epsilon_{\mathrm{reg}}>0$ such that
\begin{equation}
K_0:=\overline{B_{\rho_0}(\bCcal^k)}\subset \bigcap_{a\neq b}V_{ab},
\,\,
\Phi_{ab}(\bCcal)>-\epsilon_{\mathrm{reg}}
\,\forall\,\bCcal\in K_0,\ a\neq b.
\end{equation}
Thus every ordered-pair shared-normal map is single-valued on $K_0$. Applying Lemma~\ref{lemma::uniform_interior_ball} pairwise on $K_0$ and shrinking the common tangent-ball radius if necessary yields $r_*>0$ and compact body-fixed center sets $\Sigma_{ab}^a$ and $\Sigma_{ab}^b$ for every ordered pair $(a,b)$. Shrink $\epsilon_{\mathrm{reg}}$ further, if necessary, so that
\(
\epsilon_{\mathrm{reg}}\le 2r_*.
\)
Define the geometric packing constant
\begin{equation}
C_{\max}:=\sum_{a\neq b}P\!\left(\Sigma_{ab}^a\times\Sigma_{ab}^b,\,\epsilon_{NCP};\,d_\times\right)<\infty.
\end{equation}
This depends only on the fixed compact regularity set $K_0$. All subsequent occurrences of $r_*$, $\epsilon_{\mathrm{reg}}$, and $\Sigma_{ab}^{a},\Sigma_{ab}^{b}$ refer to these final choices.

Apply the finite-prefix construction from the proof of Lemma~\ref{lemma::timestep_regularity} to the prescribed compact ball $K_0$ with prefix length $C_{\max}+1$. Under Assumption~\ref{assume:small-dt}, after decreasing the timestep bound if necessary, we may ensure that
\begin{equation}
\bCcal_n\in K_0
\qquad \forall n=0,1,\dots,C_{\max}+1.
\end{equation}
Hence
\begin{equation}
\Phi_{ab}(\bCcal_n)>-\epsilon_{\mathrm{reg}}\ge -2r_*
\,\, \forall a\neq b,\ \forall n\le C_{\max}+1,
\end{equation}
so condition~\eqref{eq:n_body_pairwise_overlap_regime} holds on this finite prefix with $U_{ab}:=K_0$.

We prove solvability for every $n=0,1,\dots,C_{\max}+1$ by induction. At $n=0$, there are no retained constraints, so every constraint is new. Choosing the interior reference points in~\eqref{eq:admissible_reference_region} to be the tangent-ball centers supplied by Lemma~\ref{lemma::uniform_interior_ball} gives, for each step-$0$ constraint $\alpha$,
\(
\eta_{\alpha,0}^{\mathrm{new}}=2r_*>0.
\)
Since $\bCcal_0=\bCcal^k$ is overlap-free, $\Phi_\alpha(\bCcal_0)\ge 0$, so Lemma~\ref{lemma::self_jamming_exclusion} applies at step $0$ and yields
\begin{equation}
\Ker\bDcal_0\cap\mathbb R_+^{N_C^0}=\{\mathbf 0\}.
\end{equation}
By~\eqref{eq:kernel_equality}, also $\Ker\bAcal_0\cap\mathbb R_+^{N_C^0}=\{\mathbf 0\}$, so Theorem~\ref{theorem::is_a_relcp} gives solvability at $n=0$.

Assume solvability through step $n-1$, where $1\le n\le C_{\max}+1$. For each retained constraint $\alpha\in\mathcal A_{n-1}$, Assumption~\ref{assume:retention} carries forward the corresponding scalar component, and the previous LCP solution is nonnegative, so Lemma~\ref{lemma::pairwise_exclusion_from_relcp} gives the retained frozen-normal exclusion property~\eqref{eq:n_body_normal_exclusion} at step $n$. For each new constraint $\alpha\in\Delta\mathcal A_n$, the inclusion $\bCcal_n\in K_0$ lets us choose the tangent-ball centers from Lemma~\ref{lemma::uniform_interior_ball} as the interior reference points in~\eqref{eq:admissible_reference_region}. Then
\(
\eta_{\alpha,n}^{\mathrm{new}}=2r_*.
\)
Since $\Phi_\alpha(\bCcal_n)>-\epsilon_{\mathrm{reg}}\ge -2r_*$ on $K_0$, we obtain
\begin{equation}
\Phi_\alpha(\bCcal_n)+\eta_{\alpha,n}^{\mathrm{new}}>0.
\end{equation}
Thus Lemma~\ref{lemma::self_jamming_exclusion} applies at step $n$, so
\begin{equation}
\Ker\bDcal_n\cap\mathbb R_+^{N_C^n}=\{\mathbf 0\}.
\end{equation}
Using~\eqref{eq:kernel_equality}, Theorem~\ref{theorem::is_a_relcp} gives solvability at step $n$. By induction, the recursion is solvable for every $n\le C_{\max}+1$.

Let $c$ denote the first recursion index at which the stopping rule~\eqref{eq:gap_stopping_rule} holds. Suppose for contradiction that $c>C_{\max}+1$ (or that no such index exists). Then the stopping rule fails at every recursion $n=1,\dots,C_{\max}+1$. Because Assumption~\ref{assume:retention} augments all violating pairs, for each such $n$ we may choose one violating ordered pair $(a_n,b_n)$ satisfying
\begin{equation}
-\Phi_{a_n b_n}(\bCcal_n)\ge \epsilon_{NCP}.
\end{equation}
The finite sequence $\{\bCcal_n\}_{n=1}^{C_{\max}+1}$ together with these designated active pairs satisfies the hypotheses of Theorem~\ref{theorem::finite_epsilon_normal_exclusion_n_body}: the compact sets $U_{ab}=K_0$ lie in the local regularity regime, condition~\eqref{eq:n_body_pairwise_overlap_regime} holds on $K_0$, and the retained pairwise exclusion property follows from the induction above via Lemma~\ref{lemma::pairwise_exclusion_from_relcp}. Therefore
\begin{equation}
\#\{1\le n\le C_{\max}+1:-\Phi_{a_n b_n}(\bCcal_n)\ge \epsilon_{NCP}\}
\le
C_{\max},
\end{equation}
contradicting the construction of $(a_n,b_n)$ for every $n=1,\dots,C_{\max}+1$. Hence $c\le C_{\max}+1<\infty$. At this first terminal index, the stopping-rule implication in Theorem~\ref{theorem::is_a_relcp} gives configurational convergence, so the recursion terminates after finitely many iterations.

For uniqueness, view the contact recursion as Case~2 in Definition~\ref{def::relcp} with auxiliary state $\bs_n:=\bCcal_n$. The initial state $\bCcal_0=\bCcal^k$ is fixed, and Eq.~\eqref{eq:relcp} determines $\bCcal_{n+1}$ from the previously computed wrench $\bDcal_n\bx_n^*$. On $K_0$, Lemma~\ref{lemma::local_regularity} makes the violated-pair set and corresponding Jacobian rows functions of $\bCcal_n$ alone, while Assumption~\ref{assume:retention} fixes which earlier rows persist. Thus $\bDcal_n$ is determined by $\bs_n$, and Corollary~\ref{lemma::relcp_uniqueness} yields uniqueness of the terminal wrench $\bDcal_c\bx_c^*$, hence also of the terminal velocity $\bUcal_c^*=\bMcal\bDcal_c\bx_c^*$.
\end{proof}

\subsection{Proof of Corollary~\ref{cor:relcp_reduction_to_snsd_lcp}: Reduction of the adaptive constraint generation scheme to the classical single-constraint SNSD LCP}\label{app:proof:cor_relcp_reduction_to_snsd_lcp}
\begin{proof}
At recursion $0$, the ReLCP is exactly the classical single-constraint shared-normal signed-distance LCP formulation of Stewart and Anitescu \cite{stewart_first_lcp_1996, anitescu_2006}. Fix the discrete state $\bCcal^k$ and let $\bCcal_{1}^{k+1}(\Delta t)$ denote the trial configuration produced by this initial SNSD-LCP solve. Define
\begin{equation}
\eta_1(\Delta t)
\coloneqq
\max_{a\neq b}
\left|\min\!\left(\Phi_{ab}\!\left(\bCcal_1^{k+1}(\Delta t)\right),0\right)\right|,
\end{equation}
the maximum true overlap remaining after the recursion-$0$ update.

By the classical consistency results for the SNSD-LCP time-stepping scheme \cite{stewart_convergence_lcp_1998, anitescu_convergence_2008}, summarized in Section~\ref{sec:preliminaries}, the recursion-$0$ update converges as $\Delta t\to0$ to the corresponding nonpenetrating solution of the underlying differential variational inequality. Denote the limiting post-step configuration by $\bCcal^{\mathrm{DVI}}(t^{k+1})$. Then
\begin{equation}
\Phi_{ab}\!\left(\bCcal^{\mathrm{DVI}}(t^{k+1})\right)\ge0
\qquad \text{for every body pair }(a,b).
\end{equation}
By Lemma~\ref{lemma::timestep_regularity} with $c=1$, there exists $\Delta t_{\mathrm{reg}}(1)>0$ such that, for every $0<\Delta t\le\Delta t_{\mathrm{reg}}(1)$, the trial configuration $\bCcal_1^{k+1}(\Delta t)$ remains in the local regularity regime determined by $\bCcal^k$. On that regime, each relevant true signed separation function $\Phi_{ab}$ is continuous in configuration. Hence, for every body pair $(a,b)$ and in the limit $\Delta t\to0,$
\begin{equation}
\min\!\left(\Phi_{ab}\!\left(\bCcal_1^{k+1}(\Delta t)\right),0\right)
\rightarrow
\min\!\left(\Phi_{ab}\!\left(\bCcal^{\mathrm{DVI}}(t^{k+1})\right),0\right)
=0.
\end{equation}
Because the number of bodies is finite, the maximum over all body pairs preserves this limit, and therefore
$
\eta_1(\Delta t)\to0
\,\, \text{as } \Delta t\to0.
$
Now fix $\epsilon_{NCP}>0$. By the preceding limit, there exists
\begin{equation}
\Delta t_{\mathrm{red}}(\bCcal^k,\epsilon_{NCP})>0
\end{equation}
such that
\begin{equation}
\eta_1(\Delta t)\le \epsilon_{NCP}
\qquad \forall \, 0<\Delta t\le \Delta t_{\mathrm{red}}(\bCcal^k,\epsilon_{NCP}).
\end{equation}
But this is exactly the no-violation condition in the stopping rule~\eqref{eq:gap_stopping_rule} evaluated after the recursion-$0$ solve. Hence no new constraints are generated, i.e., the adaptive scheme appends nothing beyond the initial SNSD-LCP constraint set. Therefore the recursion terminates immediately after the initial solve, and the converged ReLCP solution is exactly the classical SNSD-LCP solution for every $0<\Delta t\le \Delta t_{\mathrm{red}}(\bCcal^k,\epsilon_{NCP})$.
\end{proof}


\bibliographystyle{asmejour}   
\bibliography{ref} 



\end{document}